\documentclass[runningheads]{llncs}

\usepackage{eccv}

\usepackage{eccvabbrv}

\usepackage{graphicx}
\usepackage{array}
\usepackage{booktabs}
\usepackage{pifont}     
\usepackage{amsmath}
\usepackage{amssymb}
\usepackage{subcaption}
\usepackage{multirow}
\usepackage{fontspec}
\usepackage{xcolor}
\usepackage{fontspec}

\usepackage{hyperref}

\usepackage{orcidlink}

\usepackage{todonotes}

\newcommand{\mural}{\textsc{Mural}}

\begin{document}

\title{\mural: Transferring LLM knowledge to image generation via Mixture-of-Transformers}

\titlerunning{\mural}

\author{Achin Jain
\and
Jie An \and
Siddharth Chaudhary \and
Davide Modolo
}

\authorrunning{A.~Jain et al.}

\institute{Amazon AGI \\
\email{\{achij,jiean,sidchoud,dmodolo\}@amazon.com}}

\maketitle

\begin{abstract}

    Leveraging capabilities of large language models (LLMs) in text-to-image (T2I) synthesis is an important research direction. In this work we investigate whether the knowledge of a frozen LLM can be effectively utilized in T2I generation when trained exclusively on standard text–image pairs. We integrate a frozen, reasoning-capable LLM with a diffusion-based image generator via shared attention within the Mixture-of-Transformers (MoT) architecture. Our experiments span two critical questions: (1) what degree of the LLM’s intrinsic knowledge remains accessible during T2I training, and (2) what novel capabilities emerge in the resulting system. Across established benchmarks, our models achieve strong performance among unified understanding-generation systems: 0.85 on GenEval, 86.75 on DPG-Bench, and 0.66 on WISE with inference-time reasoning, using only text-image data. Remarkably, we uncover emergent behaviors absent from training data, including cross-lingual image generation, color-guided composition, emoji / ASCII scene construction, and generation directed by world knowledge. These results demonstrate that pretrained LLM knowledge can guide image synthesis under standard text–to-image training paradigms, without interleaved multimodal signals or explicit reasoning supervision. Our findings open new avenues for harnessing frozen model capabilities in resource-constrained multimodal learning.
    
\end{abstract}

\section{Introduction}
\label{S:intro}
Adding image generation to large language models (LLMs) while preserving their understanding capabilities remains an open challenge. Recent unified multimodal models broadly fall into two categories with distinct trade-offs.
\textit{Joint multimodal training} approaches~\cite{chameleon2024,transfusion2024,januspro2025,bagel2025} update all model parameters using mixed-modality data. While this strategy can produce strong image generation quality~\cite{bagel2025}, it often degrades language understanding and requires carefully curated data mixtures such as interleaved image–text sequences~\cite{chameleon2024,bagel2025} or reasoning supervision like chain-of-thought (CoT) demonstrations~\cite{bagel2025}. Designing and optimizing these training mixtures can be costly and complex.
\textit{Frozen-LLM} approaches~\cite{metaquery2025,uniworld2025,omnigen2_2025} instead keep a pretrained LLM fixed and add generation capabilities on top, preserving language understanding by construction. However, existing methods in this category lag behind in image generation quality or rely on additional supervision such as reasoning data~\cite{omnigen2_2025} or image-to-text captioning pairs~\cite{uniworld2025,omnigen2_2025}.

Within this latter framework, a central question remains underexplored: \textit{to what extent can the knowledge of a frozen LLM be leveraged for text-to-image generation when training only on text–image pairs?} In particular, can shared attention serve as an effective architectural interface for transferring knowledge from a frozen LLM to an image generation model -- without modifying its parameters or requiring additional interleaved or reasoning supervision?

In an attempt to address this question, we present \mural, which uses the Mixture-of-Transformers (MoT) architecture~\cite{mot2024} to pair a trainable diffusion-based image generation expert with a frozen LLM, sharing attention at every transformer layer. The generation branch is trained using only text–image pairs with the LLM parameters fixed.
This setup allows us to isolate the contribution of shared attention from data and recipe effects, preserve language understanding by construction, eliminate the complex data mixtures required by~\cite{bagel2025}, and seamlessly incorporate advances from the rapidly evolving LLM landscape.

Through controlled experiments against a dense baseline trained on identical data and recipe, we find that MoT enables the generation model to leverage the frozen LLM's representations without degrading either language understanding or image generation. More strikingly, several capabilities emerge that are absent from the training data: cross-lingual generation, color guidance, emoji and ASCII scene composition, and generation guided by world knowledge. These behaviors arise without any task-specific supervision, reasoning data, or interleaved multimodal data, confirming they stem from shared attention with the frozen LLM rather than from training data.

On image generation benchmarks, \mural~achieves strong performance while being trained using only text–image pairs. Our results suggest that shared attention with a frozen reasoning-capable LLM is an effective mechanism for transferring knowledge from language models to image generation. We summarize the contributions of this work as follows:
\vspace{-5pt}
\begin{itemize}
    \item We show that the knowledge of a frozen LLM can be leveraged for text-to-image generation when training only on text–image pairs, giving rise to emergent capabilities -- including cross-lingual generation, emoji-to-scene composition, color guidance, and world-knowledge-based generation -- without any explicit task-specific supervision.
    \item Through controlled experiments, we show that the MoT architecture with a frozen LLM strictly improves over a dense model trained with identical data and recipe across image generation benchmarks.
    \item Using only text–image pairs, \mural~achieves strong performance on GenEval~\cite{geneval2023} (0.85), DPG-Bench~\cite{dpgbench2024} (86.75), and WISE~\cite{wise2024} (0.66), demonstrating competitive among unified multimodal and dedicated text-to-image models.
    \item We analyze scaling across 1.5B, 3B, and 7B LLMs with both text and vision-language backbones, finding that multimodal initialization accelerates convergence at smaller scales.
\end{itemize}

\section{Related work}
\label{S:related}

\textbf{Text-to-image generation.}
Text-to-image generation has primarily been studied under two paradigms: diffusion-based and autoregressive models. Diffusion models~\cite{ddpm2020,ldm2022,sdxl2023,sd3_2024,flux2024} synthesize images by progressively denoising latent representations conditioned on text embeddings, and scaling diffusion transformers (DiT)~\cite{dit2023} has further improved generation quality in large-scale systems such as FLUX~\cite{flux2024} and Seedream~\cite{seedream2025}. Autoregressive approaches~\cite{llamagen2024,mar2024,infinity2024} instead generate images as sequences of discrete visual tokens using next-token prediction. Most existing systems rely on pretrained text encoders or language models primarily used for textual conditioning~\cite{qwenimage2025}, limiting the extent to which LLM knowledge and reasoning capabilities influence image generation.

\noindent\textbf{Unified understanding and generation models.}
Recent work explores unified architectures that support both multimodal understanding and image generation within a single model. Early-fusion models such as Chameleon~\cite{chameleon2024} generate interleaved text–image sequences autoregressively, while Transfusion~\cite{transfusion2024} combines autoregressive language modeling with diffusion-based image generation in a single transformer. Other approaches mitigate conflicts between understanding and generation by decoupling visual pathways, e.g., LMFusion~\cite{lmfusion2025}, Janus/Janus-Pro~\cite{januspro2025}), BAGEL~\cite{bagel2025}, Cosmos 3~\cite{cosmos32026}), often involving large interleaved multimodal pretraining. Systems such as OmniGen2~\cite{omnigen2_2025}, UniWorld~\cite{uniworld2025}, and MetaQuery~\cite{metaquery2025} investigate ways to preserve language understanding while interfacing language models with diffusion-based generation.

\noindent\textbf{Mixture-of-Transformers.}
The Mixture-of-Transformers (MoT)~\cite{mot2024} architecture extends the Mixture-of-Experts (MoE) paradigm~\cite{moe_shazeer2017} to multimodal modeling. It introduces modality-specific parameters while sharing self-attention across modalities, enabling tokens from different modalities to interact through attention while retaining modality-specialized processing. Recent unified multimodal systems such as LMFusion~\cite{lmfusion2025}, BAGEL~\cite{bagel2025}, and Cosmos 3~\cite{cosmos32026} adopt this design to support both understanding and generation within a single architecture.

\noindent\textbf{Cross-modal knowledge transfer.}
Recent work has explored incorporating reasoning processes into image generation. For example, BAGEL~\cite{bagel2025} introduces chain-of-thought reasoning during training to guide generation, while models such as GPT-4o~\cite{gpt4o2024} suggest that strong language understanding can enhance multimodal generation. In this work, we study whether knowledge from a frozen LLM can influence text-to-image generation when training is restricted to standard text–image pairs, without explicit reasoning supervision or multimodal data mixtures.

\section{Framework}
\label{S:method}

Our goal is to study how well the knowledge of a frozen LLM can be leveraged for text-to-image generation when training only on text–image pairs.
To this end, we connect a diffusion-based image generation branch to a frozen LLM using the Mixture-of-Transformers (MoT) architecture~\cite{mot2024}. 
In this framework, the LLM parameters remain fixed while a trainable image generation expert learns to synthesize images conditioned on representations produced by the LLM.

In our experiments, we instantiate the frozen backbone with instruction-tuned Qwen2.5~\cite{qwen25_2024} or Qwen2.5-VL~\cite{qwen2vl2024}. 
This design allows us to leverage the knowledge and reasoning capabilities of modern LLMs while studying how such capabilities interact with image generation.

\subsection{Model setup}
\label{SS:arch}

\begin{figure}[t]
\centering
\includegraphics[width=0.75\linewidth]{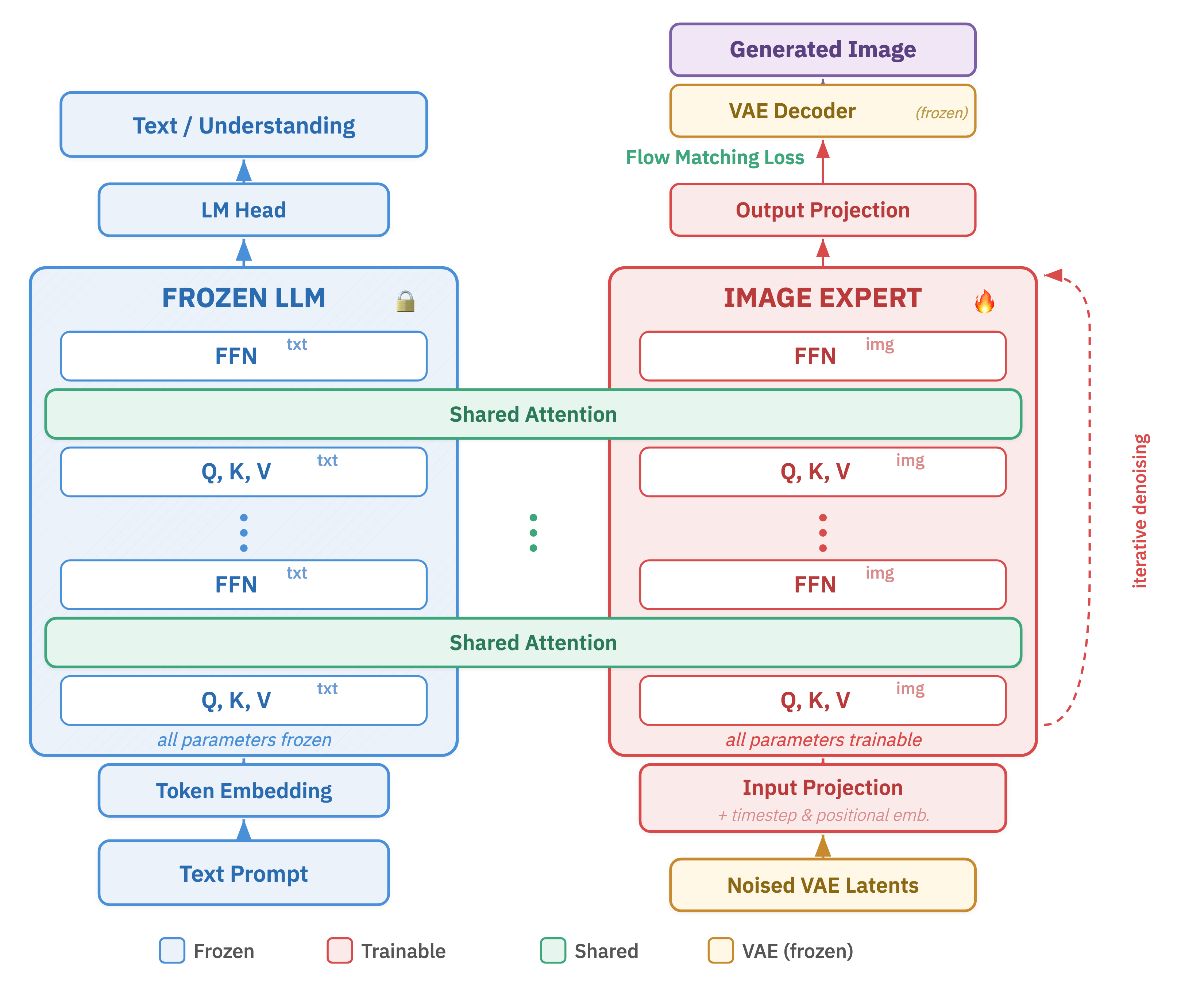}
\caption{\mural~architecture overview. The LLM is frozen and only the image generation expert is trained with a flow matching loss.}
\label{fig:architecture}
\end{figure}

Our framework adopts the Mixture-of-Transformers (MoT) architecture~\cite{mot2024}, which has previously been explored for multimodal modeling~\cite{bagel2025}. 
MoT handles multi-modal input sequences by routing tokens through modality-specific experts while sharing global self-attention across all tokens. This design enables cross-modal information exchange through attention while preserving modality-specific processing capacity.

Our setup follows~\cite{bagel2025}, with the difference that the pretrained LLM remains entirely frozen.
We attach a trainable image generation expert that processes VAE latent tokens and interacts with text tokens through the shared attention mechanism (Figure~\ref{fig:architecture}).
Specifically, we introduce an image generation expert branch. For each transformer layer, we add image-specific components: (1) layer norms, (2) query, key, value, and output projections, (3) QK norms, and (4) a feed-forward network (FFN) with the same hidden dimension as the LLM.
We note that while the Qwen2.5 and Qwen2.5-VL model families do not employ QK norms. We add them only to the image expert branch for T2I training.

During the forward pass, modality-specific tokens are first routed through their respective layer norms, QKV projections, and QK norms (image generation expert only in our case). The resulting sequences are then concatenated to compute shared attention across both modalities. Afterward, each modality's tokens are routed back through their respective FFN layers. Unlike standalone image generation models like DiT~\cite{dit2023}, UViT~\cite{uvit2023}, that concatenate text and image tokens and apply bidirectional attention to both, we preserve causal attention for text tokens while using bidirectional attention for image tokens.

In addition to the expert branch, we introduce an image input projection layer that maps VAE-encoded features into the LLM's hidden dimension, and a corresponding output projection layer that transforms hidden states back before decoding with the VAE decoder.
We use FLUX.1~\cite{flux2024} VAE to encode/decode images in latent space.
We add timestep and 2D positional embeddings at the input projection stage. 

\subsection{Training protocol}
\label{SS:training}

A key aspect of our study is the use of a simplified training setup designed to isolate the role of the frozen LLM.

\noindent\textbf{Training data.}
We train exclusively on standard text–image pairs commonly used in text-to-image generation. 
Unlike many unified multimodal models that rely on interleaved image–text sequences~\cite{chameleon2024,bagel2025}, image-to-text captioning data~\cite{januspro2025,uniworld2025}, or reasoning supervision~\cite{bagel2025,omnigen2_2025}, our model does not use any additional data types.
This setup allows us to examine how the knowledge of the frozen LLM can be utilized to influence image generation and especially demonstrate that shared attention alone unlocks capabilities that cannot be attributed to training data.
All our data is in English and went through a series of standard data cleaning and enrichment steps: filtering based on resolution, aesthetic scores, watermarks, NSFW, de-duplication, and dense captioning.

\noindent\textbf{Training loss.}
We train the image expert branch to directly denoise continuous VAE latent tokens using a flow matching objective~\cite{flowmatching2023} with a linear path from data to noise, applying no loss to text tokens. During training, we add Gaussian noise to the VAE-encoded image latents, and the image expert learns to predict the denoised latents conditioned on text representations received via shared attention.

\noindent\textbf{Training curriculum.}
We adopt a progressive resolution training strategy ($256\text{p} \rightarrow 512\text{p} \rightarrow 1024\text{p}$), with dynamic aspect ratios at each stage. We pre-train \mural~for 500K steps at 256p (batch size 2048), 300K steps at 512p (batch size 2048), and 200K steps at 1024p (batch size 1024), followed by supervised fine-tuning (SFT) for 8K steps on high-quality images (batch size 128). With a VAE downsampling factor of 8 and a patch size of 2, we train on $\sim$2T tokens in total. Following SD3~\cite{sd3_2024}, we increase the timestep shift to 2 at 512p and 4 at 1024p to concentrate sampling towards the middle of the flow path. We use AdamW~\cite{adamw2019} optimizer and a constant learning rate with linear warmup across all stages.

\subsection{Inference modes}
\label{SS:inference}

To analyze how the frozen LLM influences generation, we evaluate two inference settings.

\noindent\textbf{Standard generation.}
We prompt the model in the same format using during training. Even without chain-of-thought prompting (see next paragraph), \mural~benefits from knowledge transfer via shared attention with the frozen LLM. For instance, it can generate images from prompts in multiple languages supported by the LLM, despite all training image-text data being in English.

{\footnotesize
\begin{verbatim}
<|im_start|>{prompt}<|im_end|><|vision_start|>{noisy_latents}
\end{verbatim}
}

\noindent\textbf{Chain-of-thought enhanced generation.}
Instruction-tuned Qwen2.5 models support chain-of-thought reasoning through their chat templates. 
At inference time, we optionally allow the LLM to generate intermediate reasoning tokens before image generation. 
The generation branch attends to these tokens through the shared attention mechanism.

{\footnotesize
\begin{verbatim}
<|im_start|>system
You are an image generation assistant. Add visual details based
on real-world knowledge.<|im_end|>
<|im_start|>user
{prompt}<|im_end|>
<|im_start|>assistant
{thinking_trace}<|im_end|><|vision_start|>{noisy_latents}
\end{verbatim}
}

Importantly, this capability is a direct consequence of the frozen LLM's reasoning abilities, requiring no reasoning supervision during T2I training.
This inference setting therefore allows us to probe whether reasoning capabilities of the frozen LLM can influence generation.

\section{Experiments}
\label{S:exps}

We evaluate \mural~across four dimensions:
(1) emergent capabilities that arise from the MoT architecture without explicit training data for these tasks (Section~\S\ref{SS:emergent}),
(2) comparison with a dense model (image generation without understanding) trained on identical data and recipe to decouple the benefits of the MoT architecture (Section~\S\ref{SS:dense_vs_mot}),
(3) comparison to the state-of-the-art unified and standalone image generation models (Section~\S\ref{SS:sota}), and
(4) scaling behavior across 1.5B, 3B, and 7B model sizes with text/multimodal initialization (Section~\S\ref{SS:scaling}).

Unless otherwise noted, we use Qwen2.5-7B-Instruct~\cite{qwen25_2024} as the frozen LLM backbone, yielding a total of 14B parameters in the MoT configuration (7B frozen LLM + 7B image generation expert).
We initialize all parameters of the image generation branch with the chosen backbone and the newly introduced query/key RMS norm~\cite{rmsnorm2019} weights with ones.

\noindent\textbf{Dense image generation expert.}
To better understand the role of shared attention in \mural, we train a dense baseline consisting solely of the image expert branch, also initialized from Qwen2.5-7B-Instruct weights and trained with an identical recipe to the MoT model.
This isolates the contribution of shared attention by removing the frozen LLM from the forward pass entirely.
The dense model has only 7B parameters (compared to 14B in MoT) and naturally loses text understanding capabilities upon training on text-to-image data.

\noindent\textbf{Benchmarks.}
We evaluate text-to-image generation on three benchmarks:
(1) GenEval~\cite{geneval2023} which measures image-text alignment on compositional prompts; (2) DPG-Bench~\cite{dpgbench2024} which assesses alignment with dense text prompts; and (3) WISE~\cite{wise2024} which evaluates world knowledge in generated images.
For text and visual understanding, we report whether the base LLM's original benchmark scores are preserved or degraded.

\subsection{Emergent capabilities via shared attention}
\label{SS:emergent}
We demonstrate capabilities that emerge from shared attention between the frozen LLM and the image generation expert, without explicitly using any interleaved, multi-turn, reasoning, or task-specific training data.
These emergent capabilities are best demonstrated qualitatively as existing benchmarks do not capture such scenarios.
Note that we use Qwen2.5 (text-only LLM) as the frozen backbone for all results in this section.

\noindent\textbf{Cross-lingual generation.}
\begin{figure}[!t]
\centering
\setlength{\tabcolsep}{2pt}
\begin{tabular}{cccc}
\includegraphics[width=0.22\linewidth]{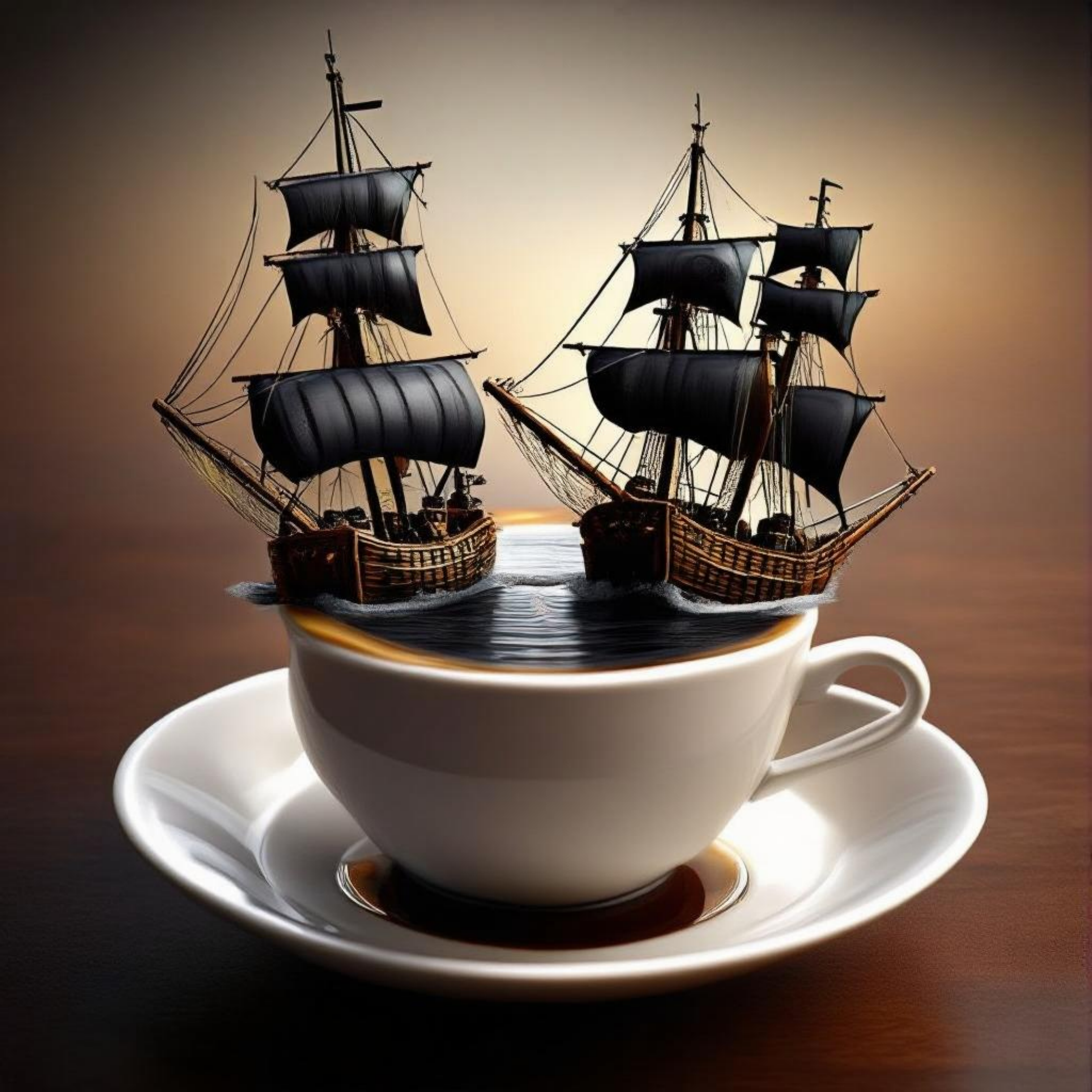} &
\includegraphics[width=0.22\linewidth, trim=0 144 0 144, clip]{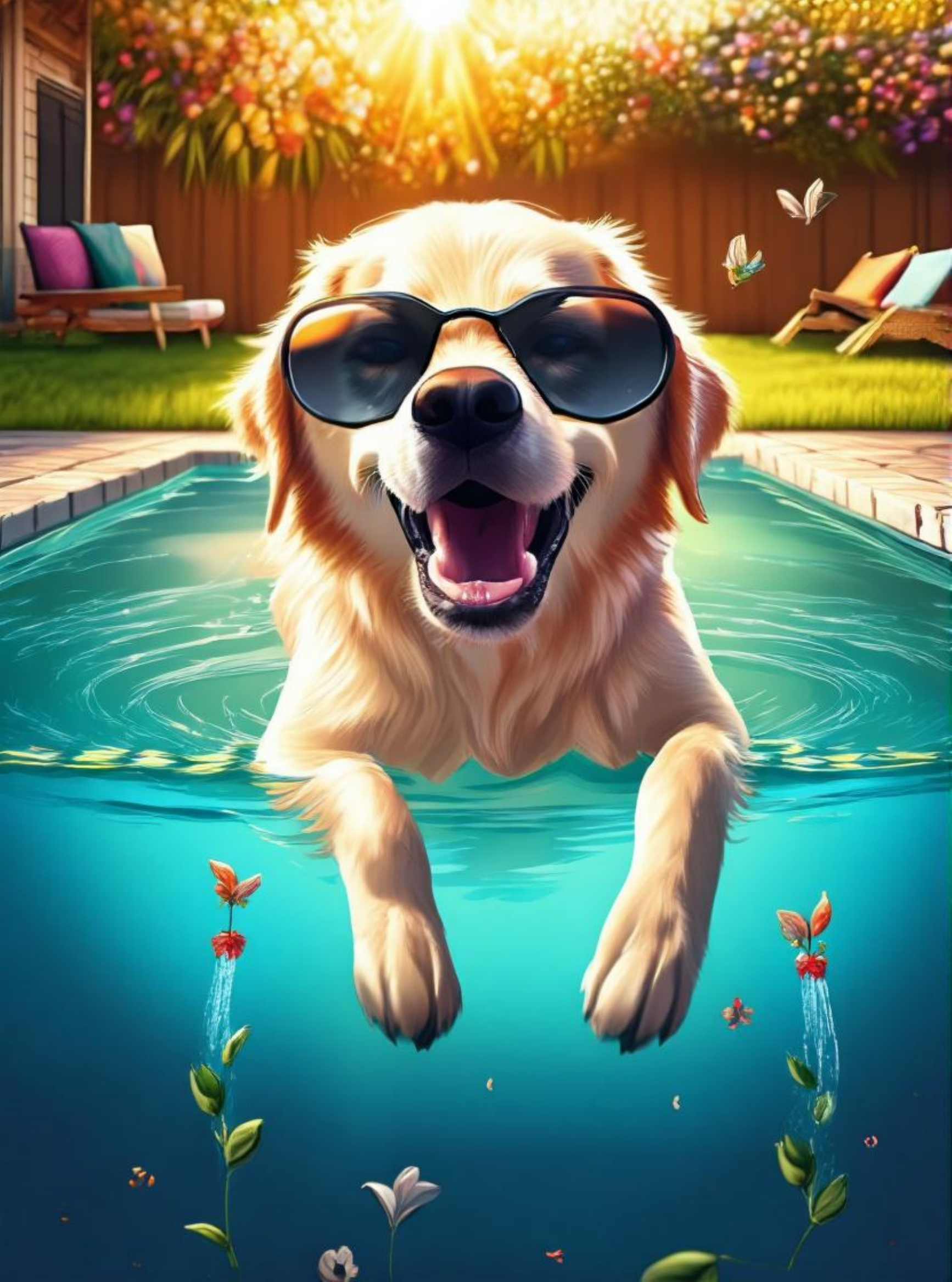} &
\includegraphics[width=0.22\linewidth]{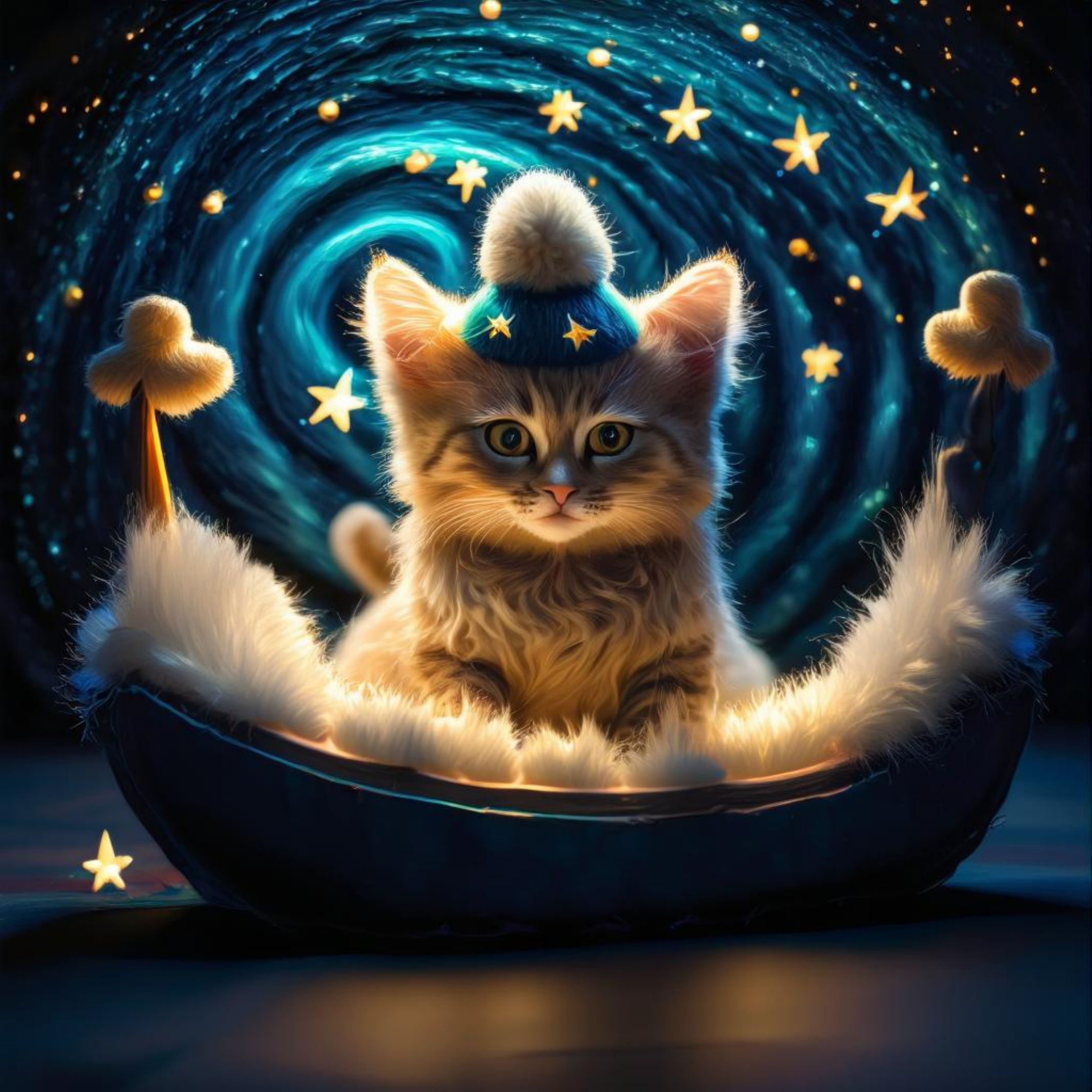} &
\includegraphics[width=0.22\linewidth, trim=144 0 144 0, clip]{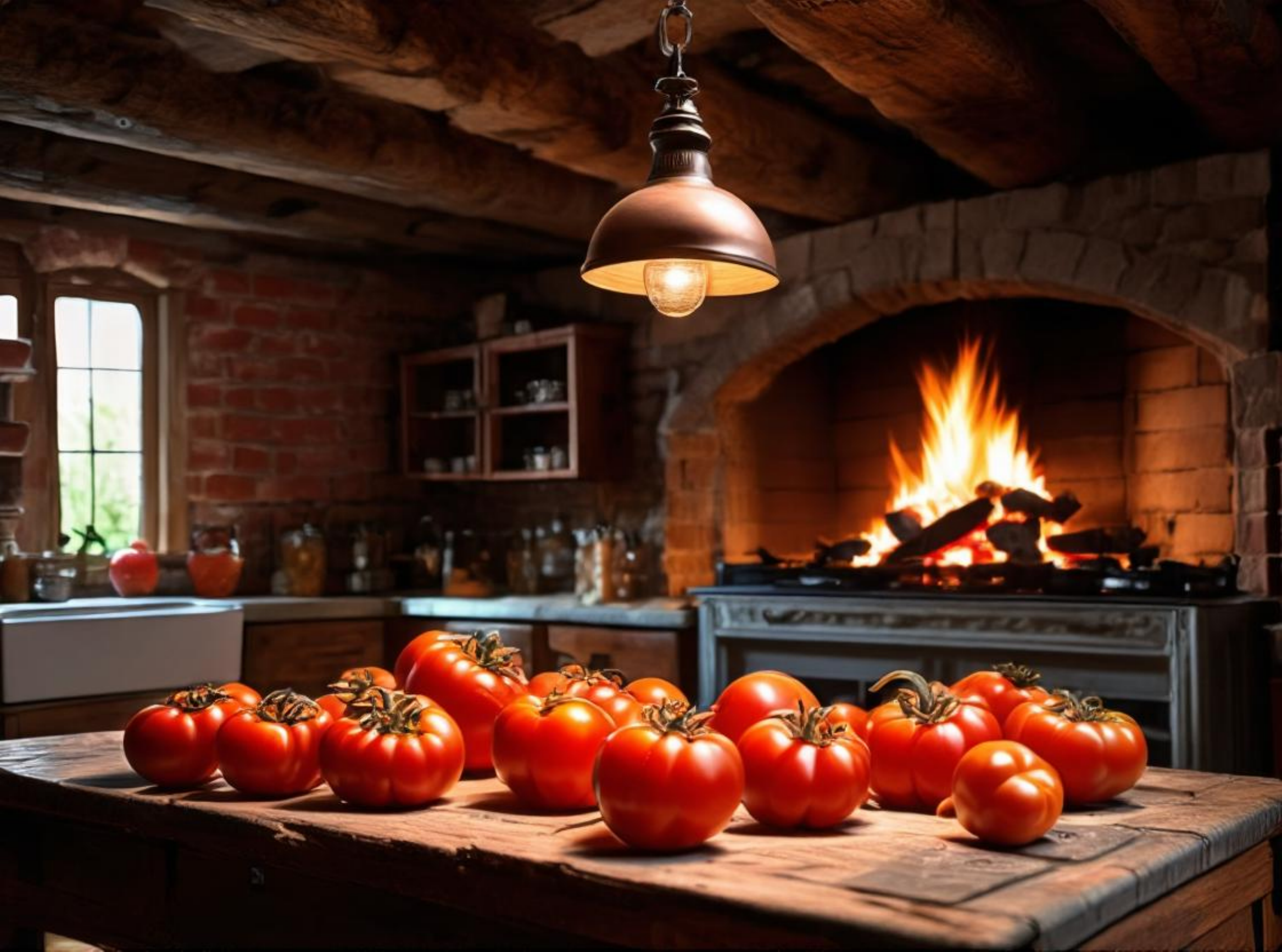} \\[1pt]
\footnotesize English & \footnotesize Hindi & \footnotesize Chinese & \footnotesize Italian \\[2pt]
\includegraphics[width=0.22\linewidth]{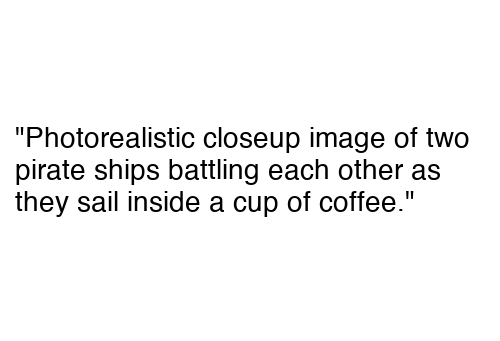} &
\includegraphics[width=0.22\linewidth]{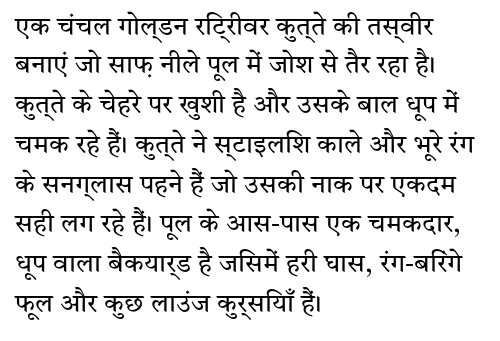} &
\includegraphics[width=0.22\linewidth]{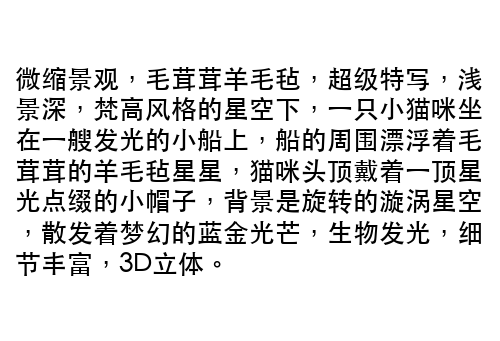} &
\includegraphics[width=0.22\linewidth]{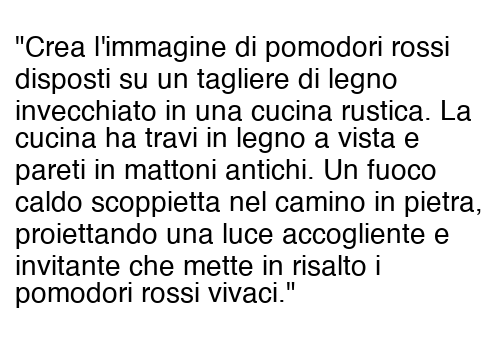} \\
\end{tabular}
\caption{Cross-lingual generation: \mural~generates images from prompts in English, Hindi, Chinese, and Italian despite training \textit{only} on English image-text pairs. The frozen LLM's multilingual understanding transfers to the generation branch via shared attention \textit{without} using CoT at inference.}
\label{fig:crosslingual}
\end{figure}
\mural~naturally enables image generation in all languages supported by the base LLM. Since Qwen2.5 supports over 29 languages, \mural~can generate images from prompts in any of these languages, despite being trained exclusively on English image-text pairs. Figure~\ref{fig:crosslingual} shows generation results on English, Hindi, Chinese, and Italian prompts, demonstrating consistent quality across languages. English and Chinese prompts are taken from~\cite{bagel2025}.
Most notably, we \textit{do not} use CoT at inference which highlights the power of shared attention alone.

\newcommand{\cpimg}[1]{\includegraphics[width=0.155\linewidth]{#1}}
\begin{figure}[!t]
\centering
\setlength{\tabcolsep}{1.5pt}
\scriptsize
\begin{tabular}{@{}cccccc@{}}
\toprule
OmniGen2 & Bagel & Bagel+CoT & \mural & \mural+CoT & NanoBanana2 \\
\midrule
\multicolumn{6}{@{}l@{}}{\scriptsize\texttt{--\textcolor[HTML]{FF1493}{car body}:\,\#FF1493;\;\;--\textcolor[HTML]{00FF00}{tires}:\,\#00FF00;\;\;--\textcolor[HTML]{FFD700}{road}:\,\#FFD700;\;\;--\textcolor[HTML]{8B00FF}{sky}:\,\#8B00FF}} \\[2pt]
\cpimg{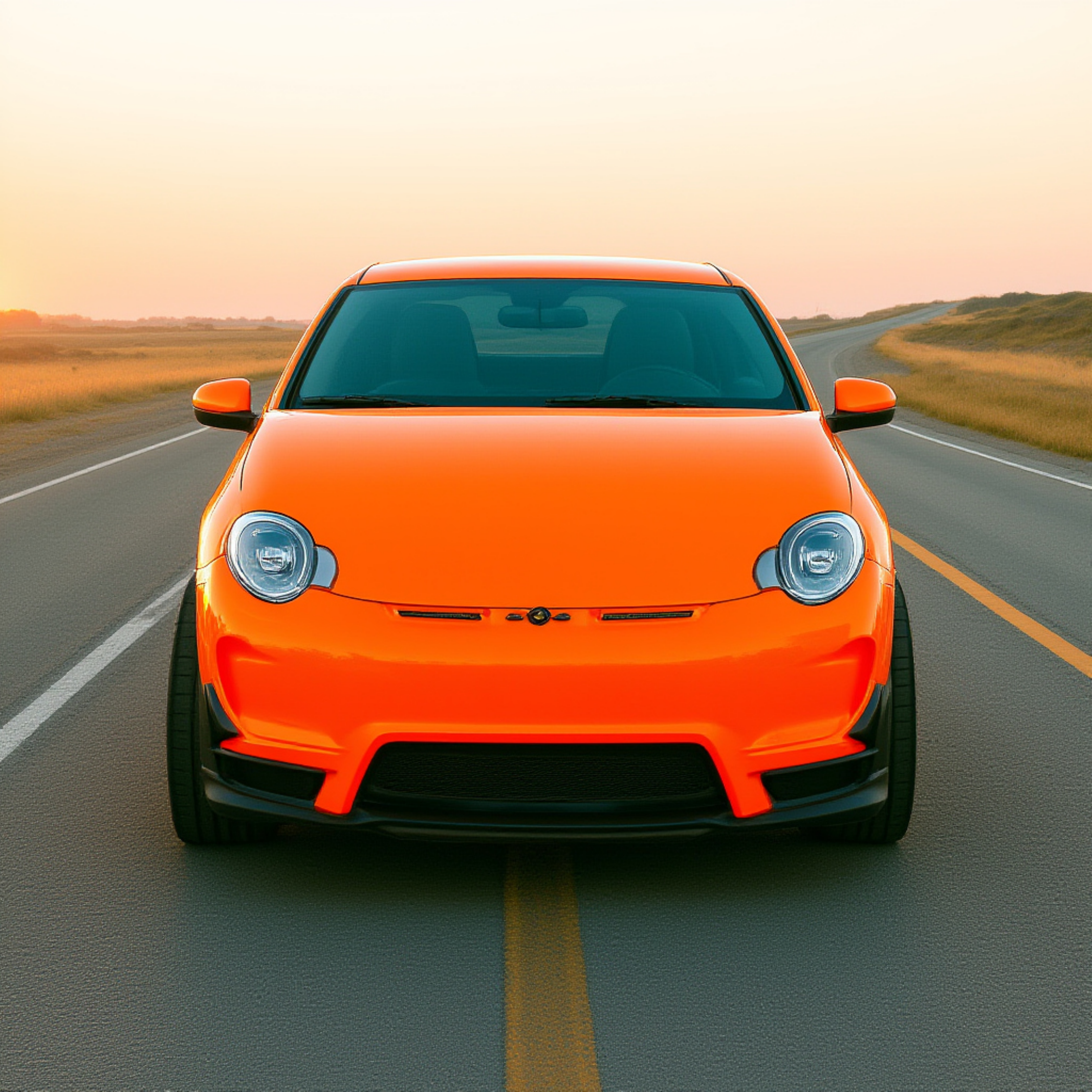} &
\cpimg{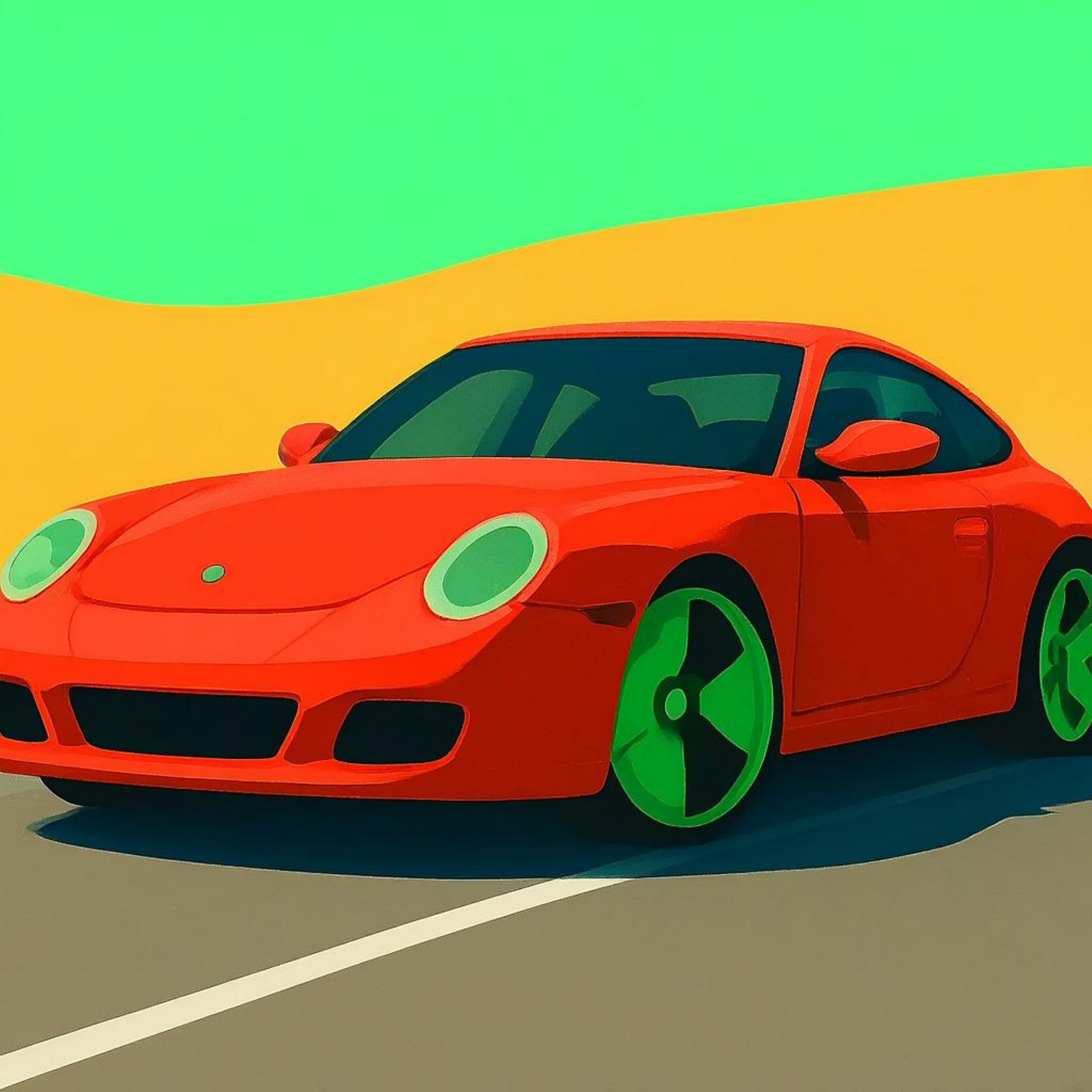} &
\cpimg{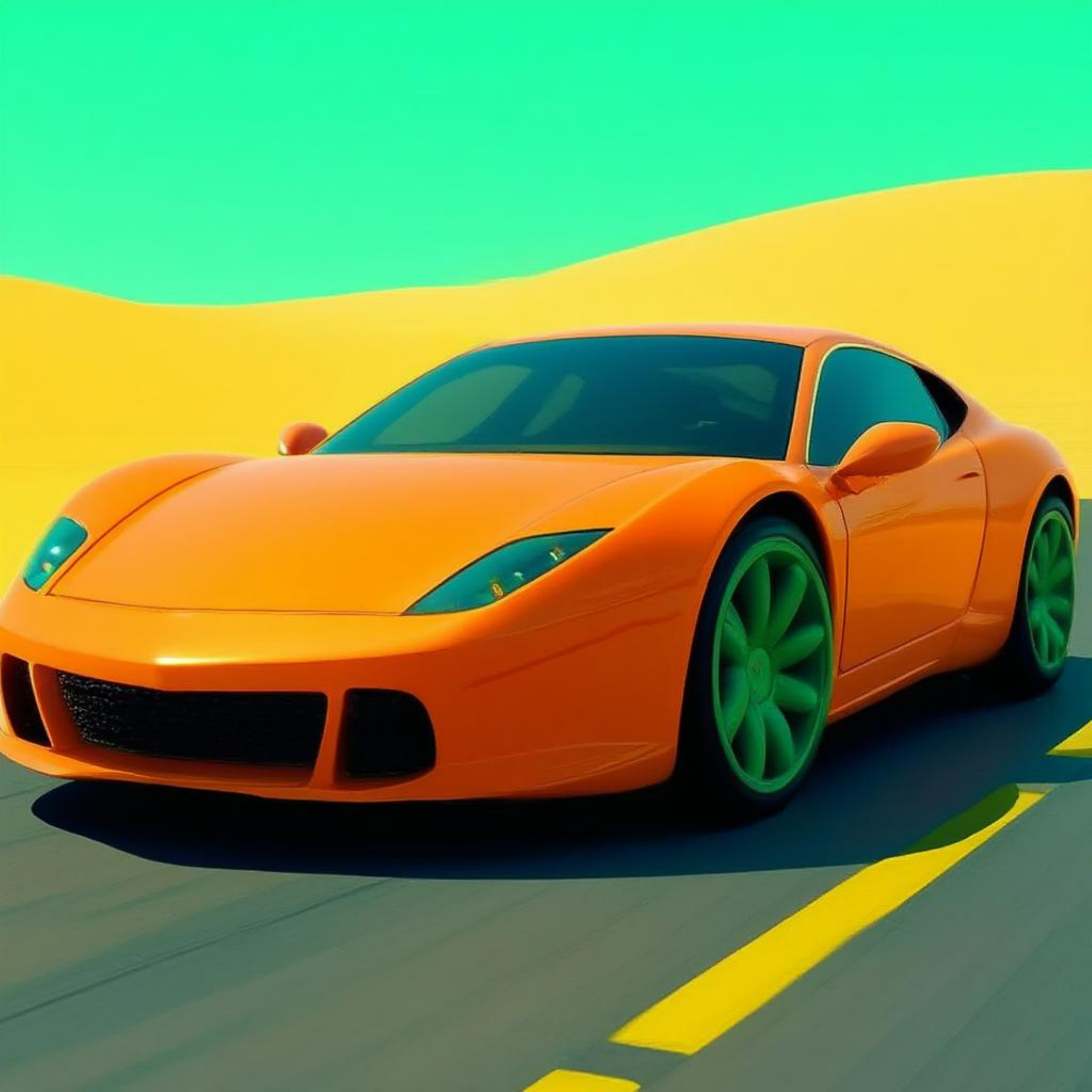} &
\cpimg{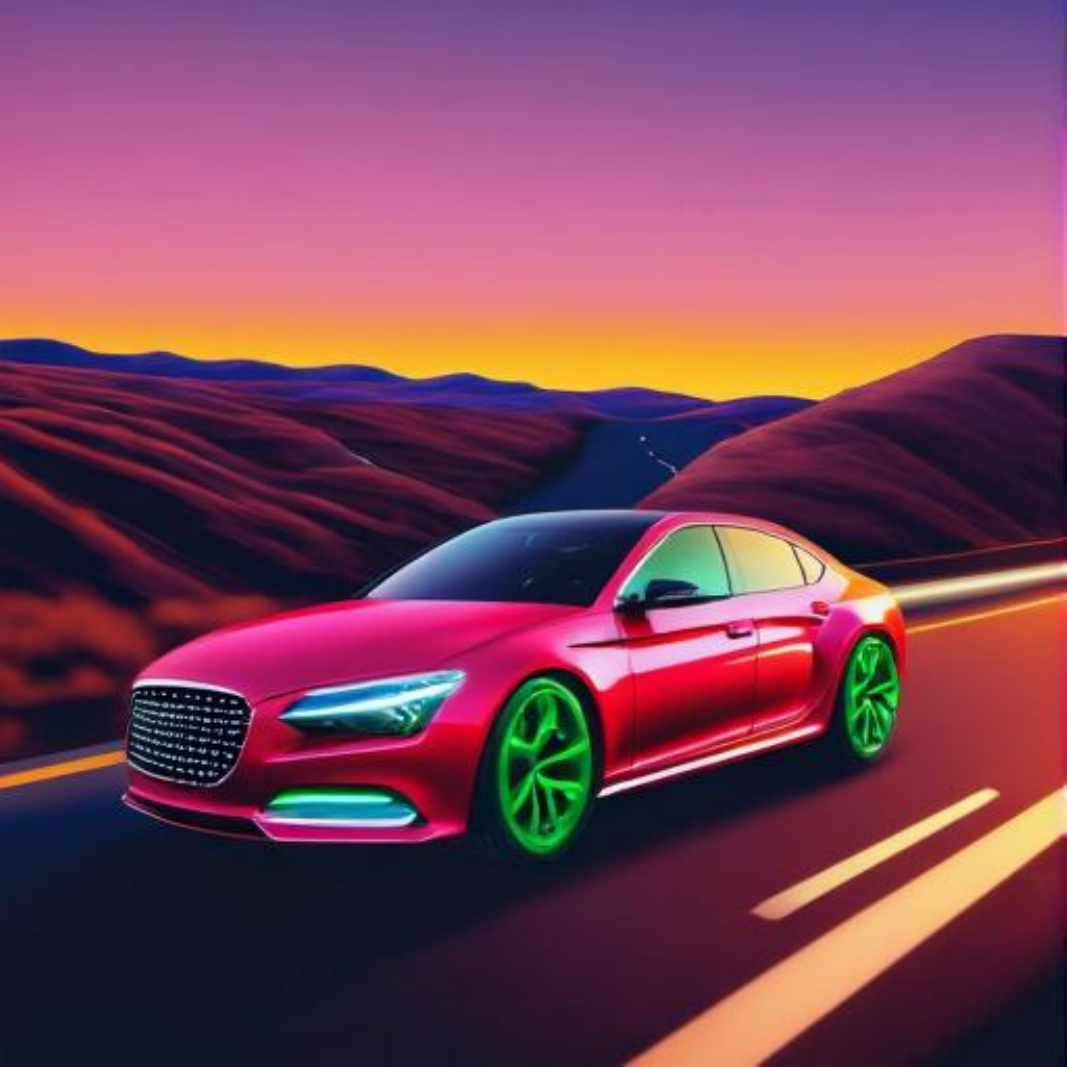} &
\cpimg{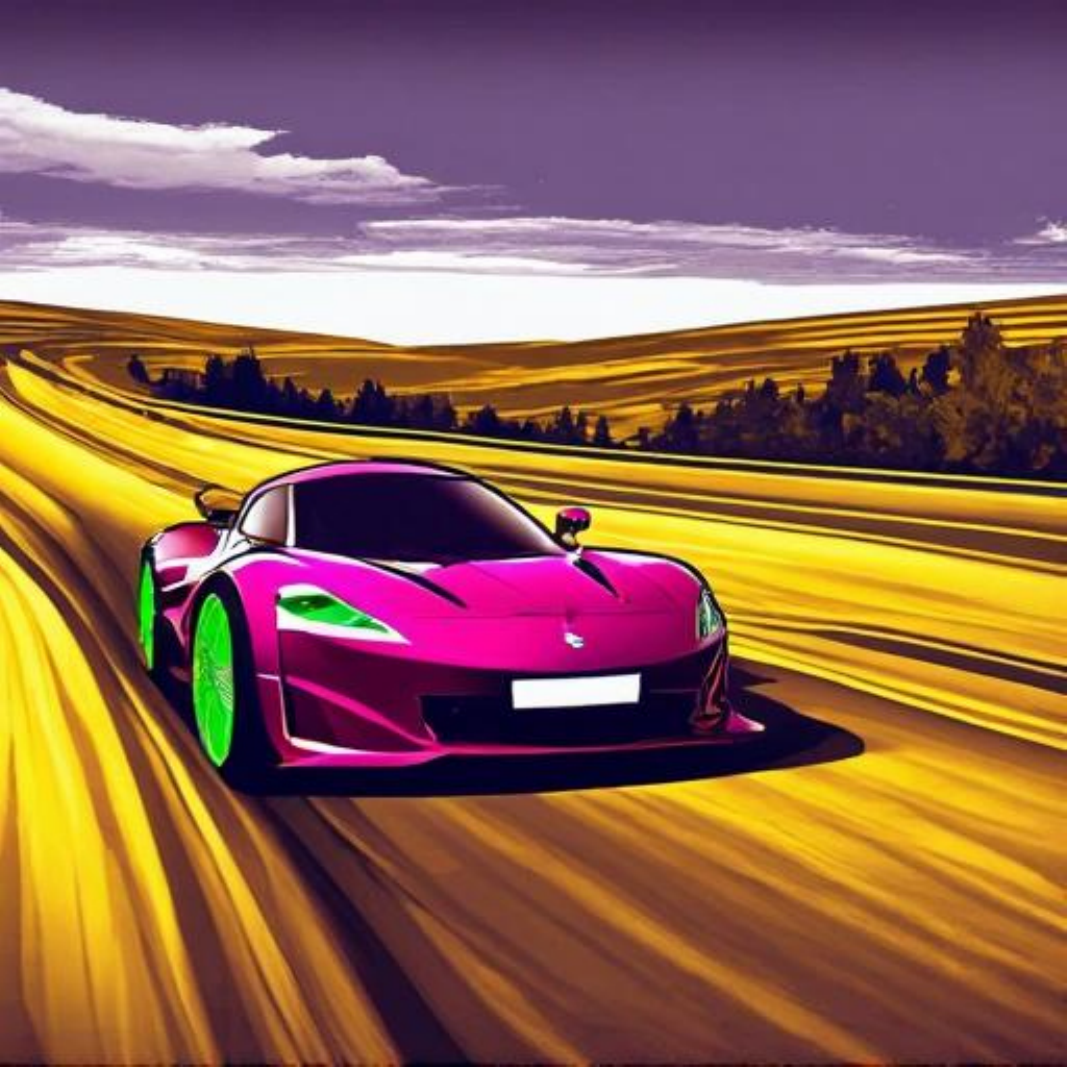} &
\cpimg{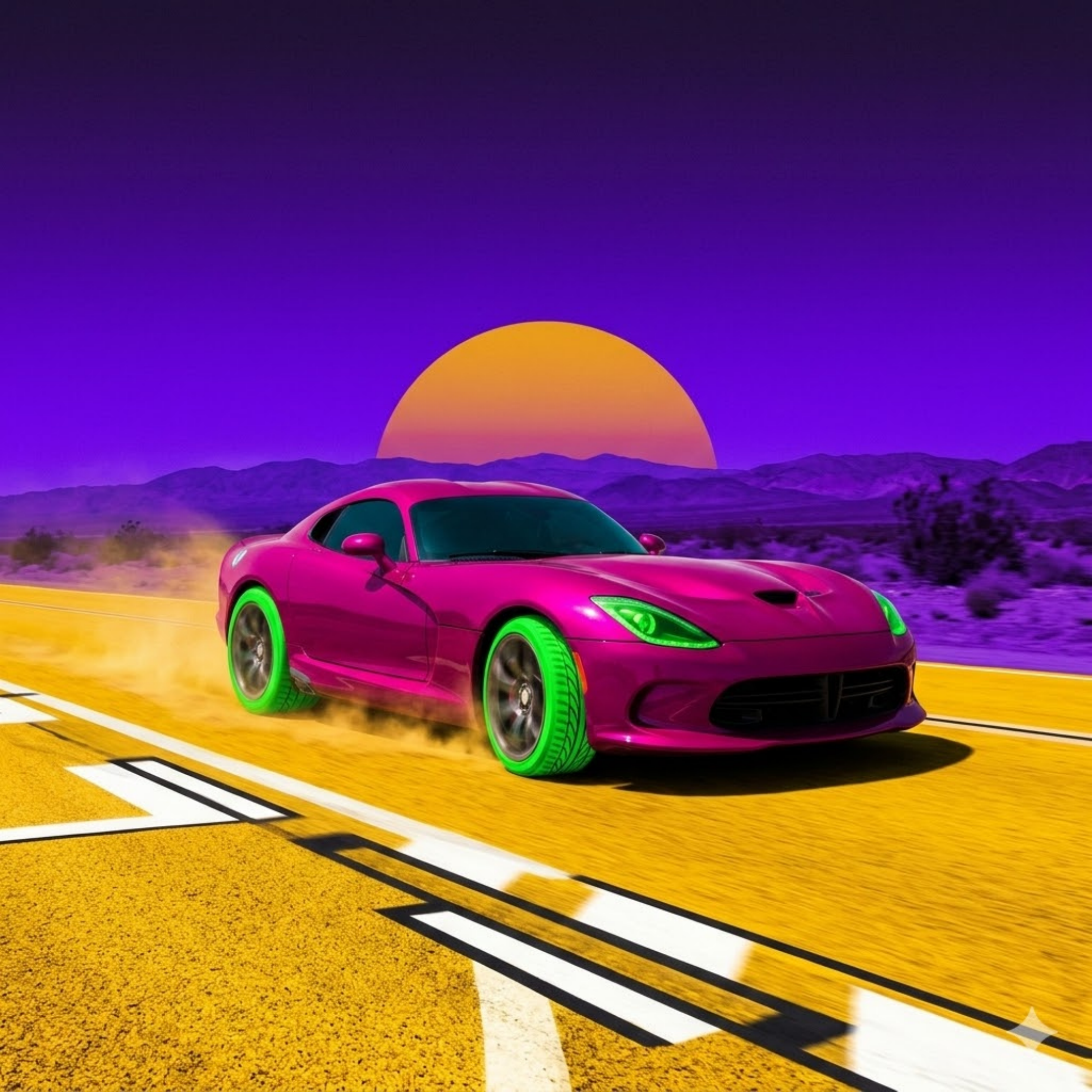} \\[4pt]
\multicolumn{6}{@{}l@{}}{\scriptsize\texttt{--\textcolor[HTML]{71eb34}{sky}:\,\#71eb34;\;\;--\textcolor[HTML]{34b1eb}{flowers}:\,\#34b1eb;\;\;--\textcolor[HTML]{9c34eb}{mountains}:\,\#9c34eb}} \\[2pt]
\cpimg{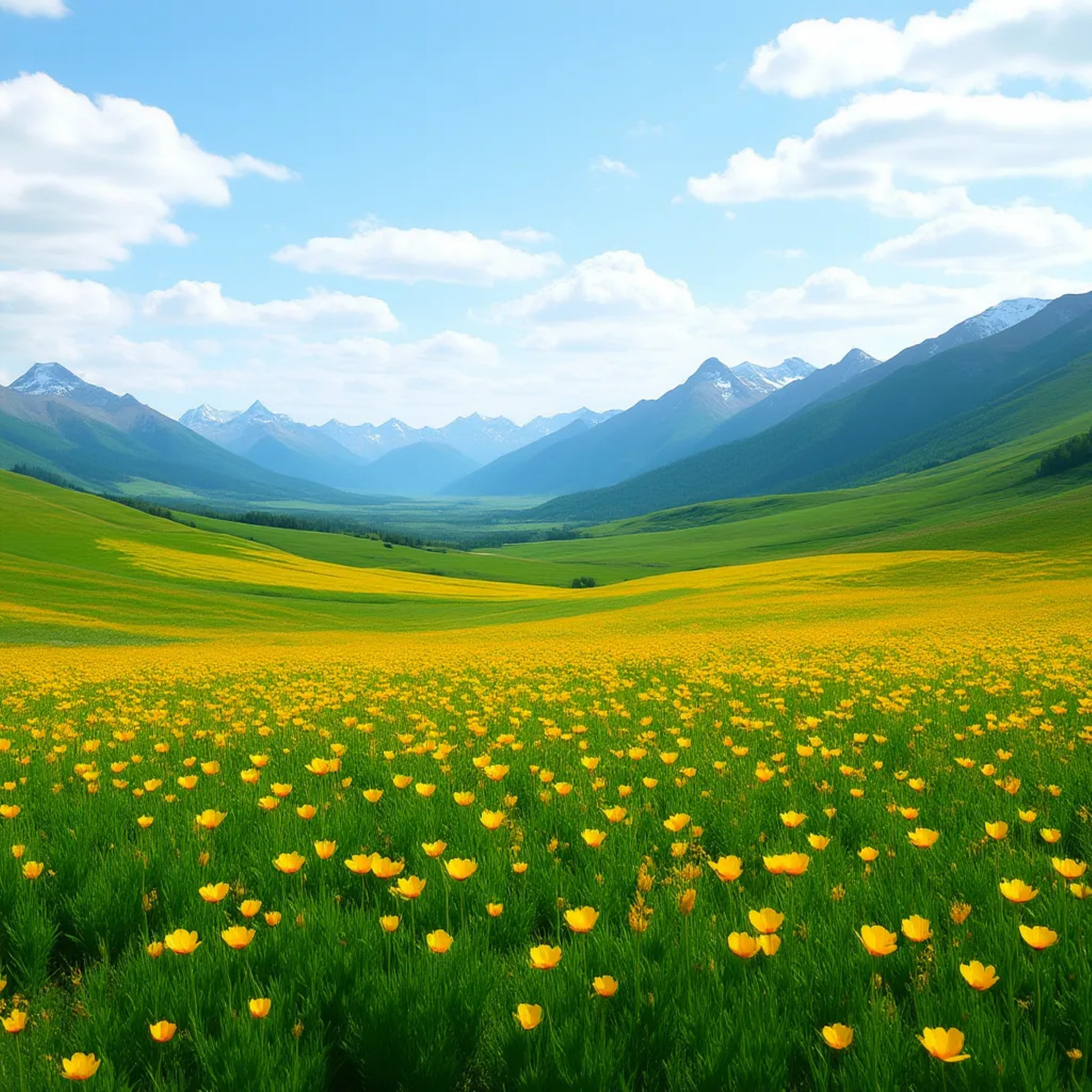} &
\cpimg{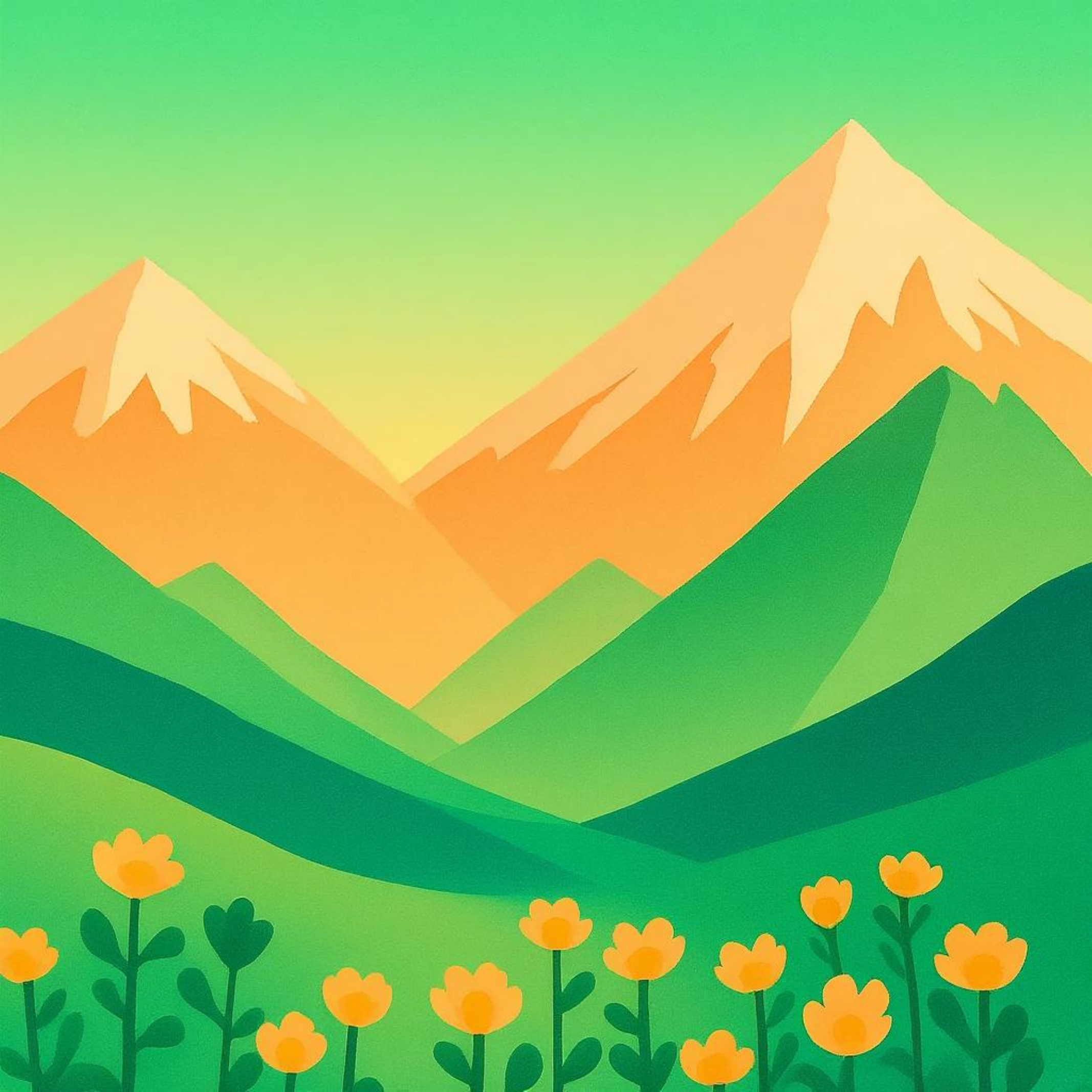} &
\cpimg{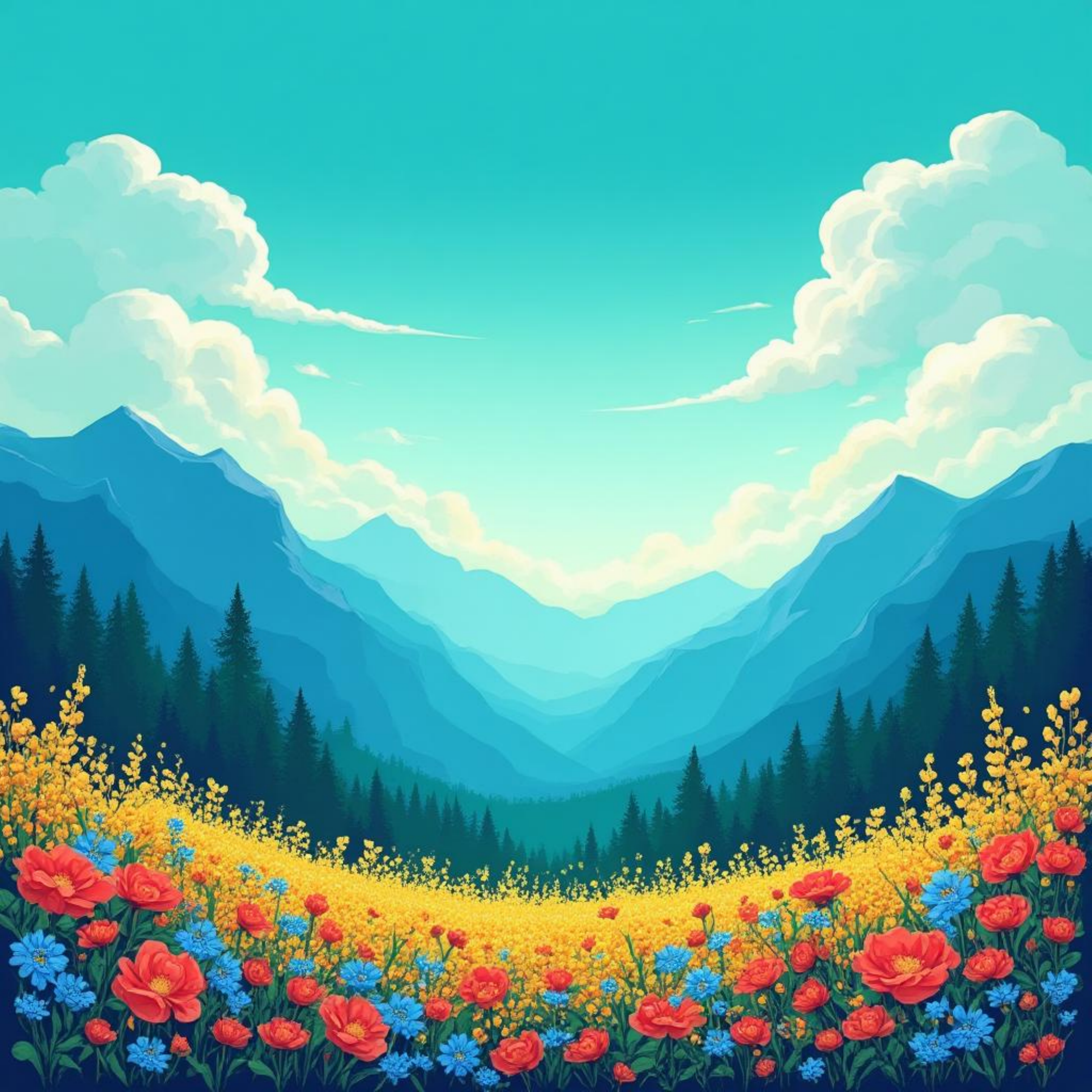} &
\cpimg{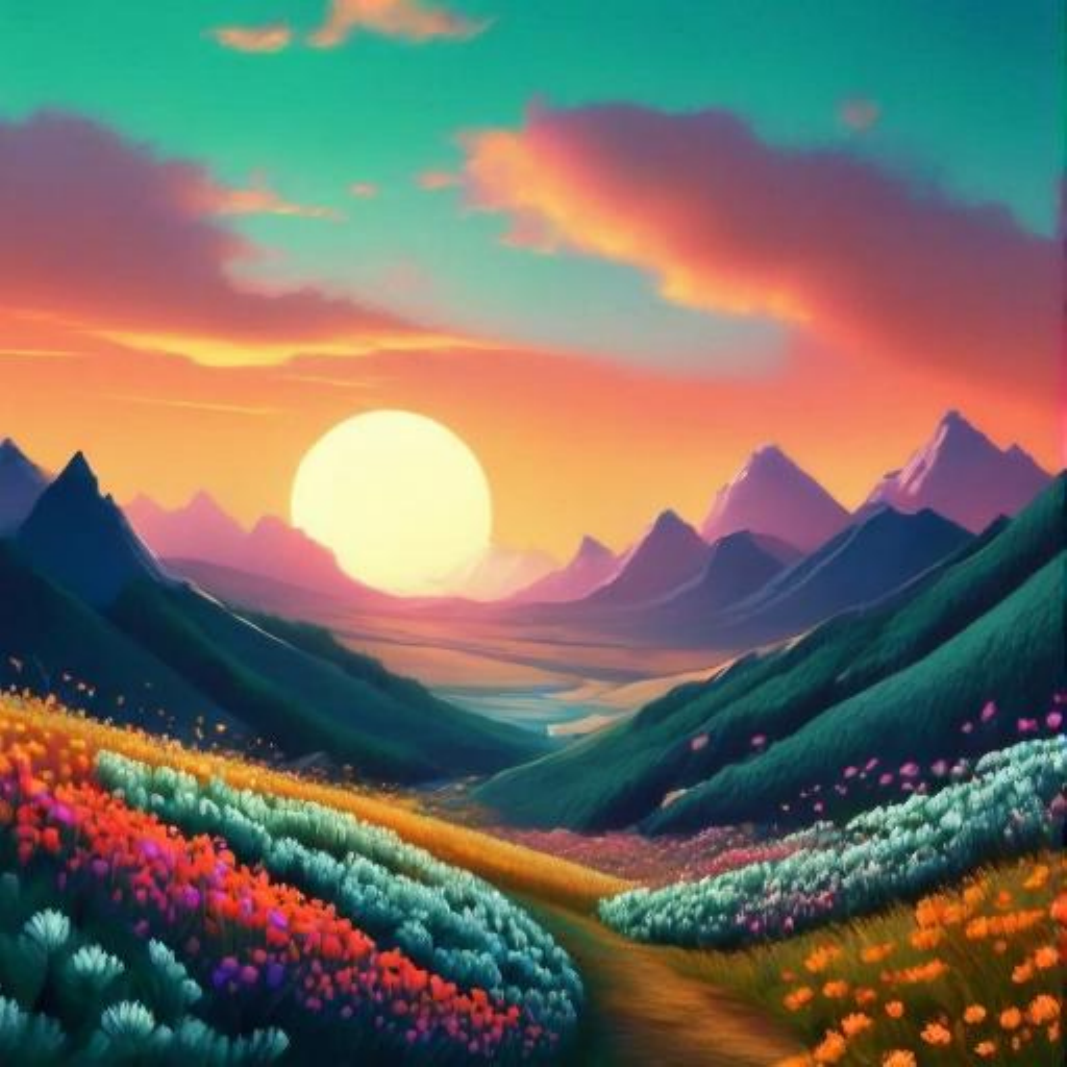} &
\cpimg{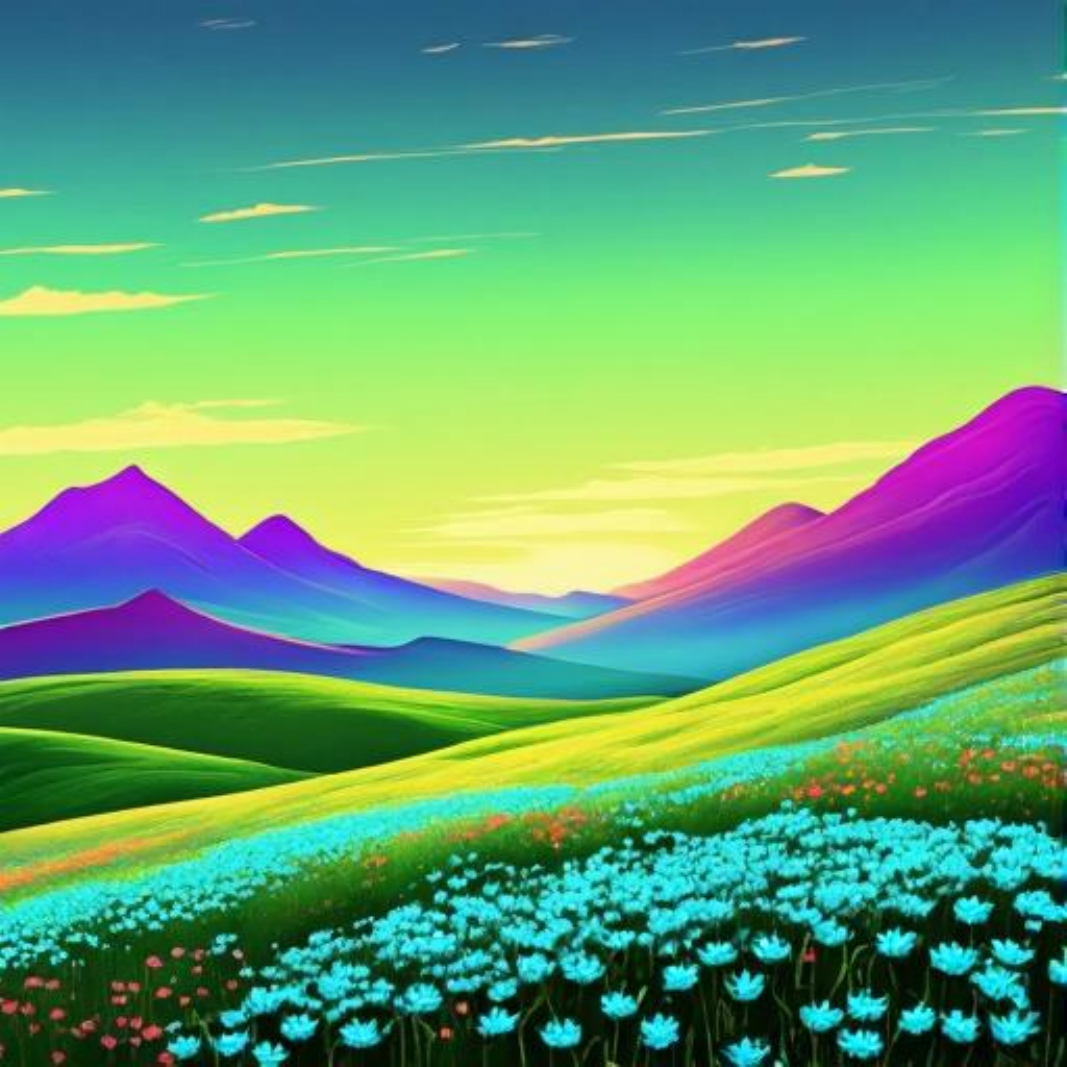} &
\cpimg{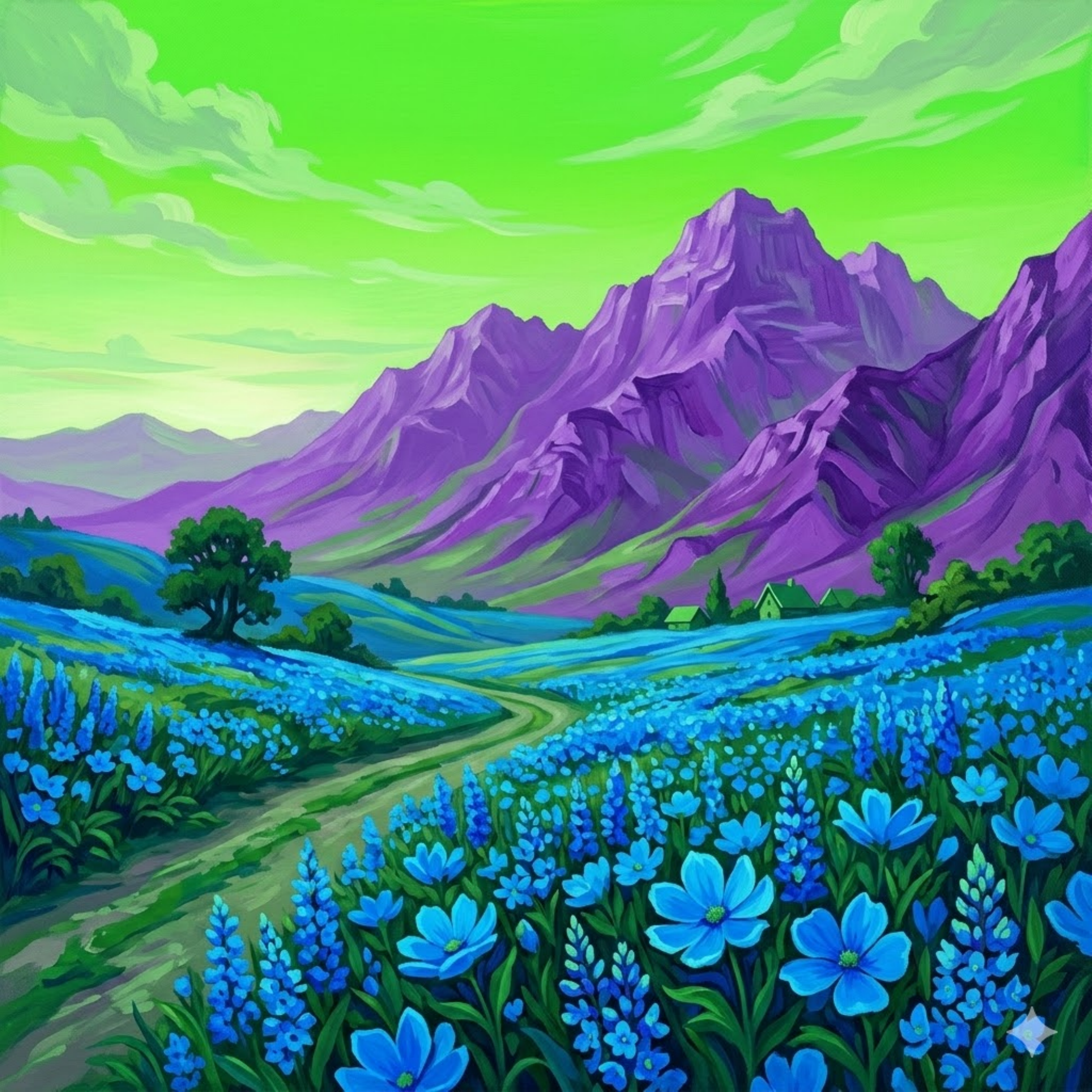} \\
\bottomrule
\end{tabular}
\caption{Color palette guidance: Given hex color codes (shown above each row), we compare how well each model adheres to the specified color specifications. \mural ~better respects the color constraints compared to OmniGen2 and Bagel.}
\label{fig:colorpalette}
\end{figure}

\noindent\textbf{Color palette guidance.}
We next prompt the model to generate images with specific freeform instruction on color composition in the generated images.
For instance, for the following prompt
{\footnotesize
\begin{verbatim}
Generate a car on a road using this color palette:
--car-body: #FF1493;
--tires: #00FF00;
--road: #FFD700;
--sky: #8B00FF;
\end{verbatim}}
\noindent
we expect the model to generate \textcolor[HTML]{FF1493}{car body}, \textcolor[HTML]{00FF00}{tires}, \textcolor[HTML]{FFD700}{road}, and \textcolor[HTML]{8B00FF}{sky} adhering to the specified color palette.
This task requires the LLM to interpret the structured color specification and communicate it to the generation branch through attention. We show qualitative results in Figure~\ref{fig:colorpalette}.
Compared to unified models such as OmniGen2 and Bagel, \mural~better follows the instruction on color guidance and the results improve with CoT at inference.
\mural~can achieve comparable results to Nano Banana 2, currently one of the top image generation models.

\noindent\textbf{Emoji and ASCII scene composition.}
We test the model's ability to interpret emoji sequences and ASCII art as scene descriptions.
For instance, when we prompt the model: ``Convert this ASCII art to a realistic image: <mountain emoji>, <tree emoji>, <deer emoji>'' (row 1 in Figure~\ref{fig:emoji_scene}) \mural~generates a mountain landscape with trees and a deer.
Again, \mural~understands these prompts via shared attention even \textit{without} CoT/reasoning at inference. We also show results with CoT enabled.
This is another demonstration of the image generation expert effectively leveraging LLM's semantic understanding of emoji and symbolic representations.
We show more qualitative examples with different number of emojis.
In row 2, \mural~correctly infers the color of the car from the emoji, and in row 4, it respects the spatial layout implied by the sequence -- generating waves on the left, beach in the center, and vegetation on the right.
Note that OmniGen2 and Bagel also use the same Qwen2.5 tokenizer which suggests that tokenizer support alone is not sufficient.
Among the models compared, Nano Banana 2 also consistently does well on these examples.

\newcommand{\emojiimg}[1]{\includegraphics[width=0.135\linewidth]{#1}}
\begin{figure*}[!t]
\centering
\setlength{\tabcolsep}{1.5pt}
\scriptsize
\begin{tabular}{@{}l@{\hskip 4pt}cccccc@{}}
\toprule
Prompt & OmniGen2 & Bagel & Bagel+CoT & \mural & \mural+CoT & NanoBanana2 \\
\midrule
\raisebox{0.3cm}{\includegraphics[width=1.2cm]{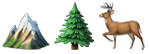}} &
\emojiimg{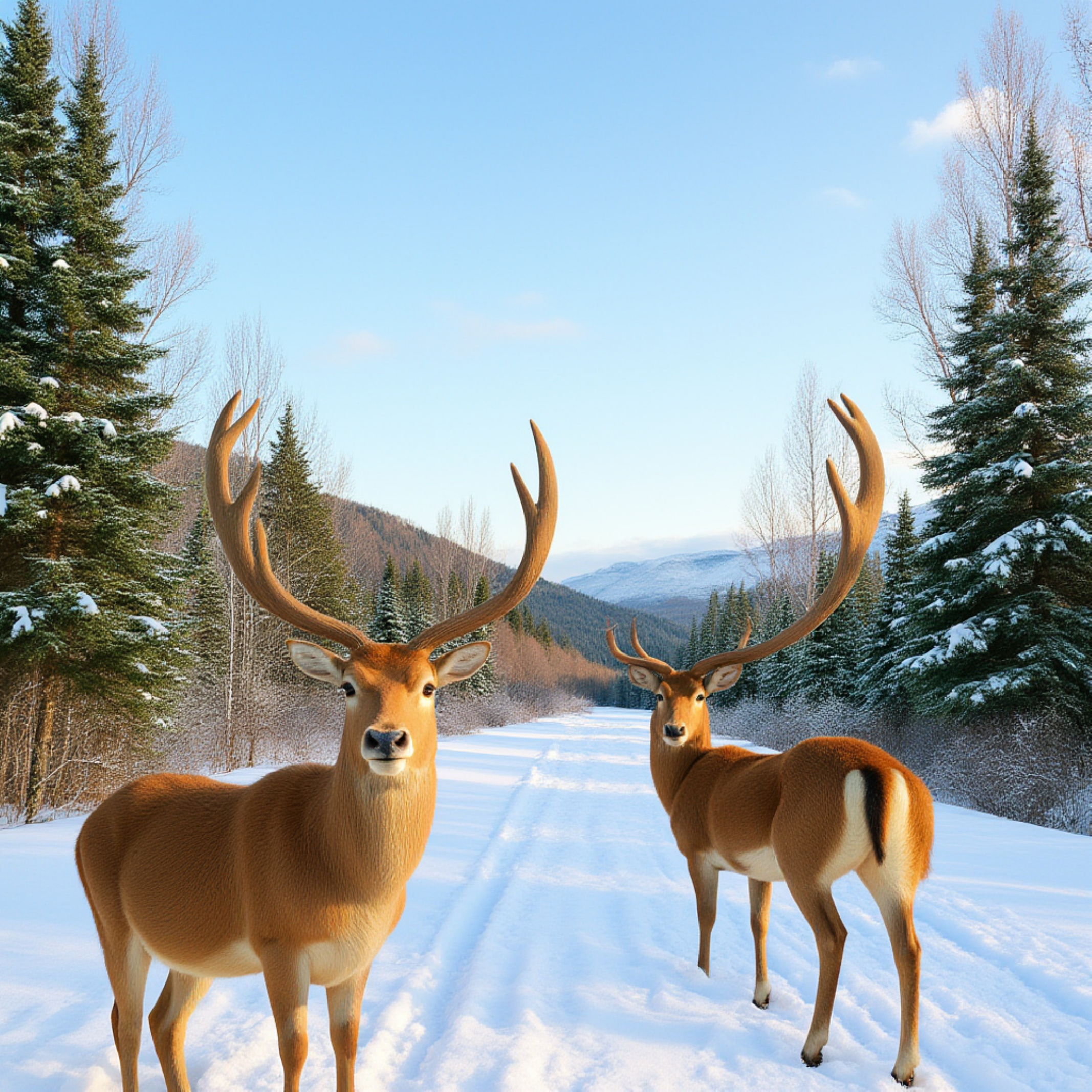} &
\emojiimg{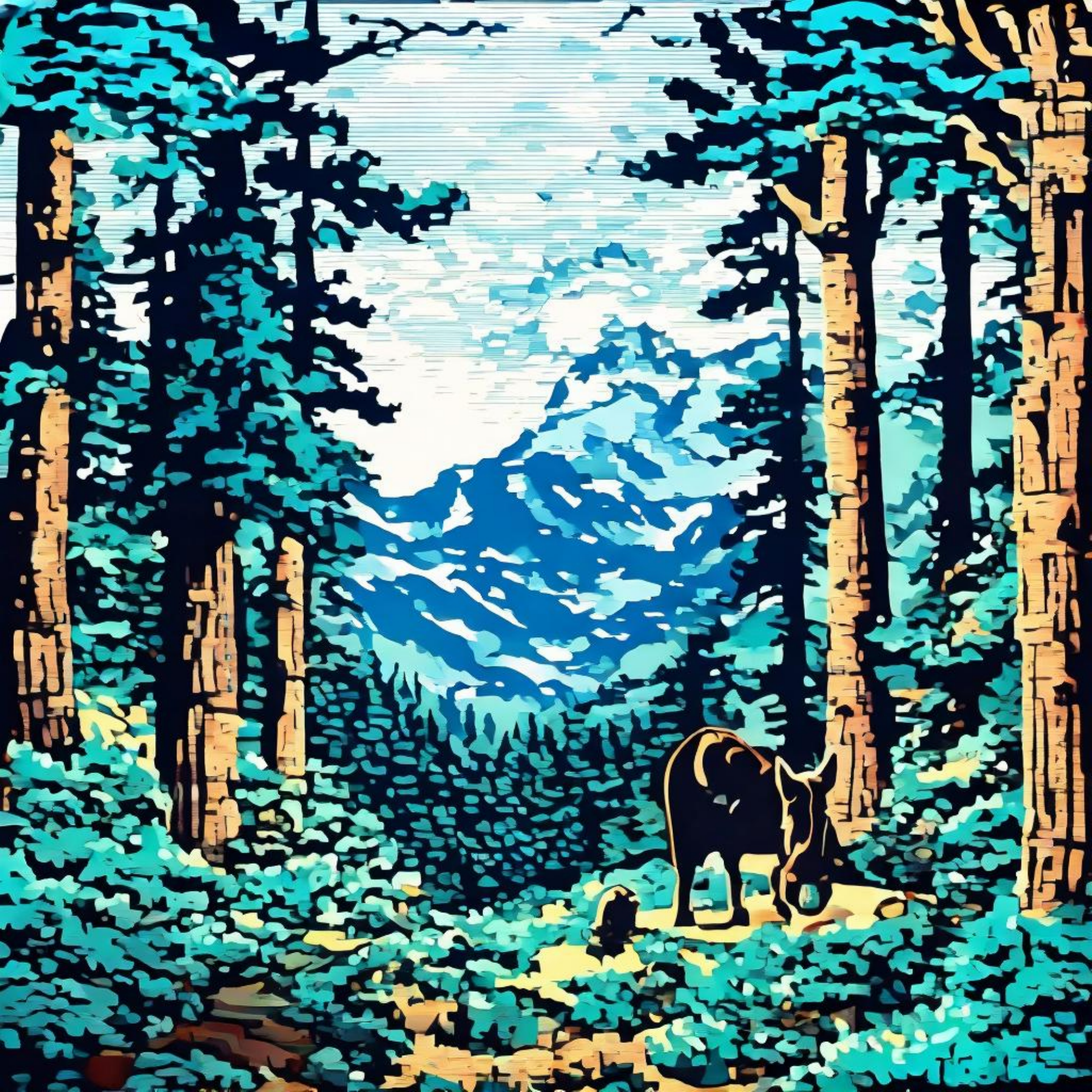} &
\emojiimg{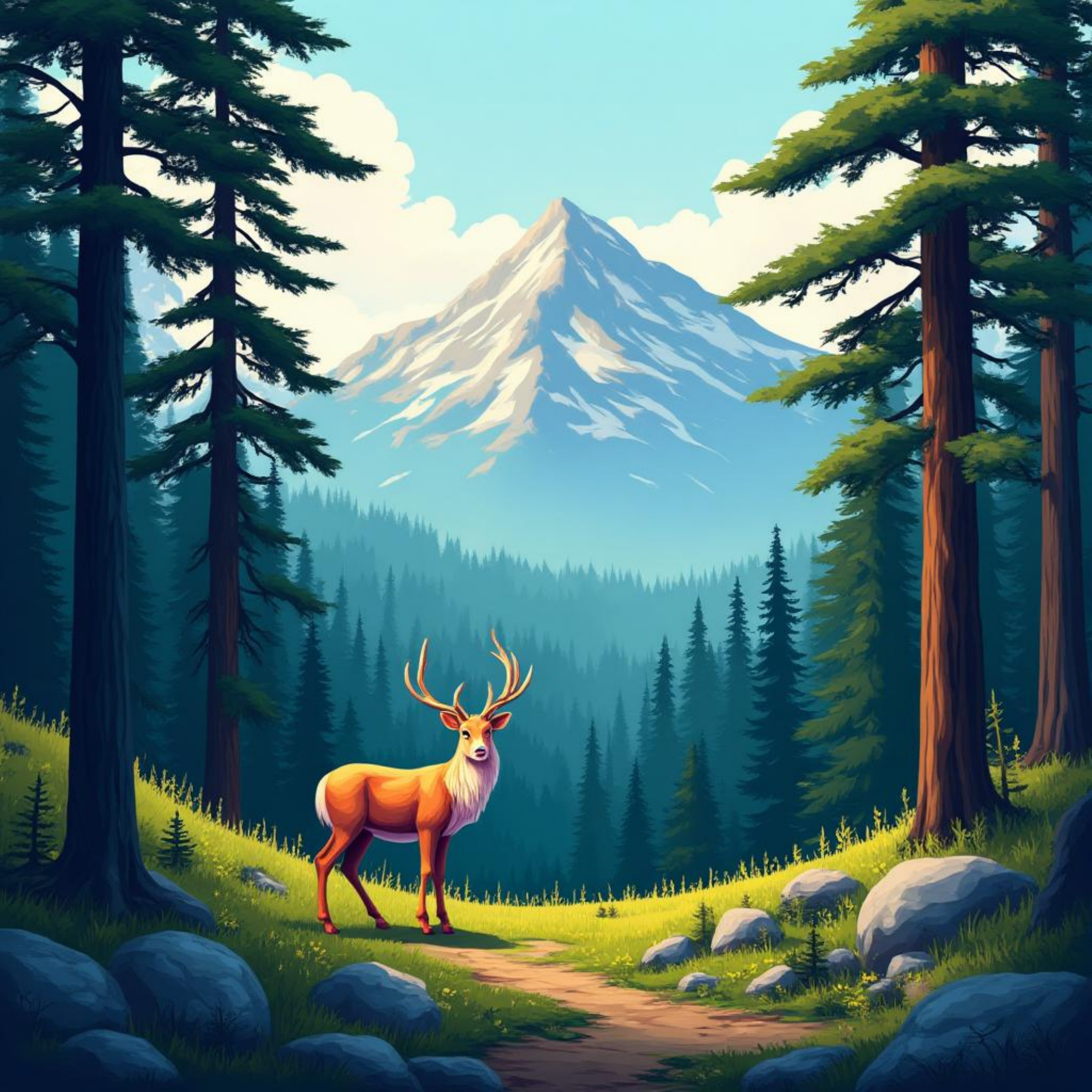} &
\emojiimg{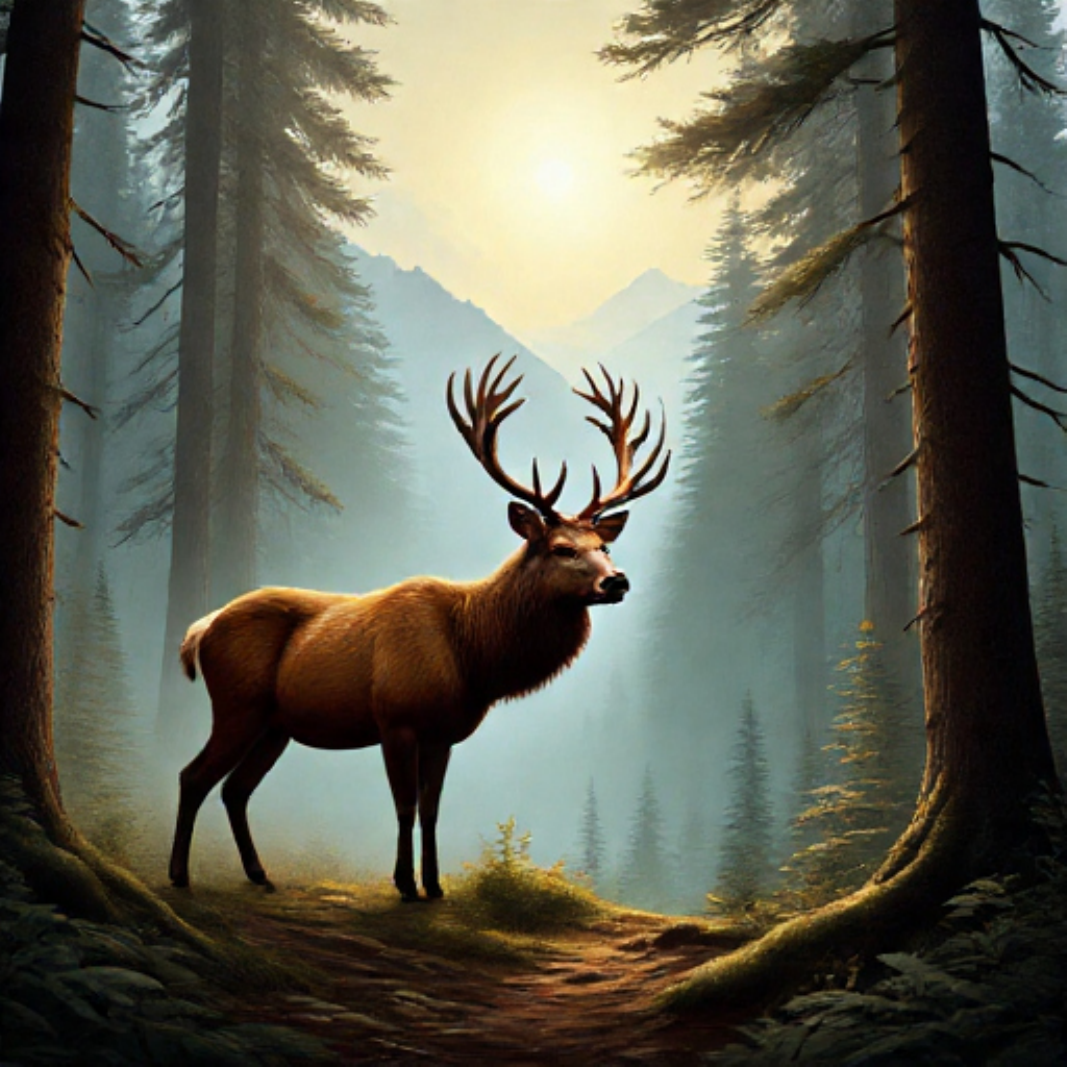} &
\emojiimg{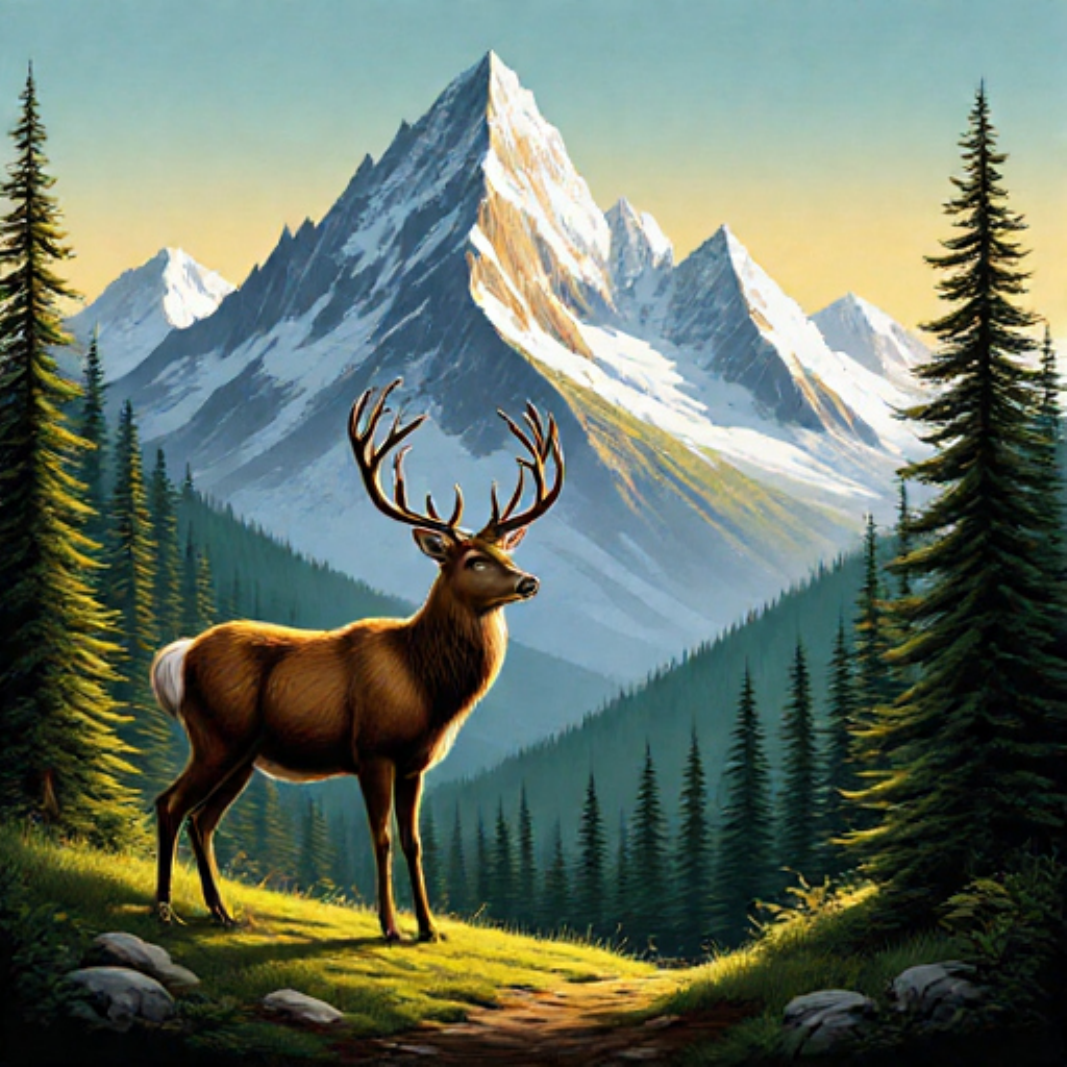} &
\emojiimg{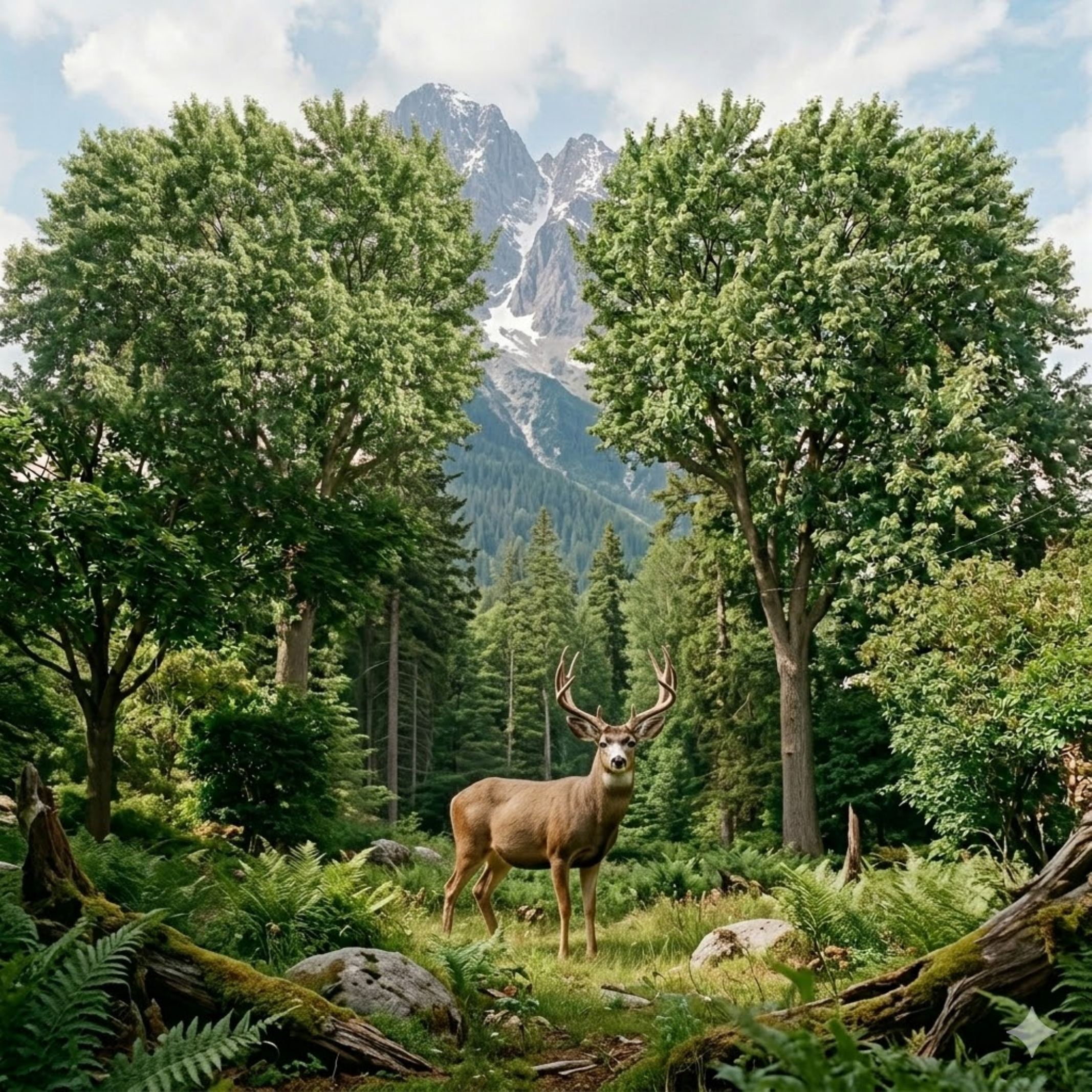} \\[2pt]
\raisebox{0.3cm}{\includegraphics[width=1.2cm]{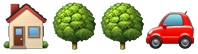}} &
\emojiimg{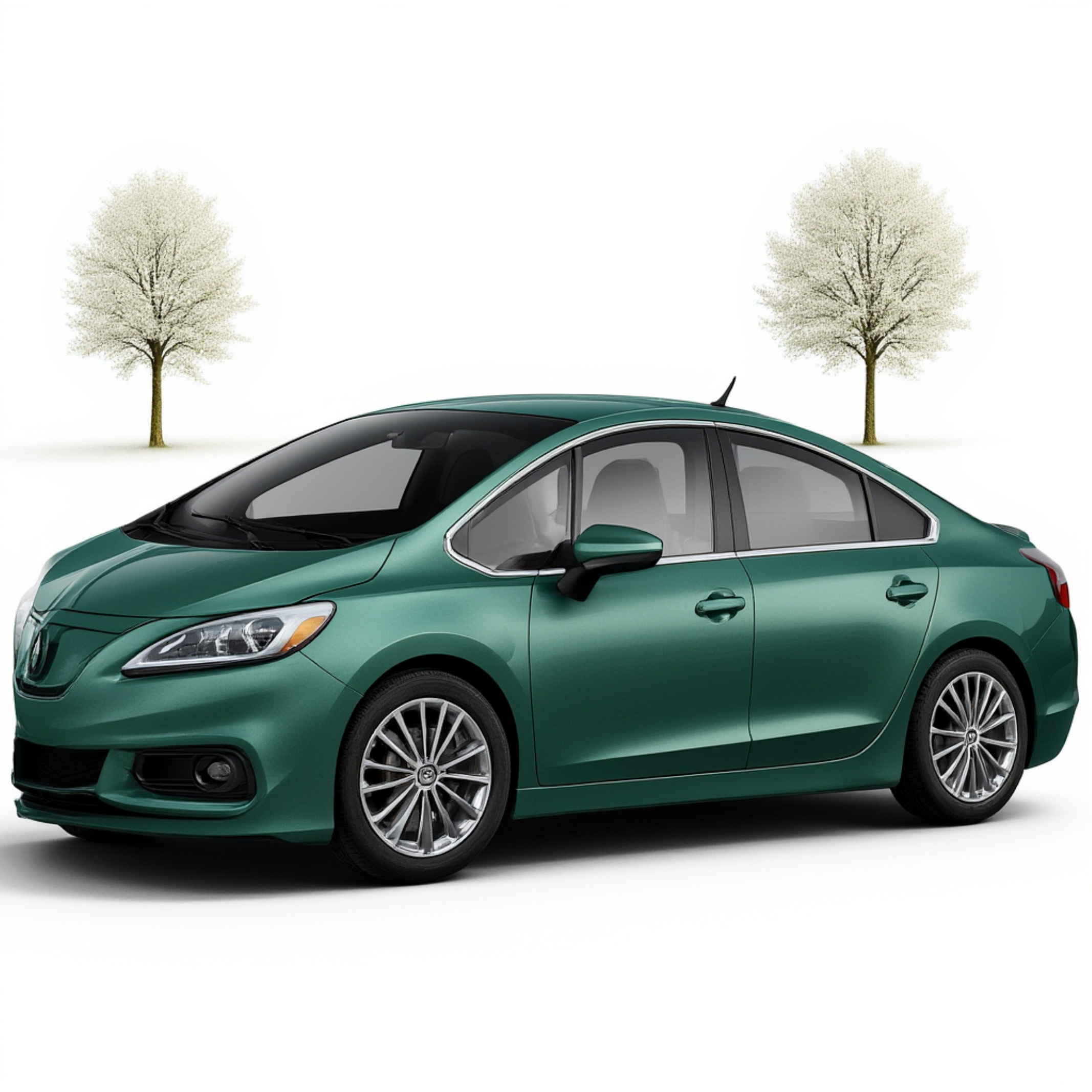} &
\emojiimg{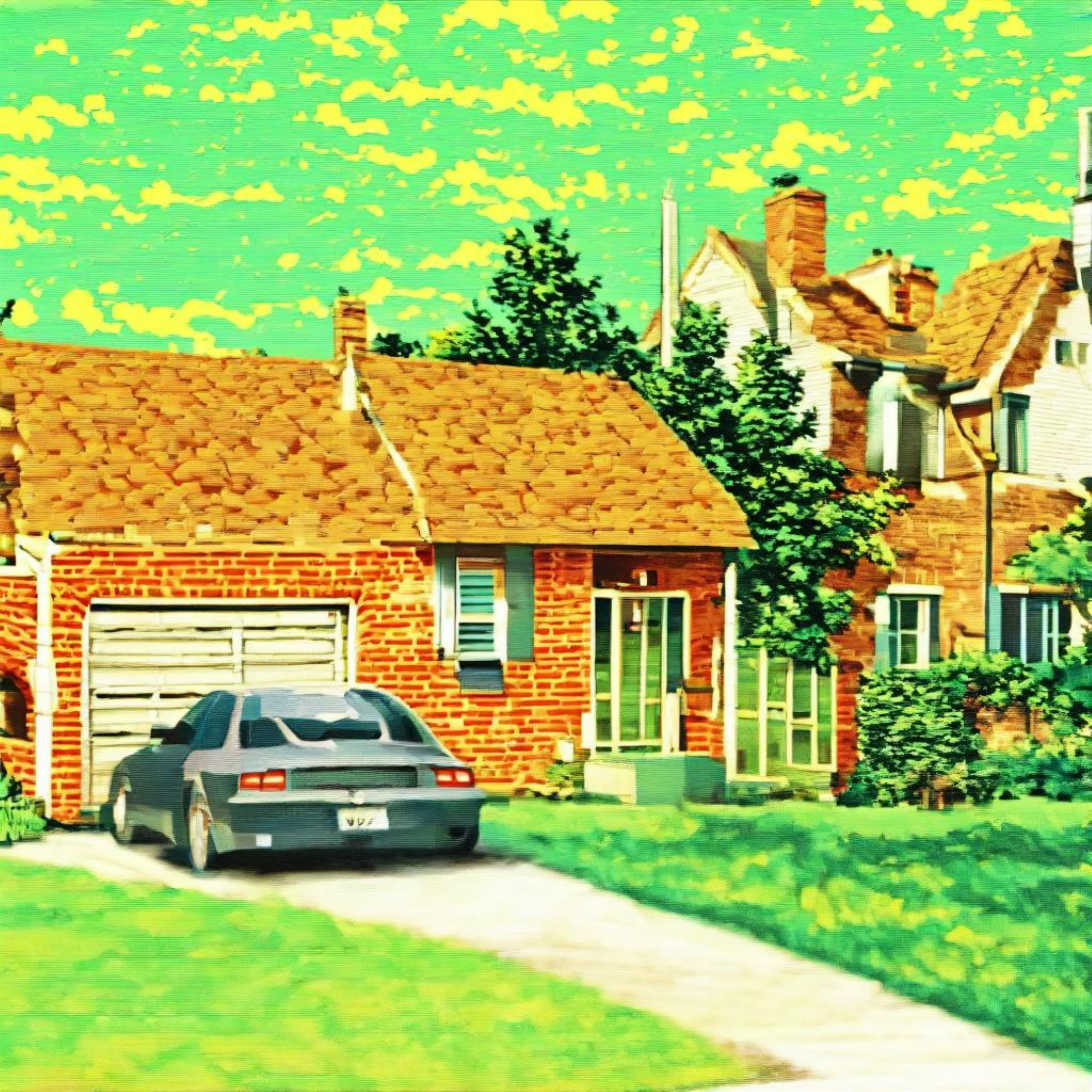} &
\emojiimg{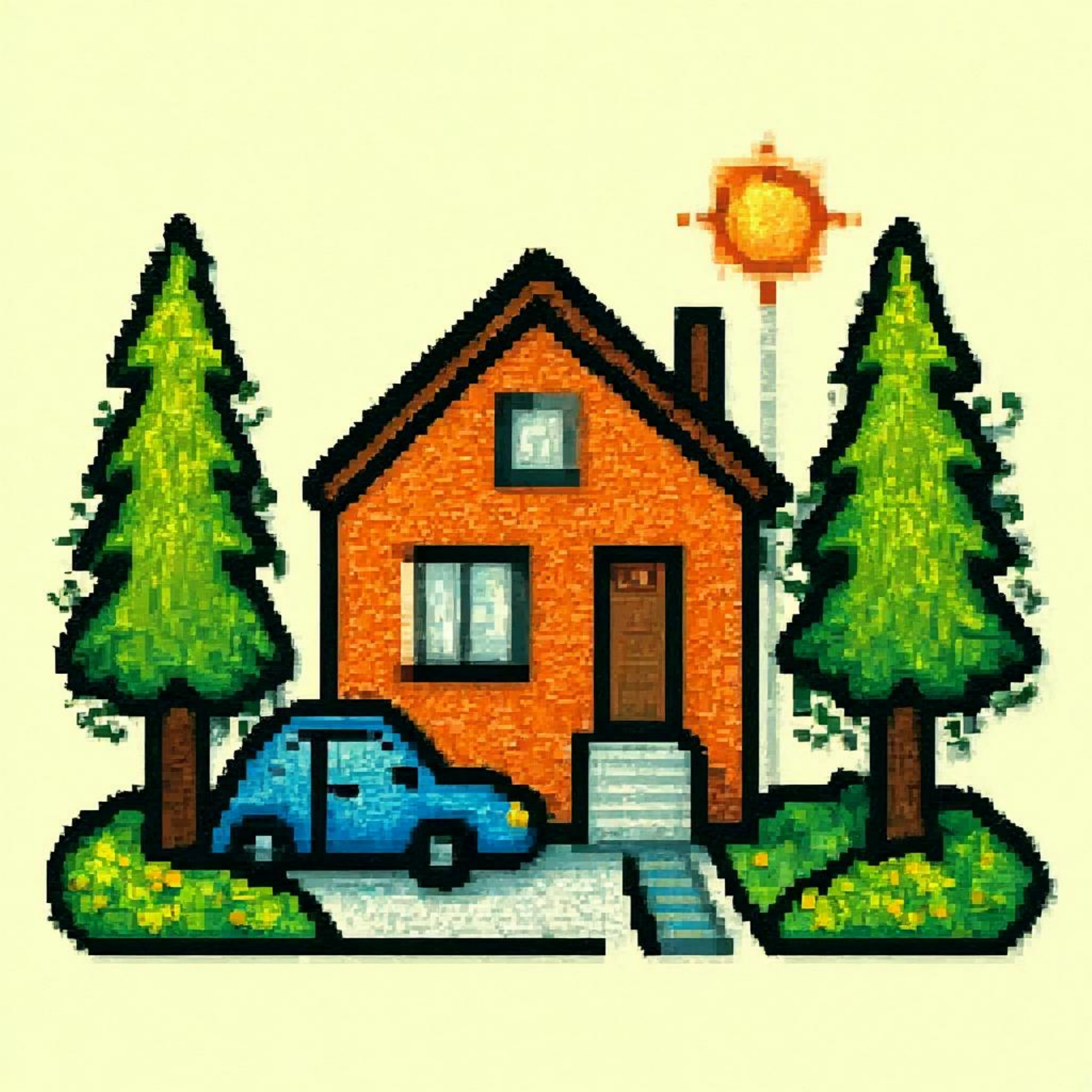} &
\emojiimg{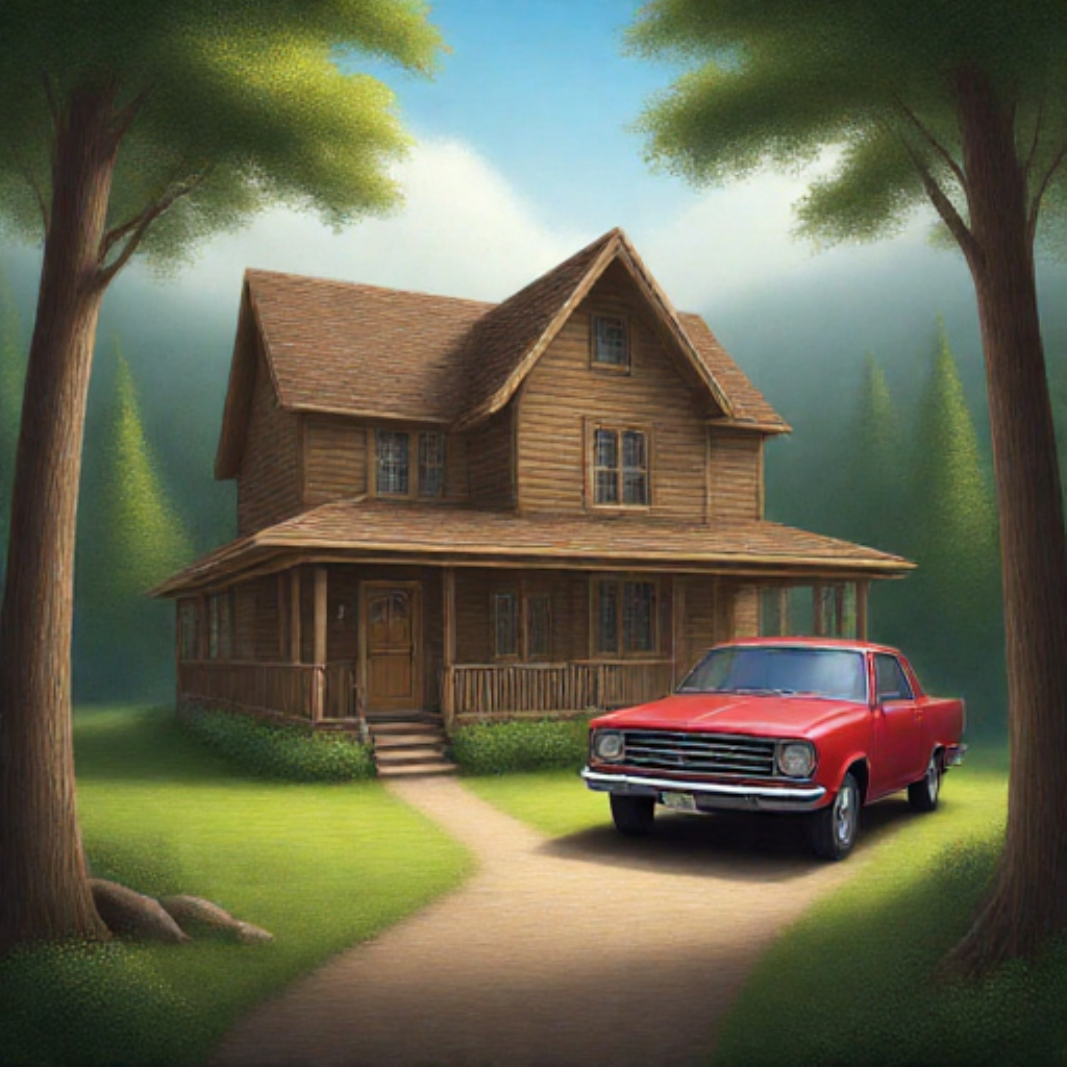} &
\emojiimg{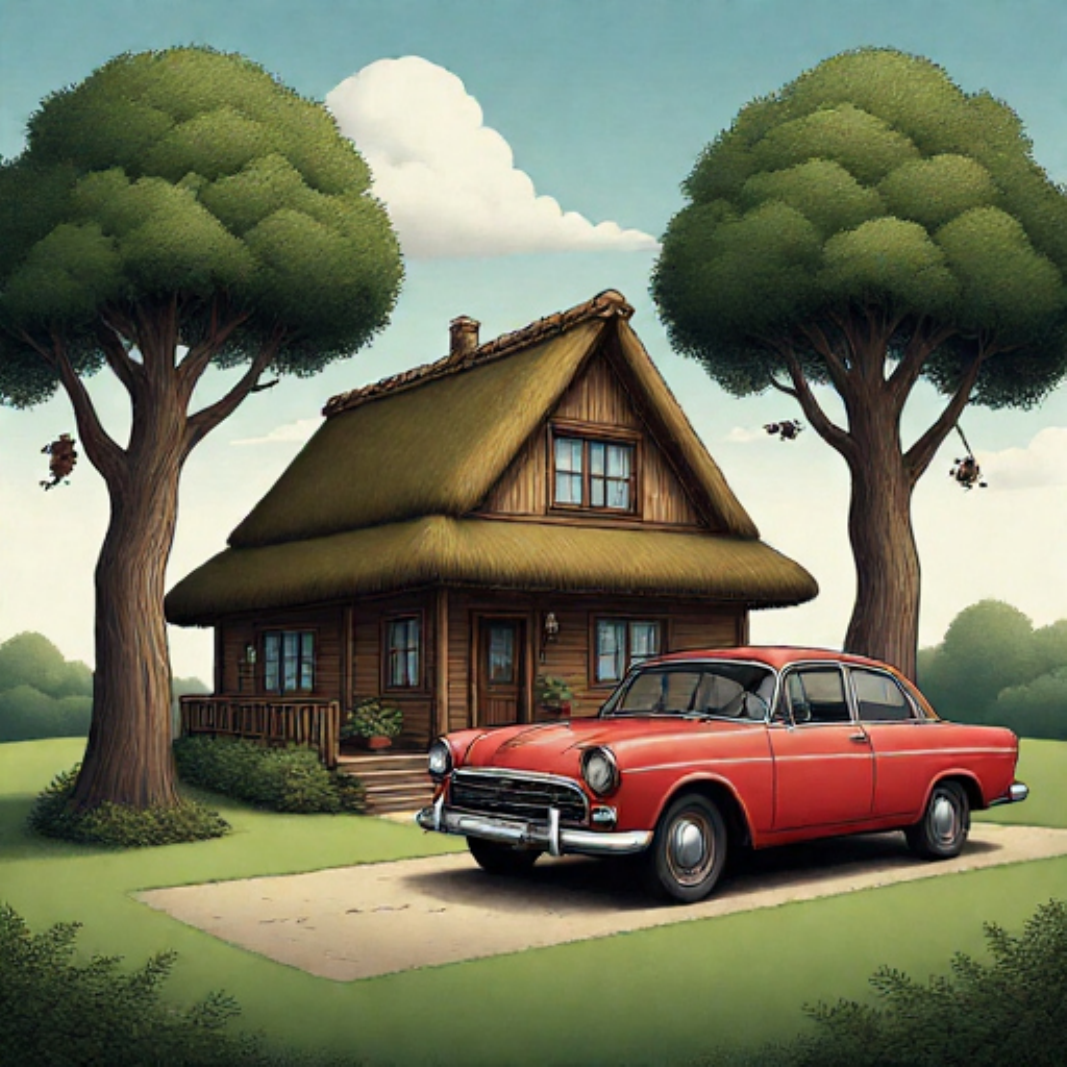} &
\emojiimg{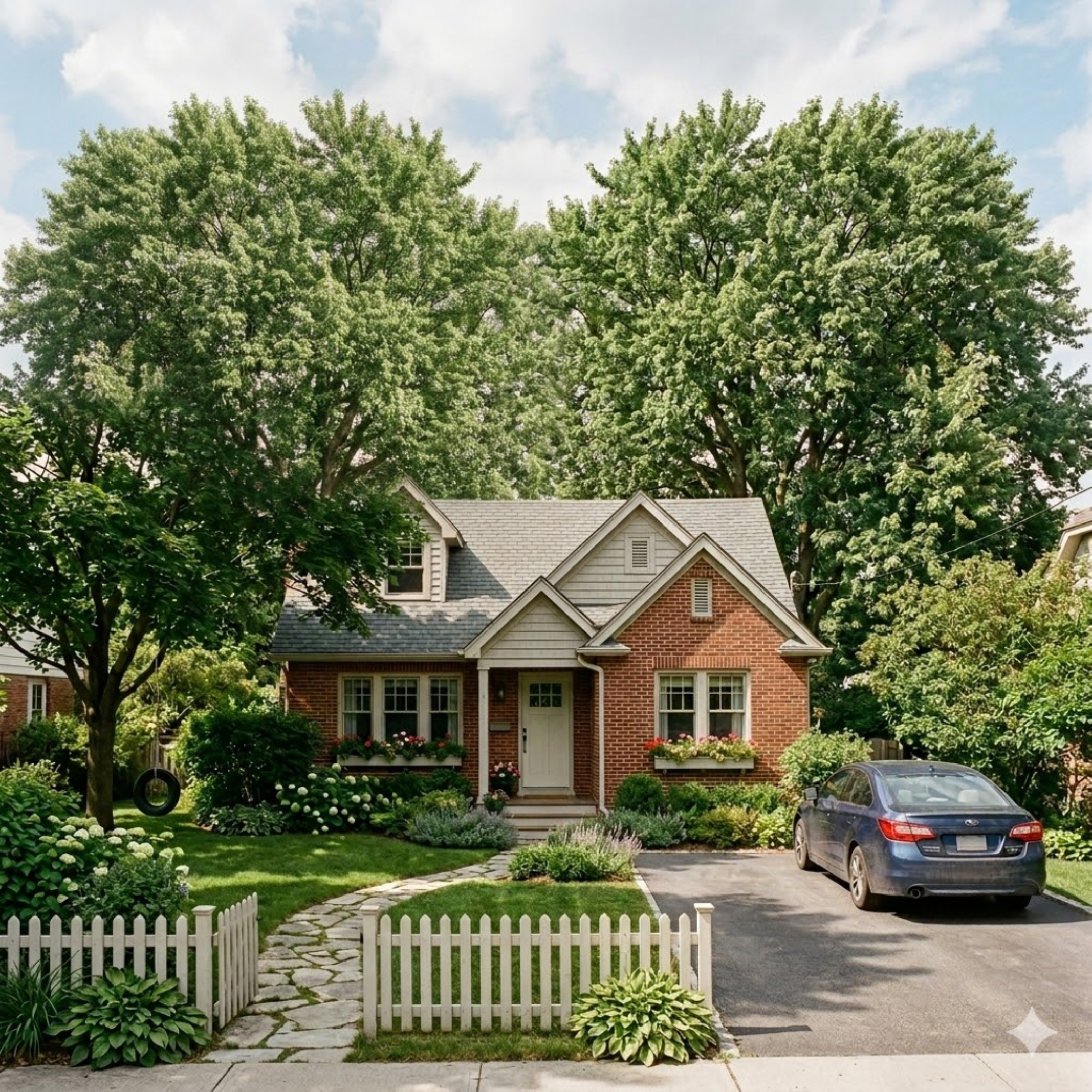} \\[2pt]
\raisebox{0.3cm}{\includegraphics[width=1.2cm]{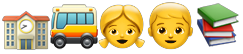}} &
\emojiimg{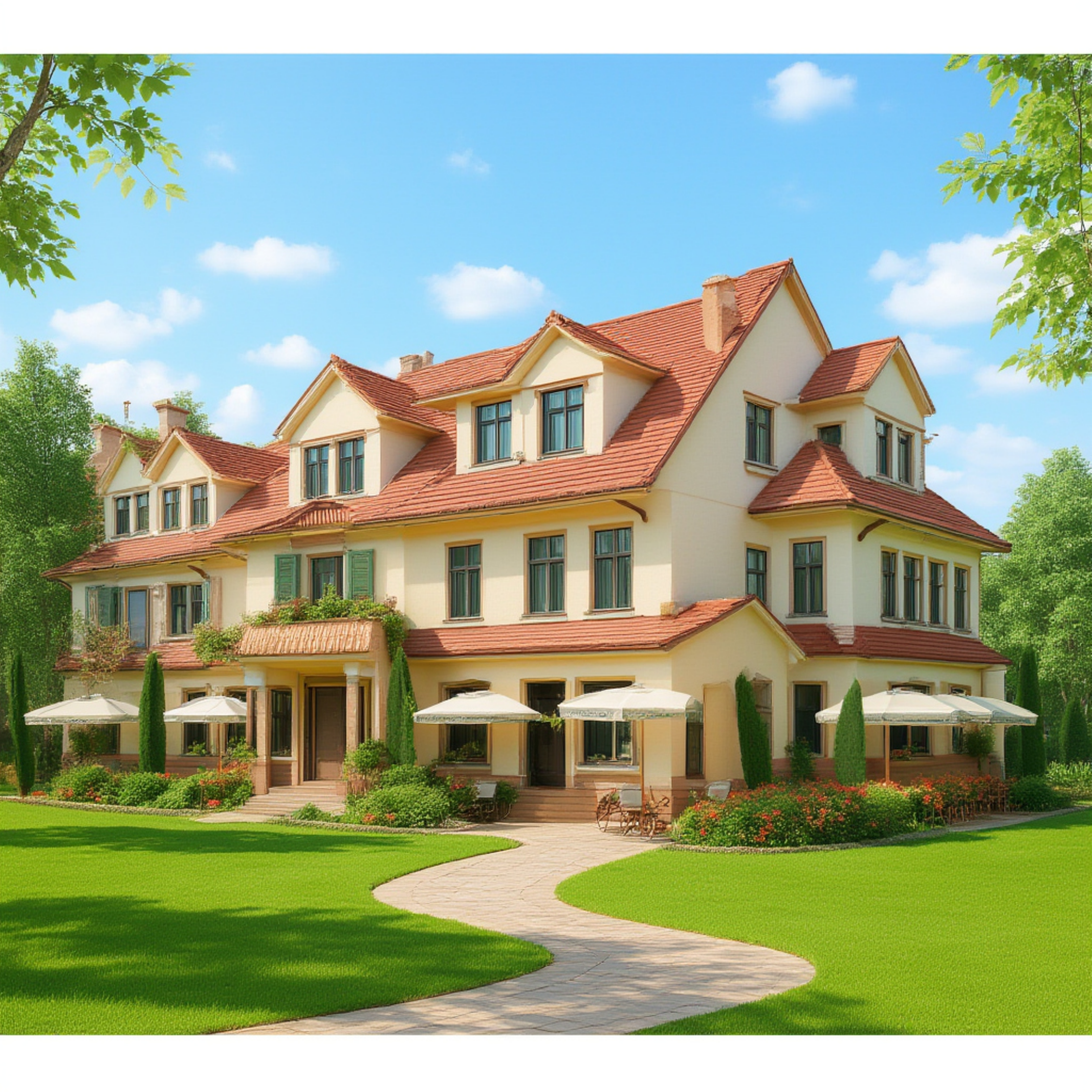} &
\emojiimg{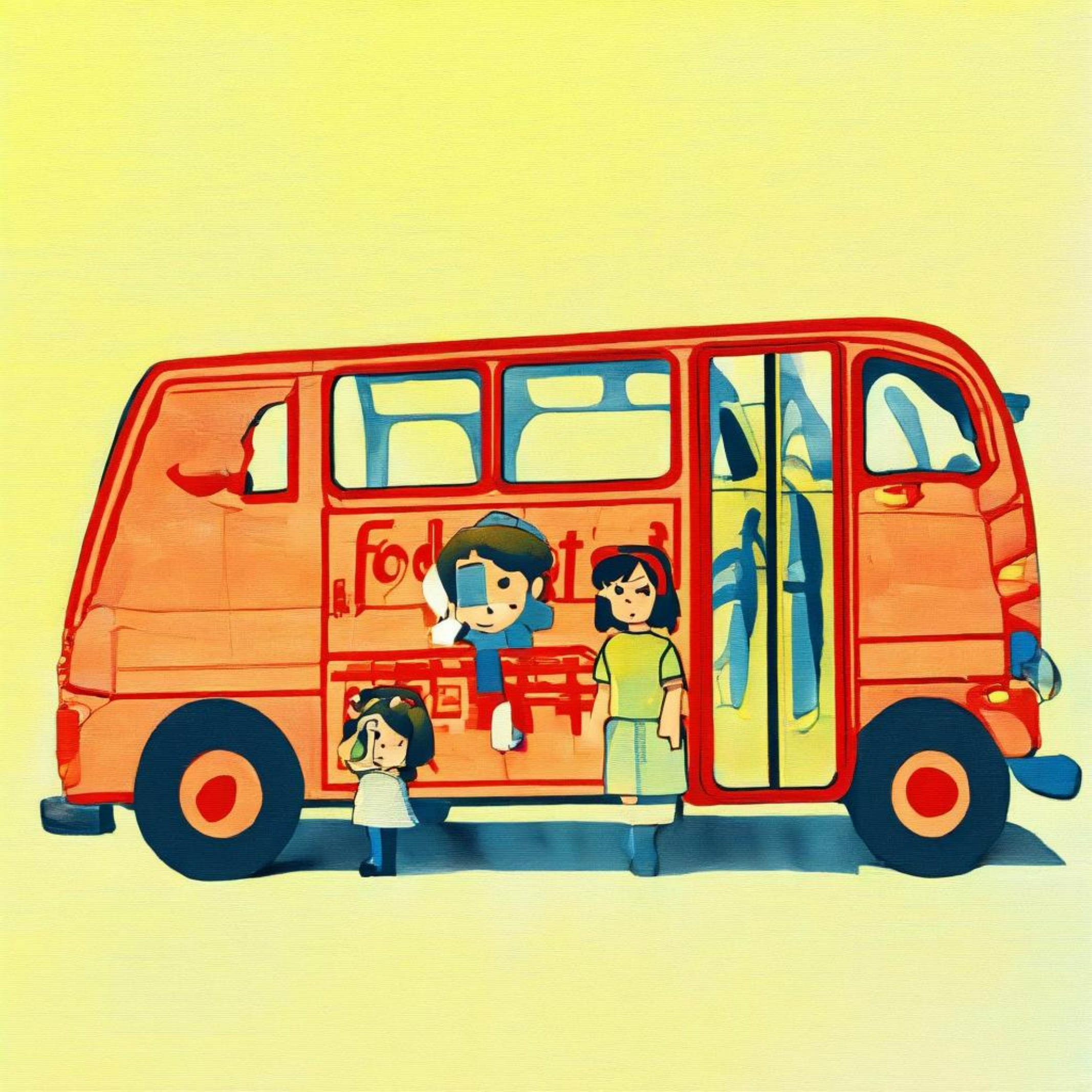} &
\emojiimg{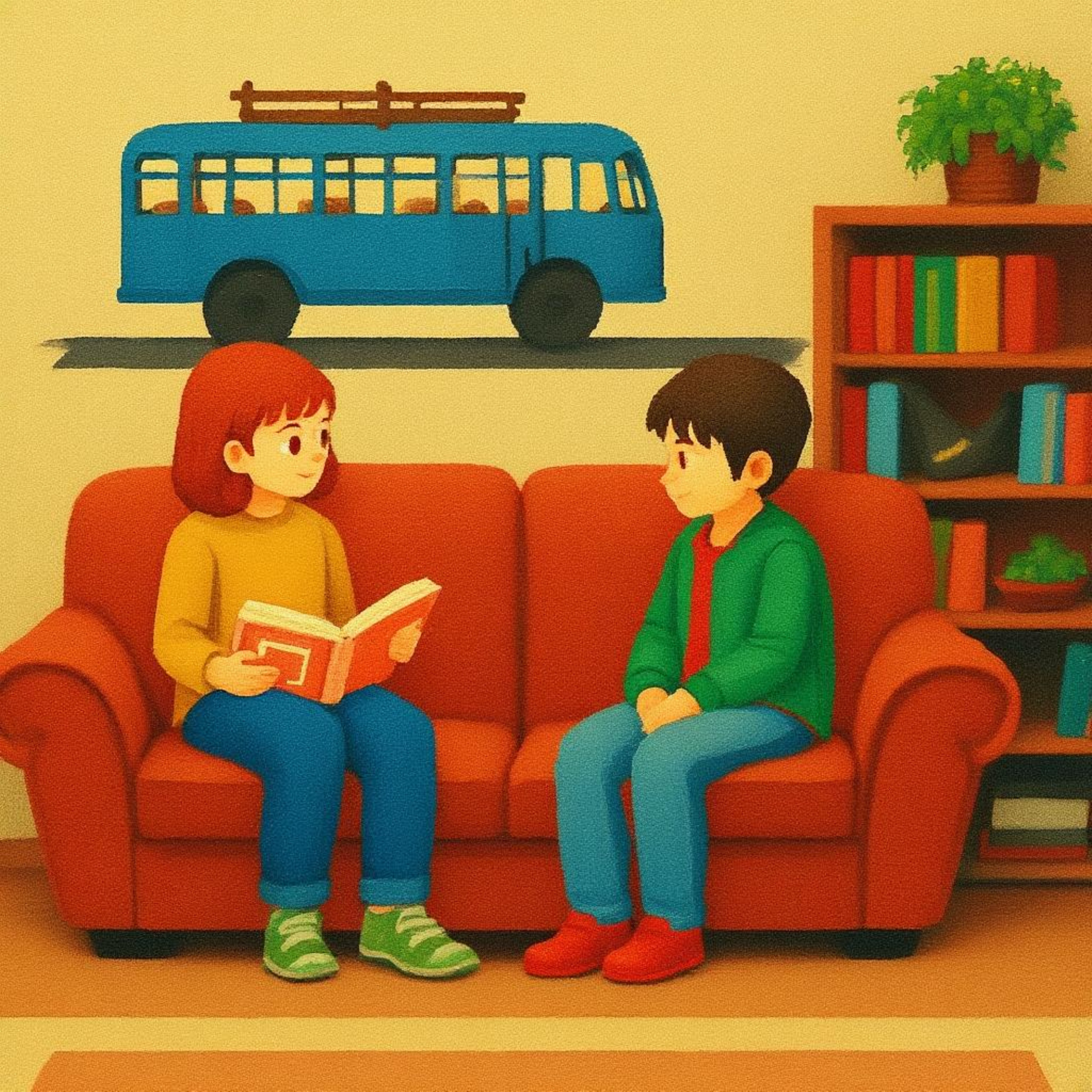} &
\emojiimg{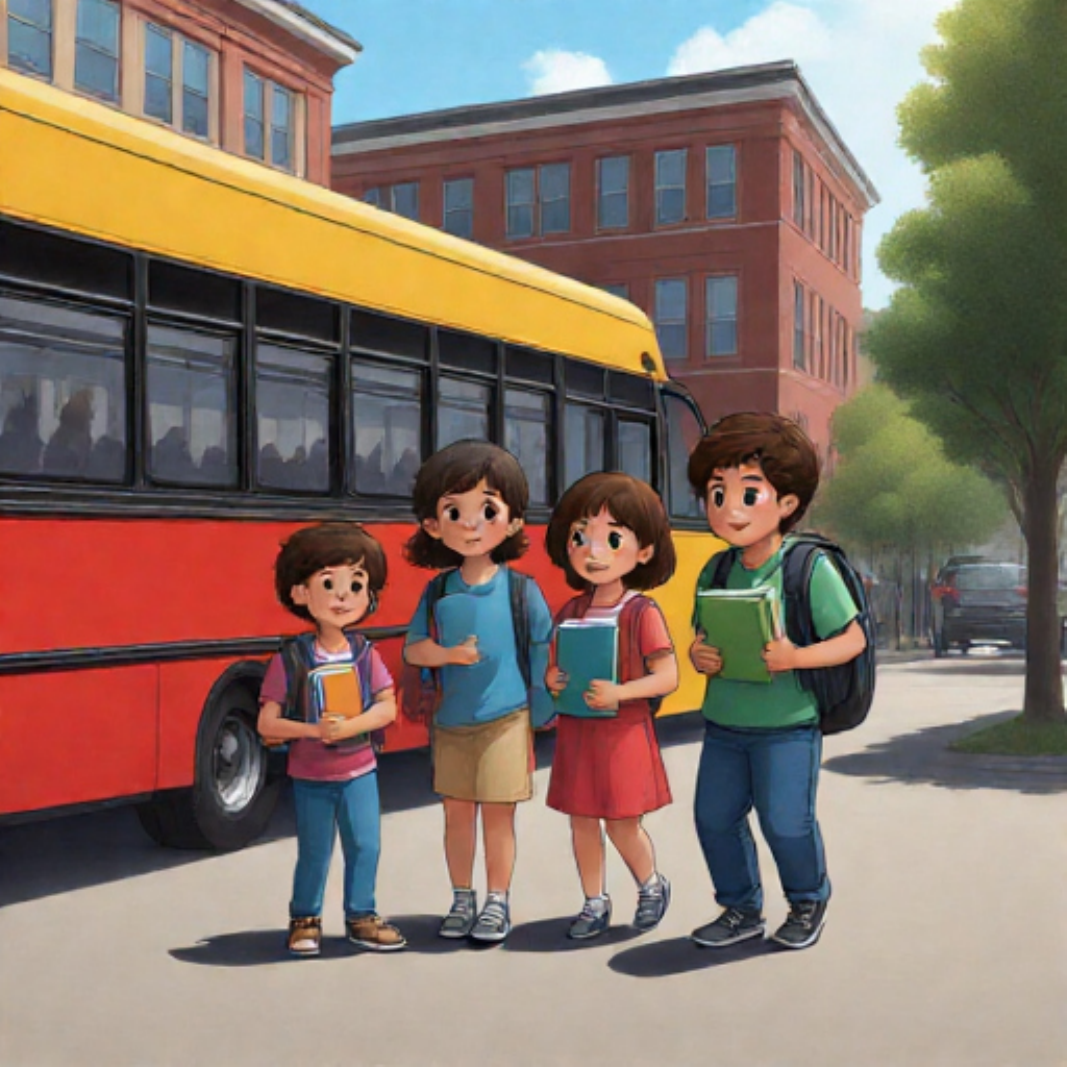} &
\emojiimg{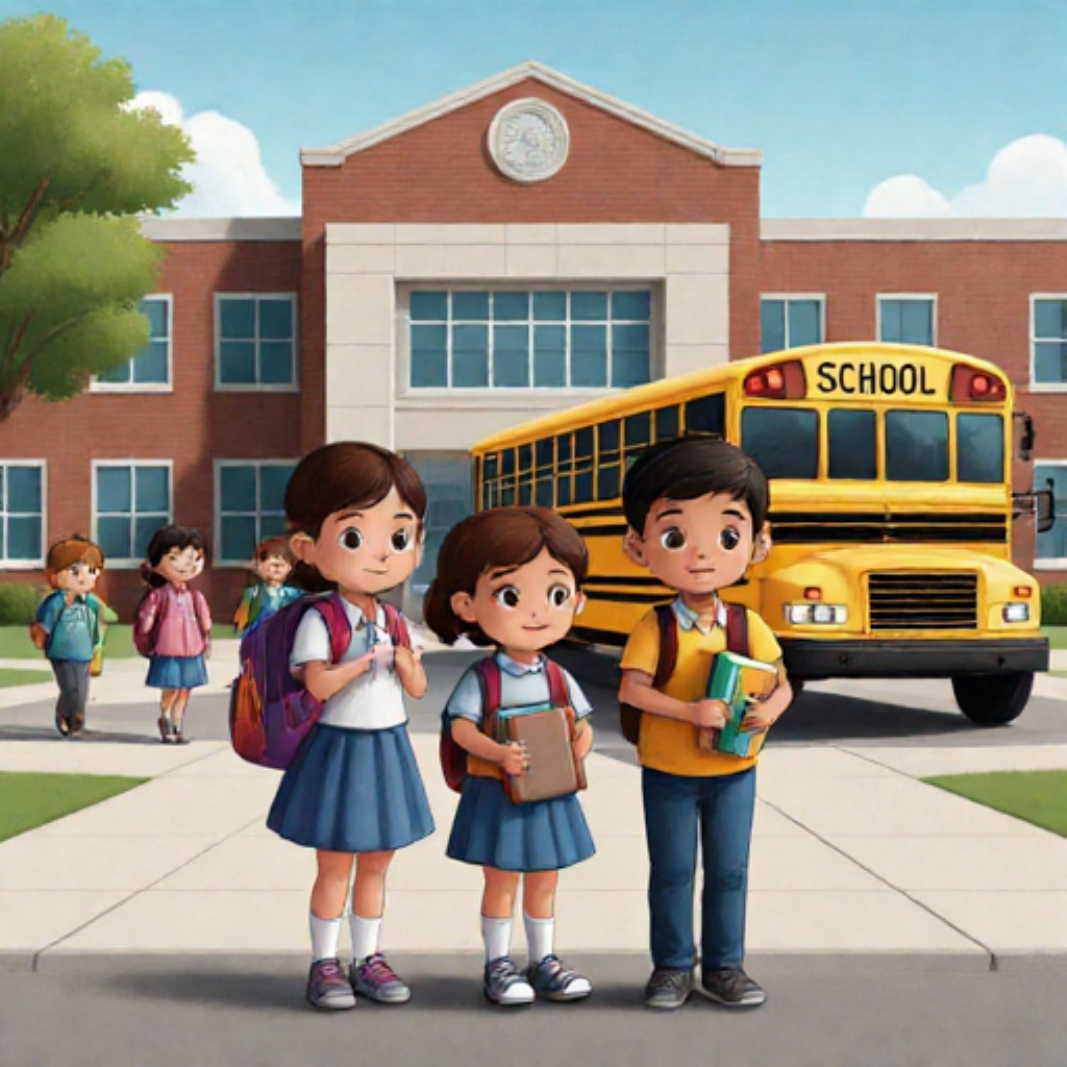} &
\emojiimg{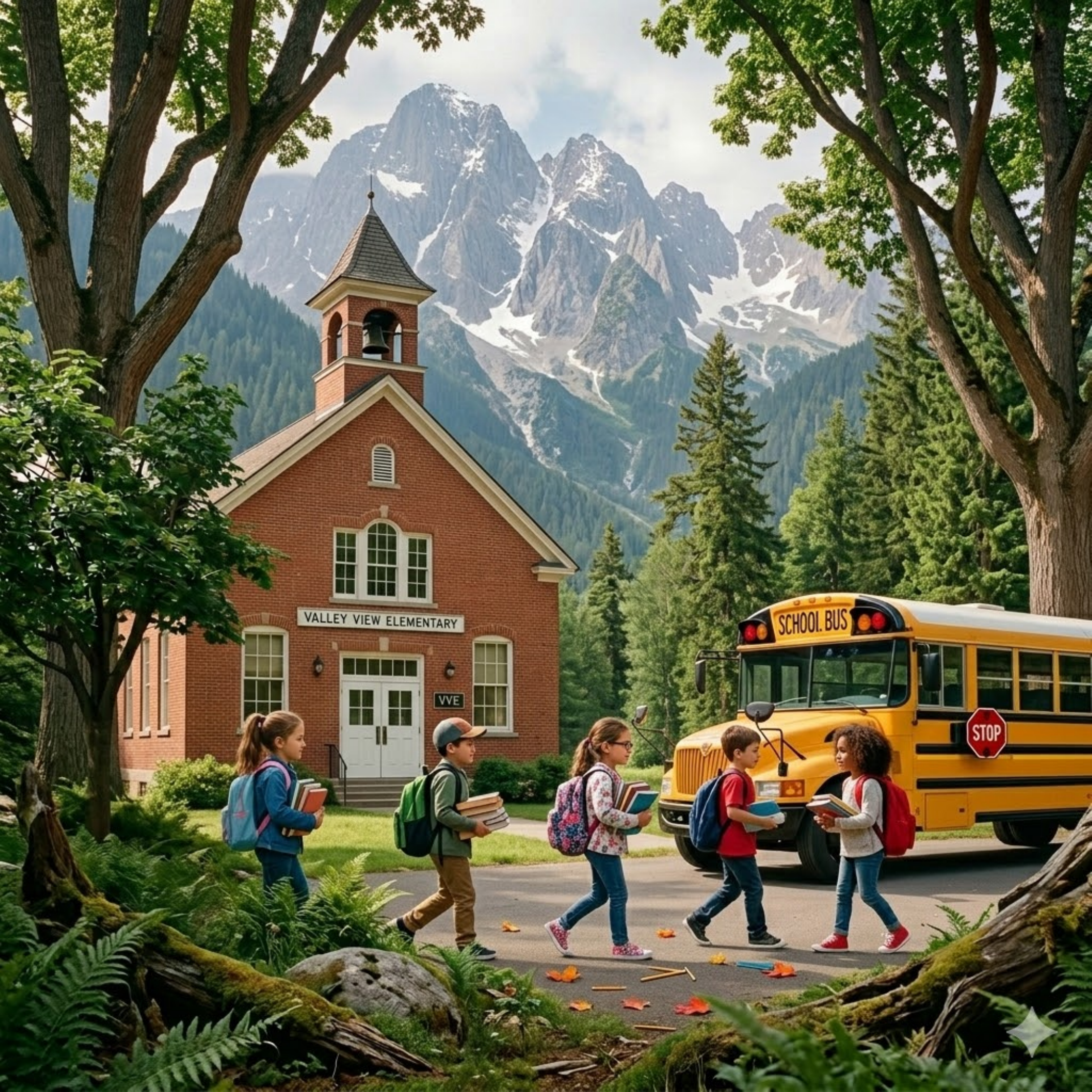} \\[2pt]
\raisebox{0.3cm}{\includegraphics[width=1.2cm]{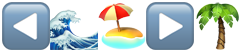}} &
\emojiimg{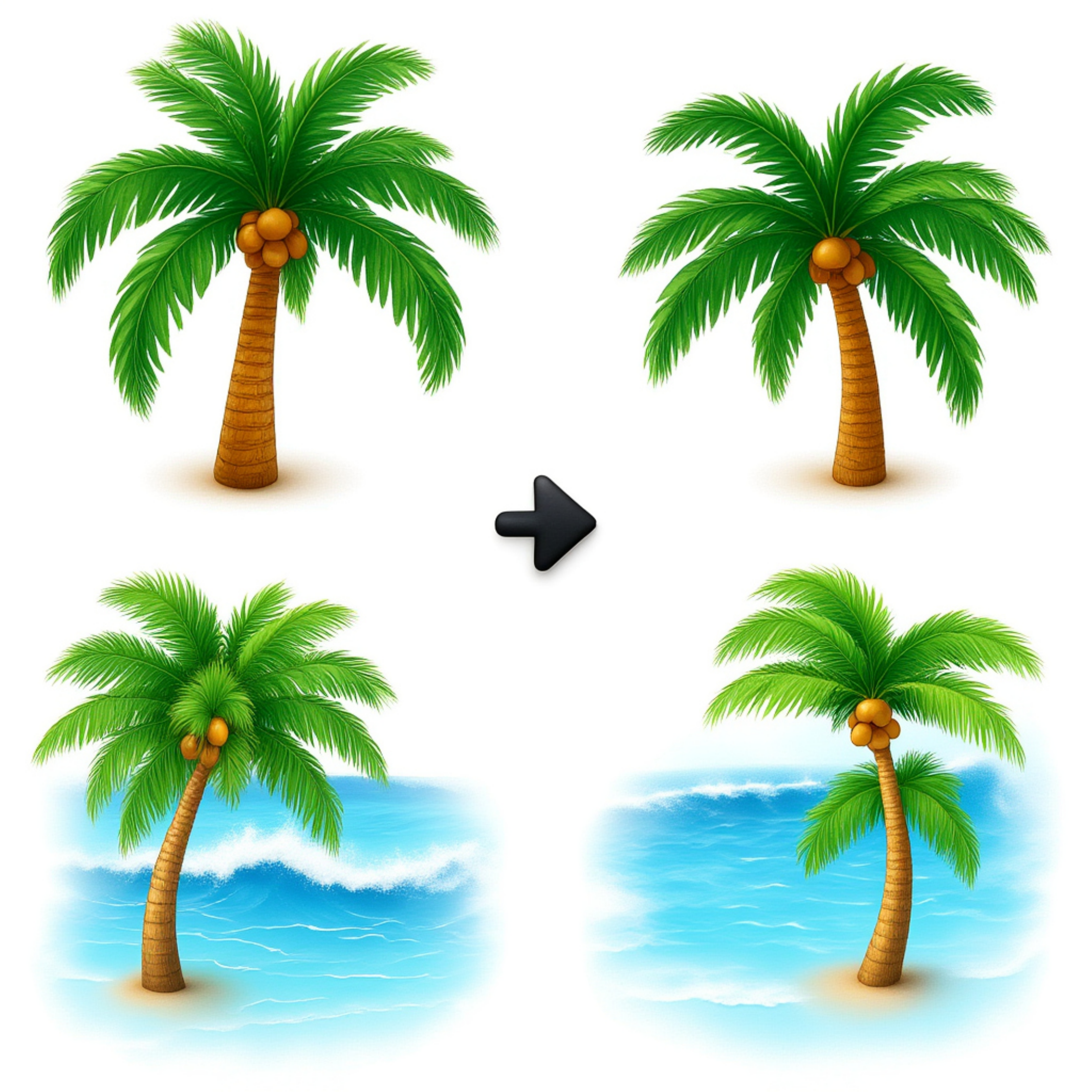} &
\emojiimg{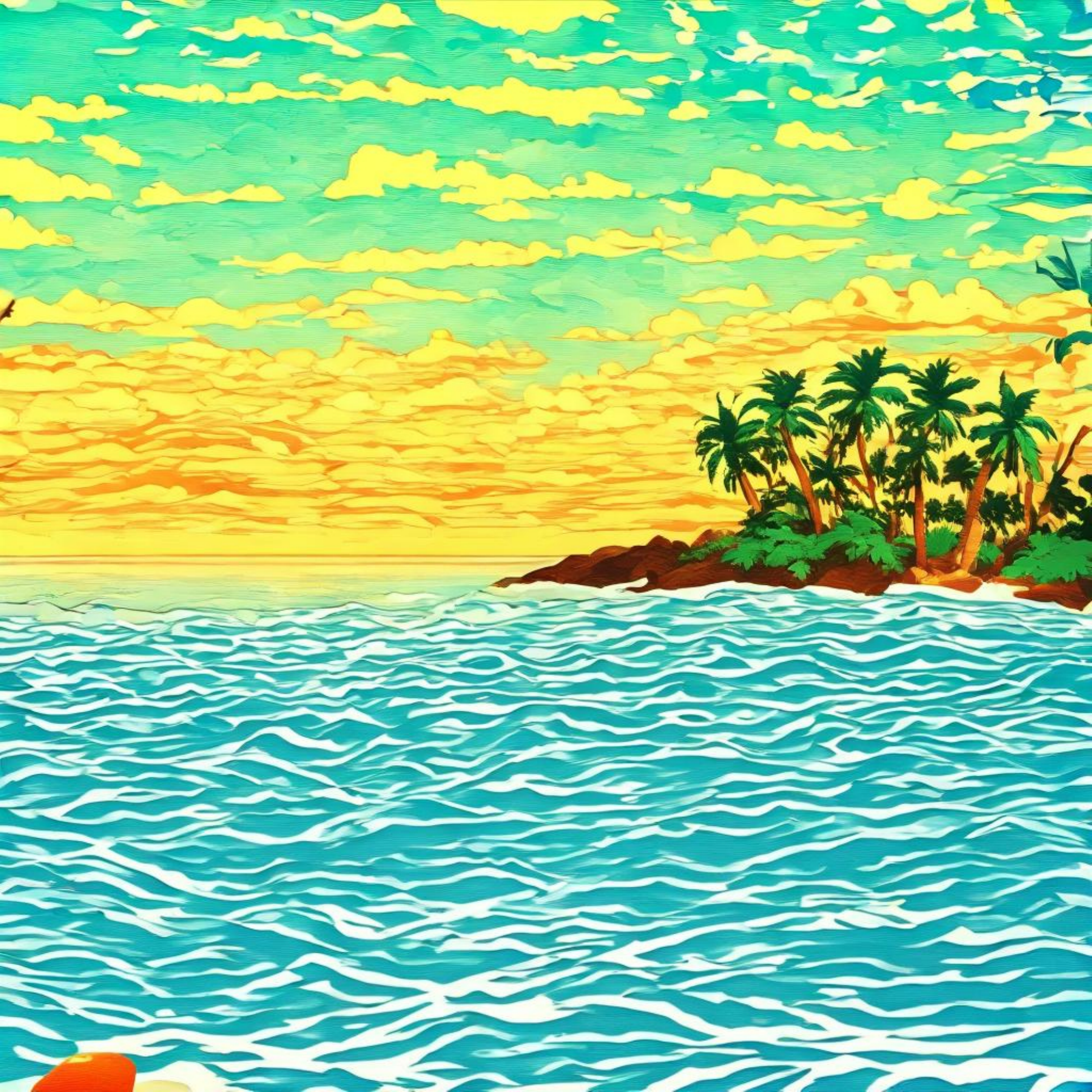} &
\emojiimg{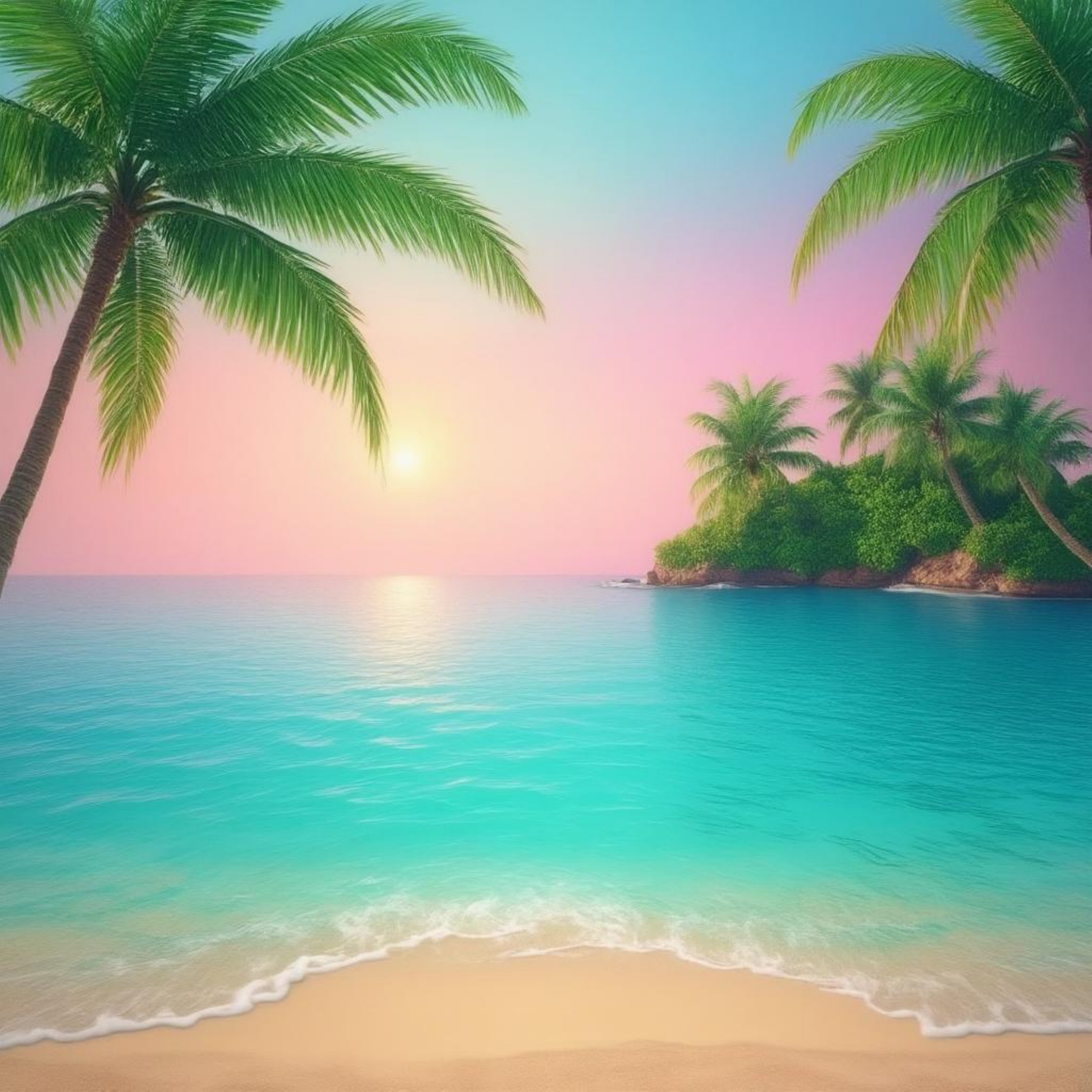} &
\emojiimg{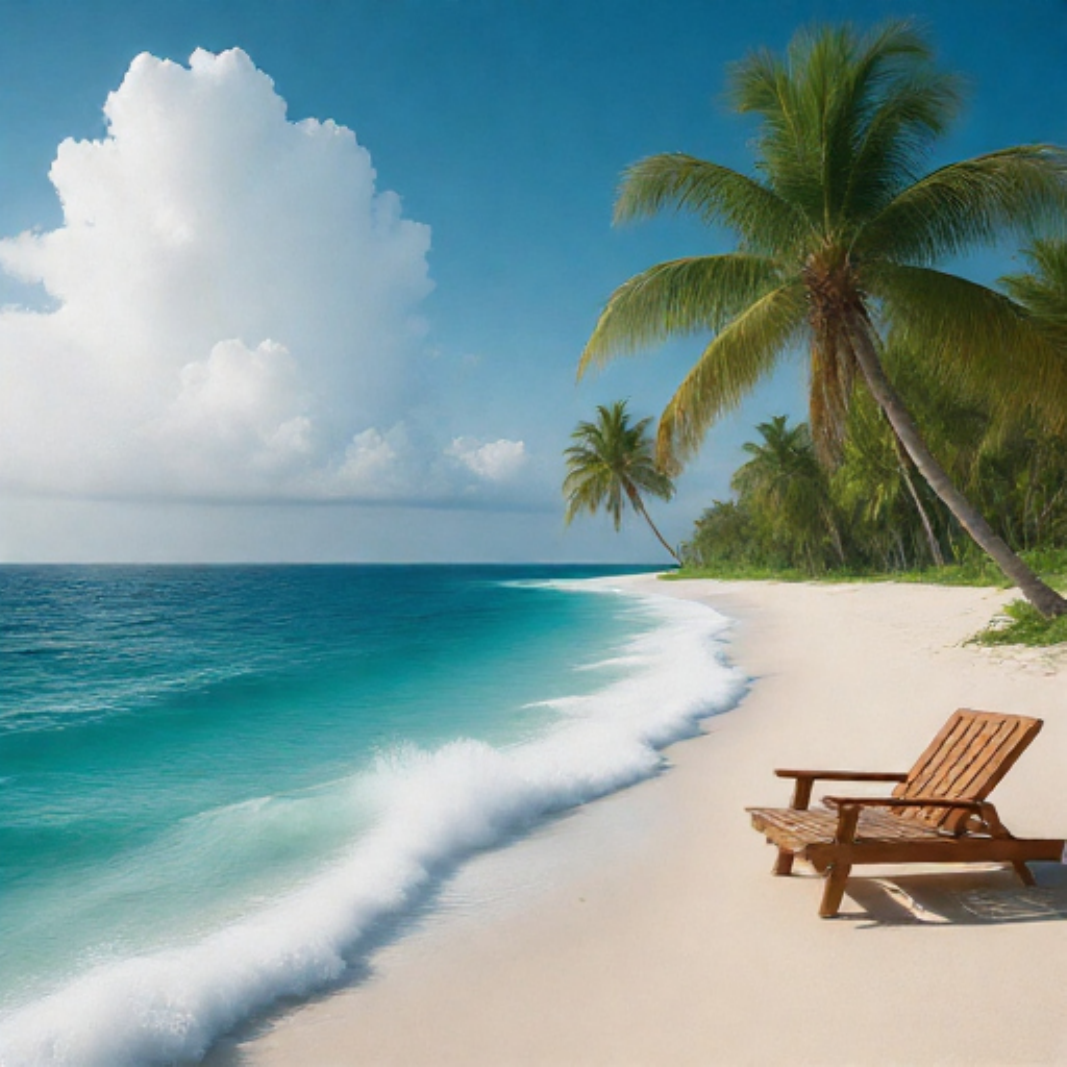} &
\emojiimg{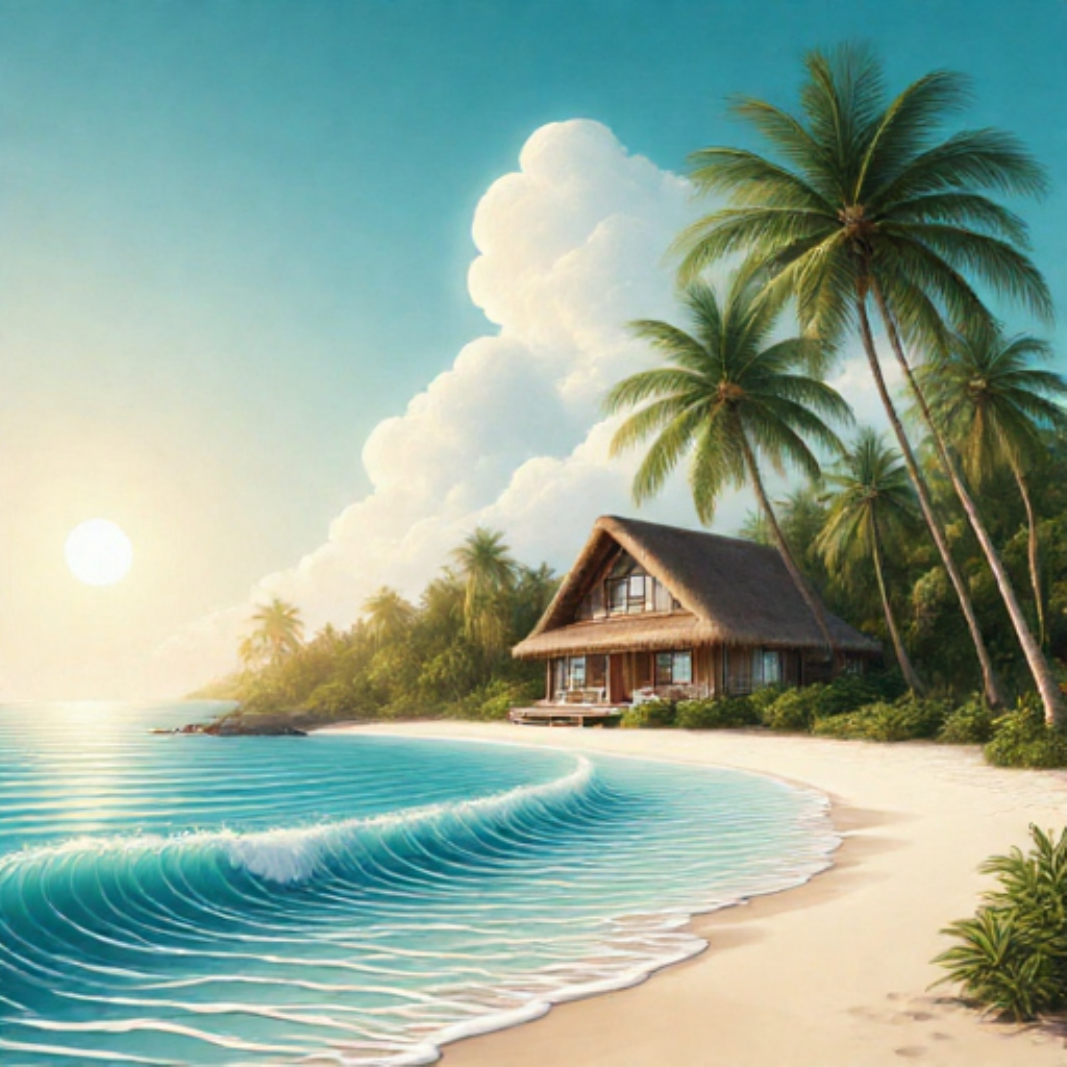} &
\emojiimg{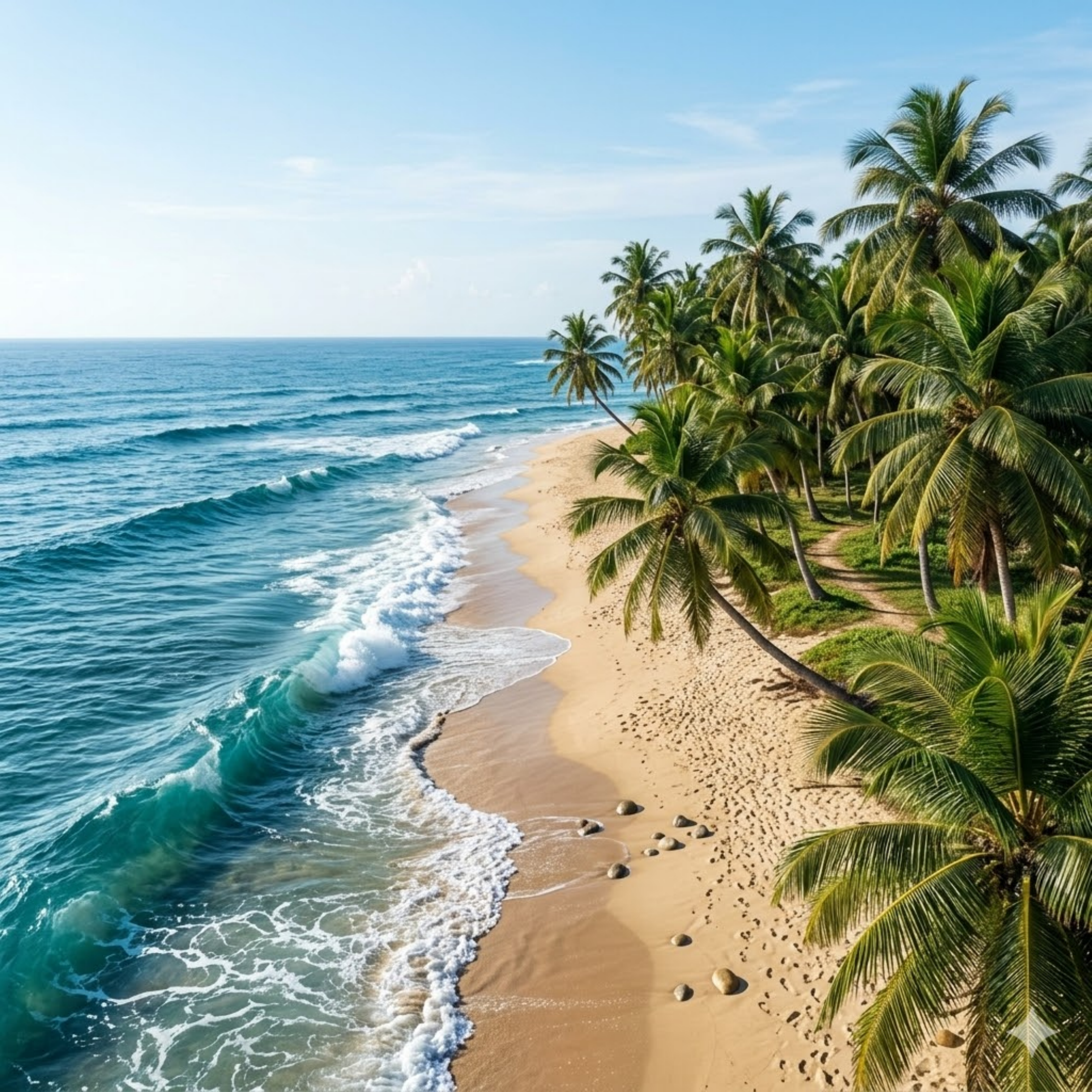} \\
\bottomrule
\end{tabular}
\caption{Emoji scene composition: All prompts use the prefix ``Convert this ASCII art to a realistic image: '' followed by the emoji sequence shown in the first column. \mural~produces realistic interpretations even \textit{without} CoT, with quality improving further when CoT is enabled.}
\label{fig:emoji_scene}
\end{figure*}

\noindent\textbf{Textual layout interpretation.}
We further demonstrate \mural~also interprets ASCII box drawings and text-based floor plans, and generates realistic images.
As an example, we prompt the model with an ASCII representation of a house denoting spatial locations of windows and a door as follows.
{\footnotesize\begin{verbatim}
Convert this drawing to a realistic image:
┌─────┐
│ □ □ │  <- windows
│     │
│ ▢   │  <- door
└─────┘
\end{verbatim}}
\noindent\mural~generates a photorealistic house matching the described layout (Figure~\ref{fig:boxdrawing}).
Even without CoT prompting, \mural~correctly resolves both the semantic content -- inferring that the diagram depicts a house -- and the spatial constraints, such as the number and placement of windows and doors (row 1), or the arrangement of a cup of coffee, a book, and flowers on a table (row 2). Results improve further with CoT enabled at inference.

\newcommand{\tlimg}[1]{\includegraphics[width=0.125\linewidth]{#1}}
\begin{figure*}[t]
\centering
\setlength{\tabcolsep}{1pt}
\scriptsize
\begin{tabular}{@{}c@{\hskip 3pt}cccccc@{}}
\toprule
Prompt & OmniGen2 & Bagel & Bagel+CoT & \mural & \mural+CoT & NanoBanana2 \\
\midrule
\raisebox{0.3cm}{\includegraphics[height=1.4cm]{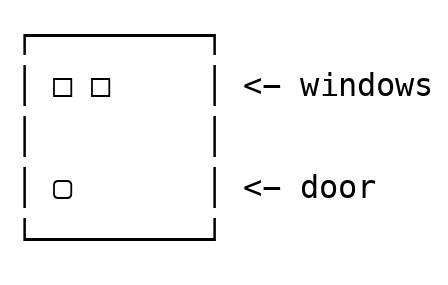}} &
\tlimg{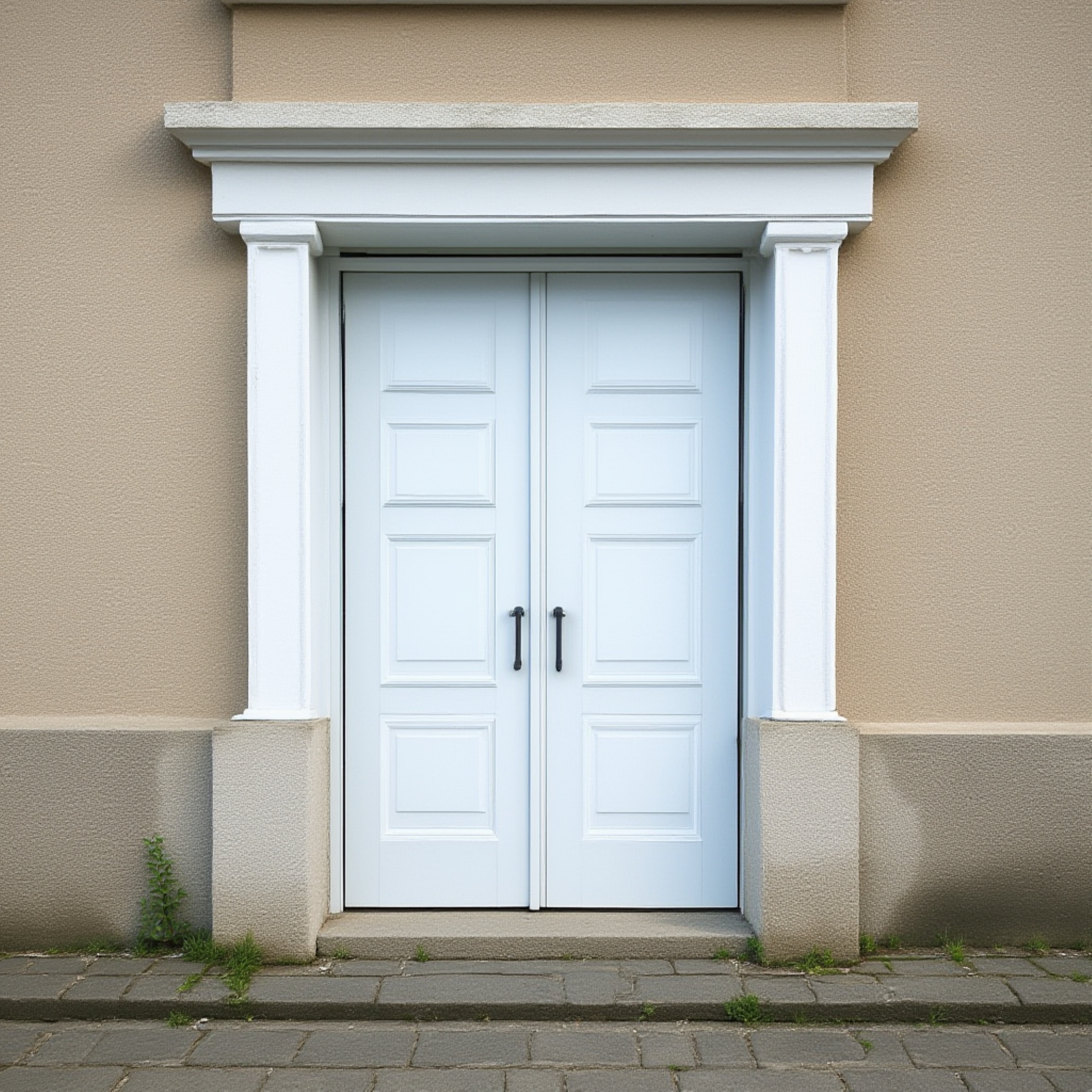} &
\tlimg{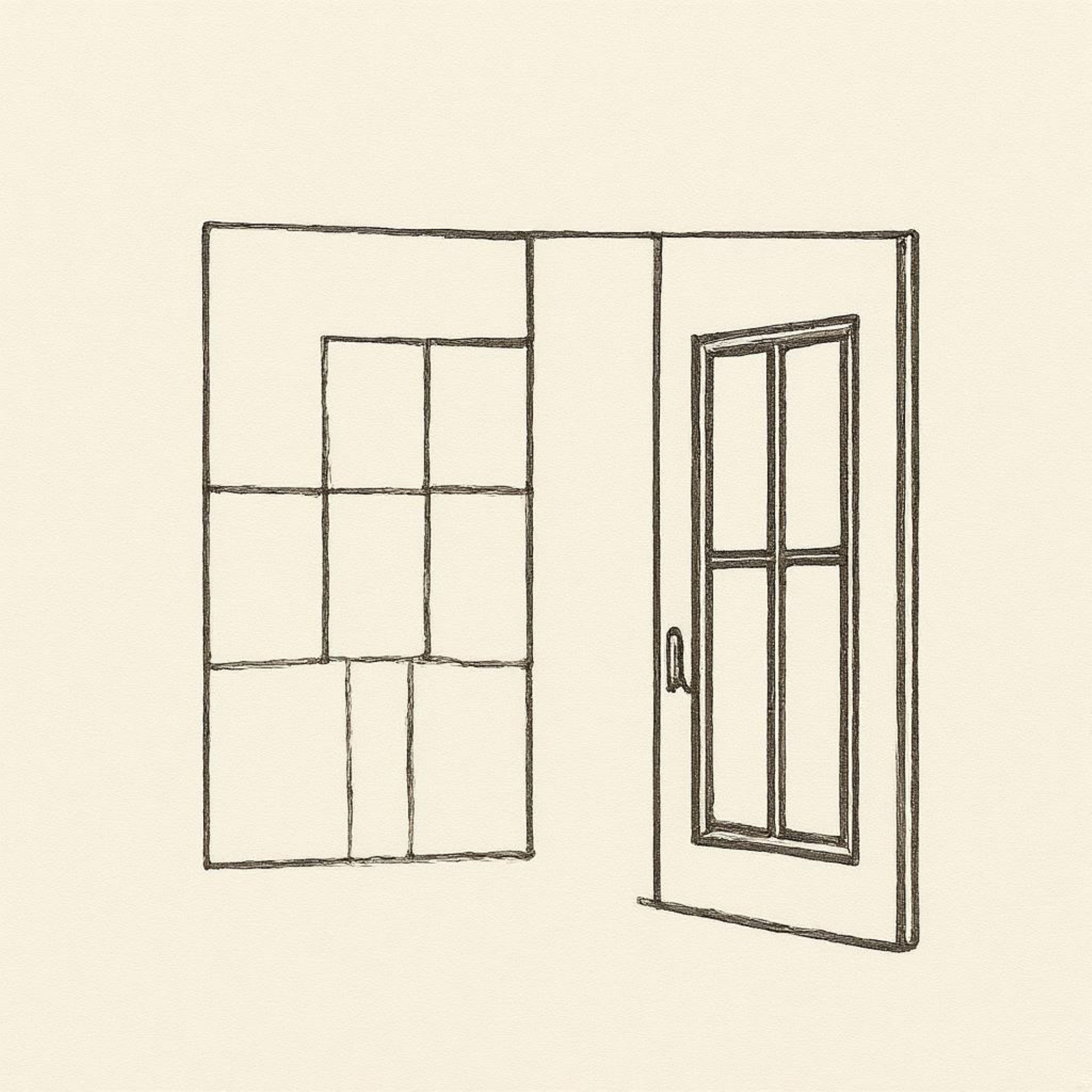} &
\tlimg{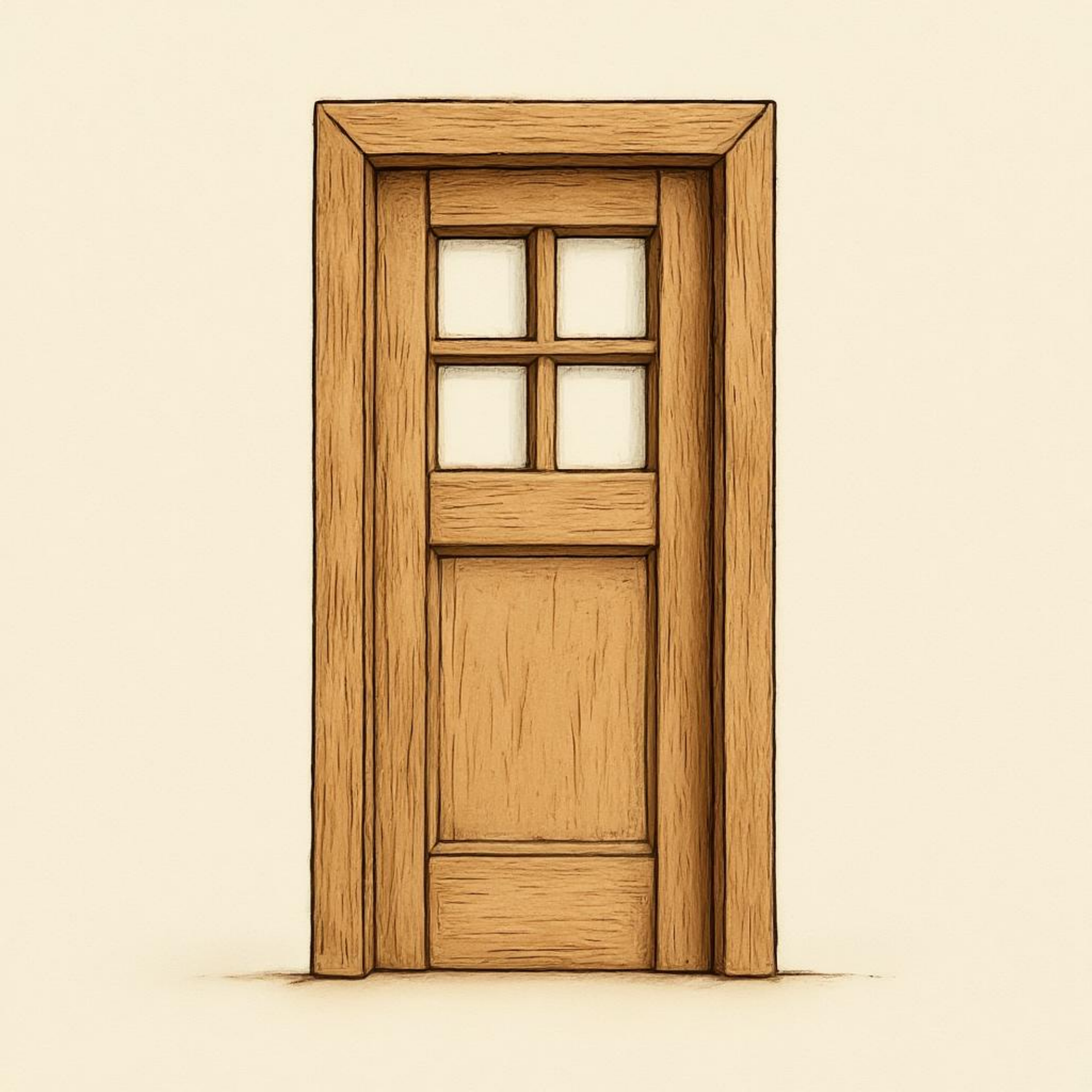} &
\tlimg{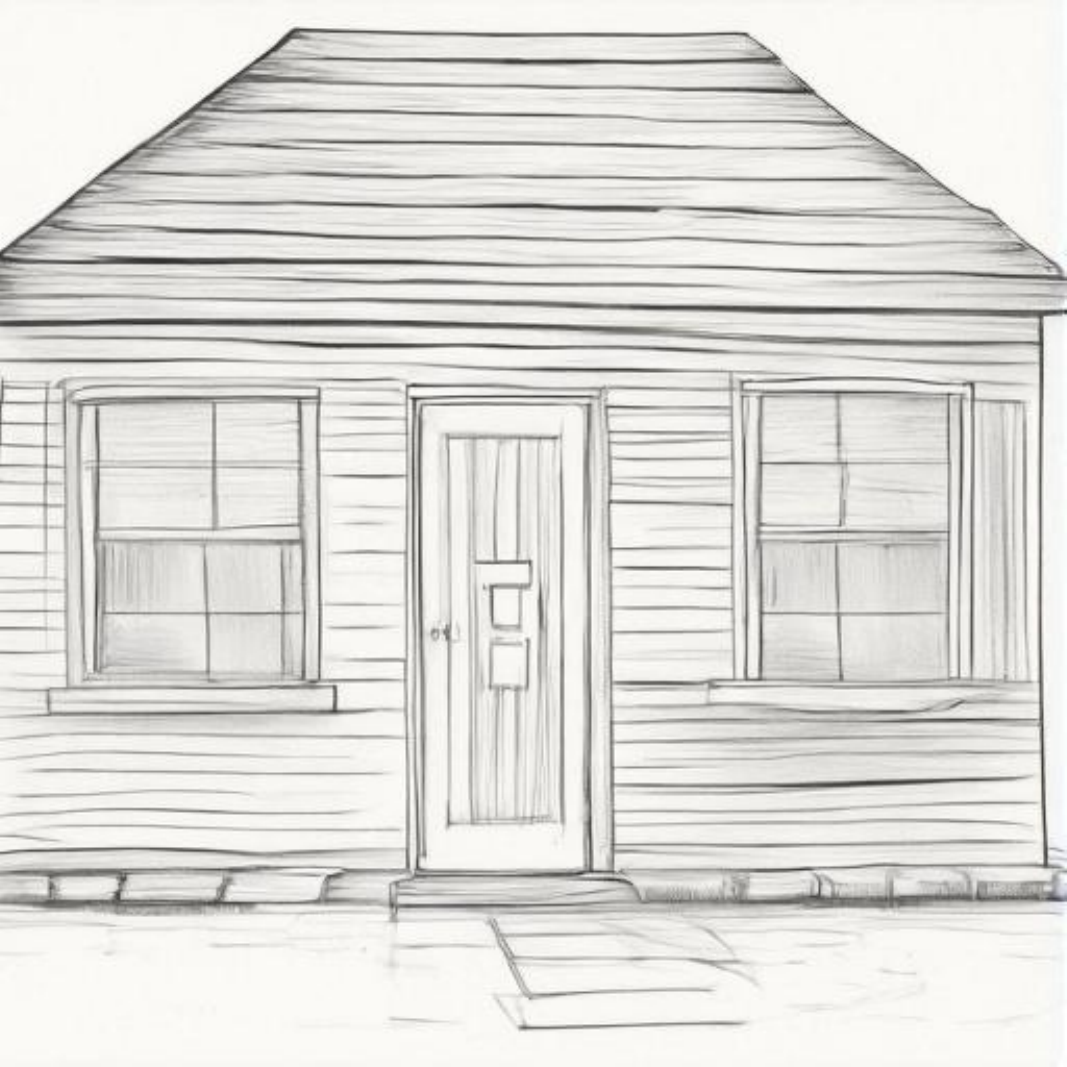} &
\tlimg{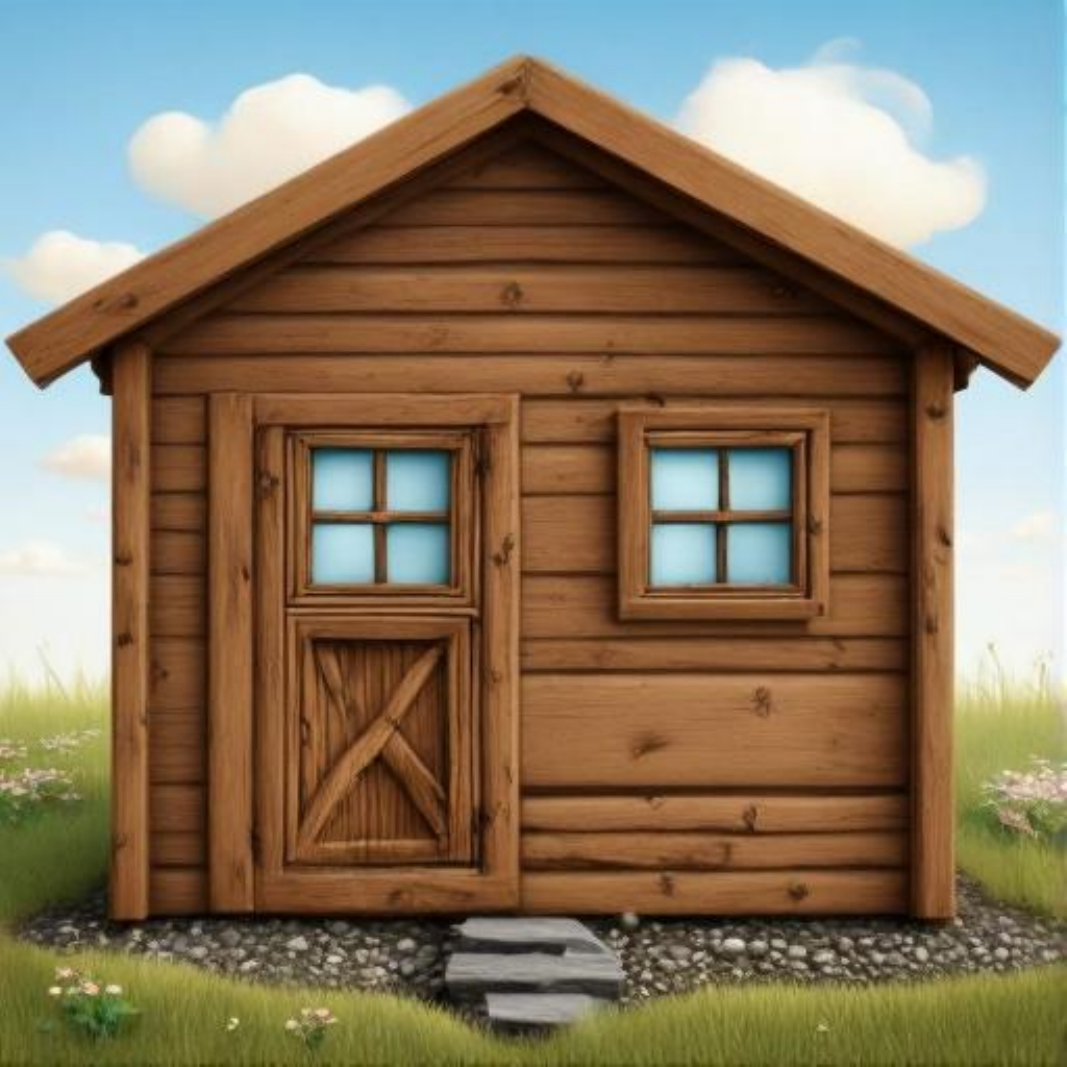} &
\tlimg{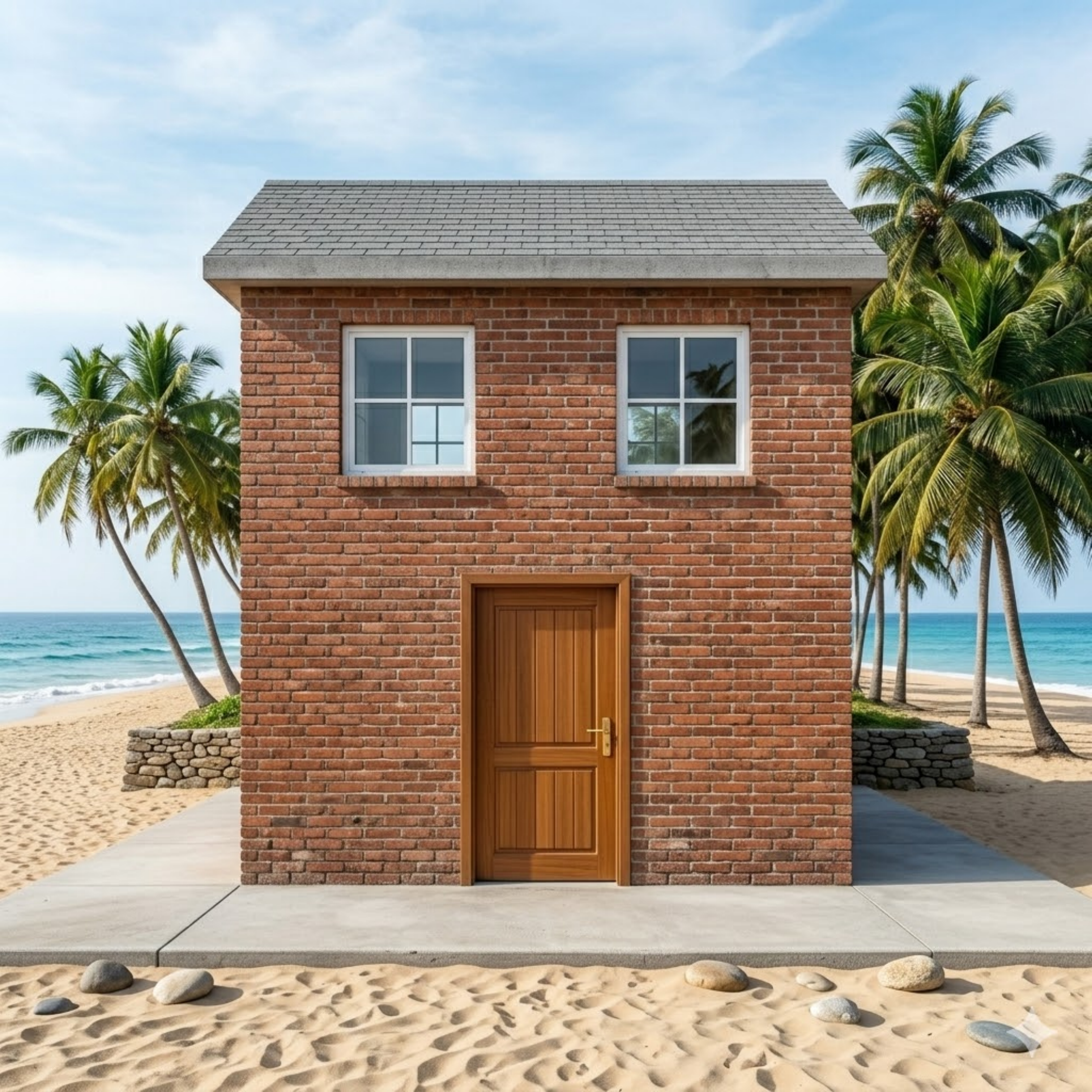} \\[2pt]
\raisebox{0.3cm}{\includegraphics[height=1.4cm]{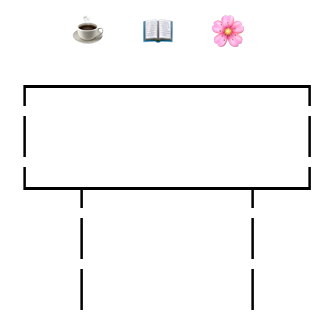}} &
\tlimg{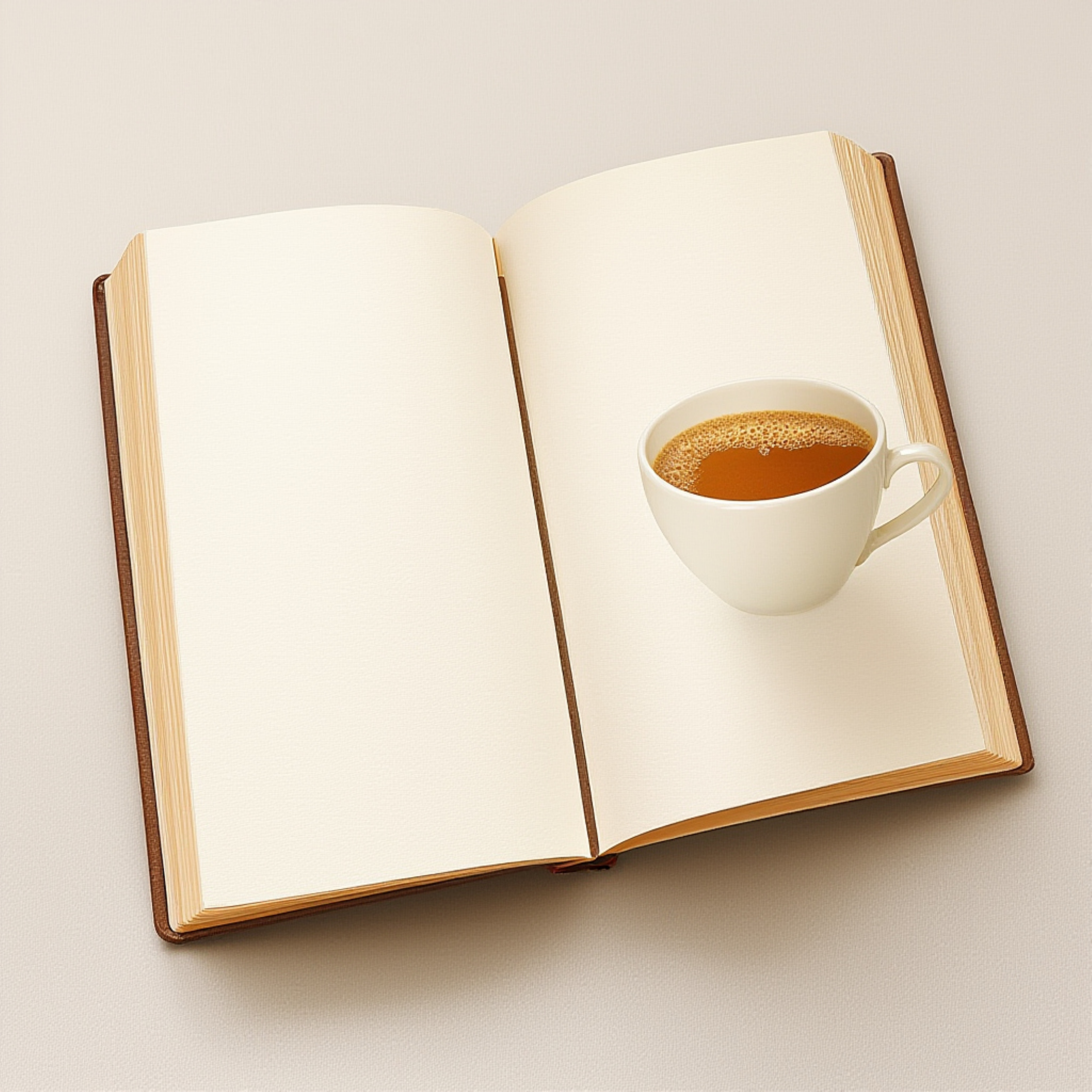} &
\tlimg{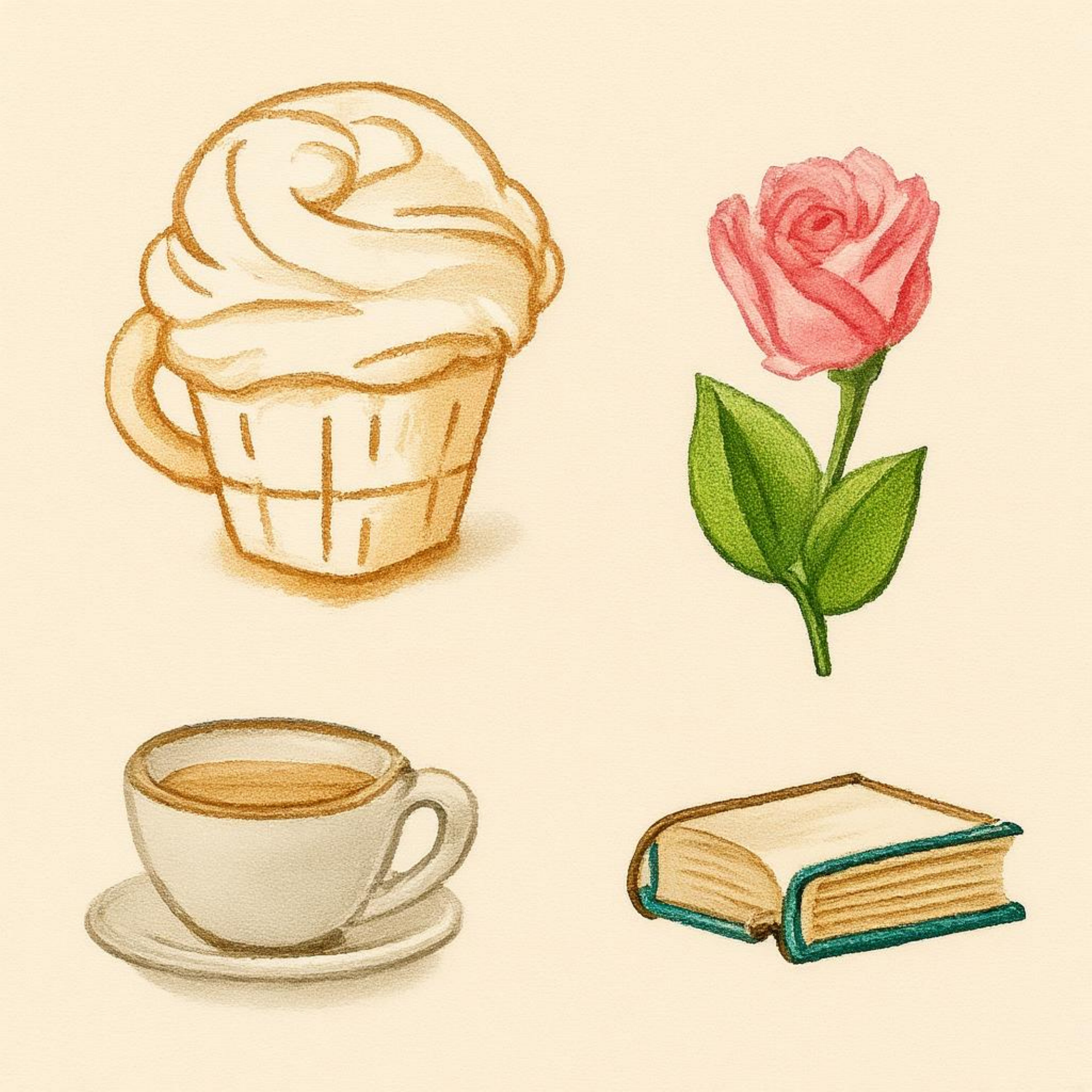} &
\tlimg{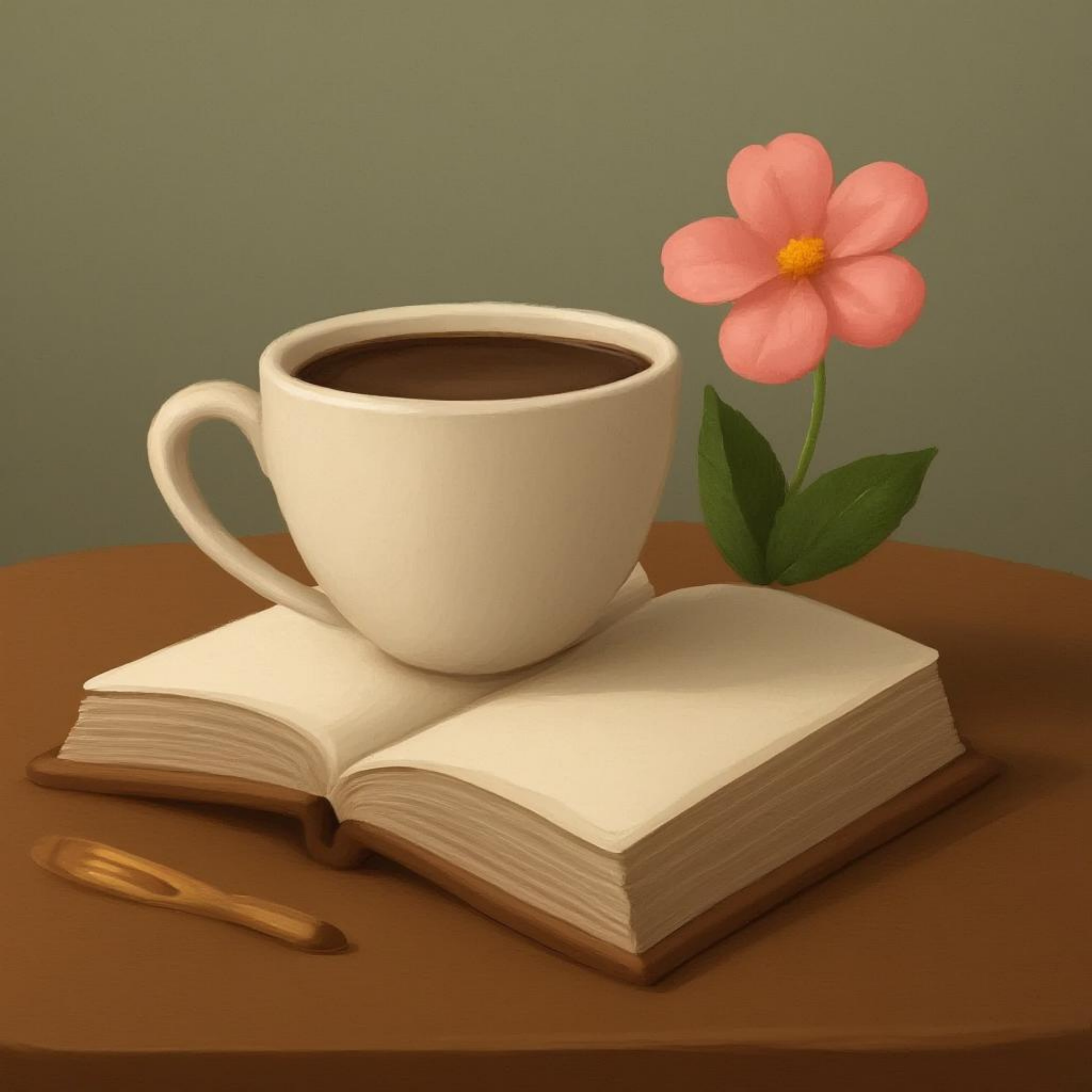} &
\tlimg{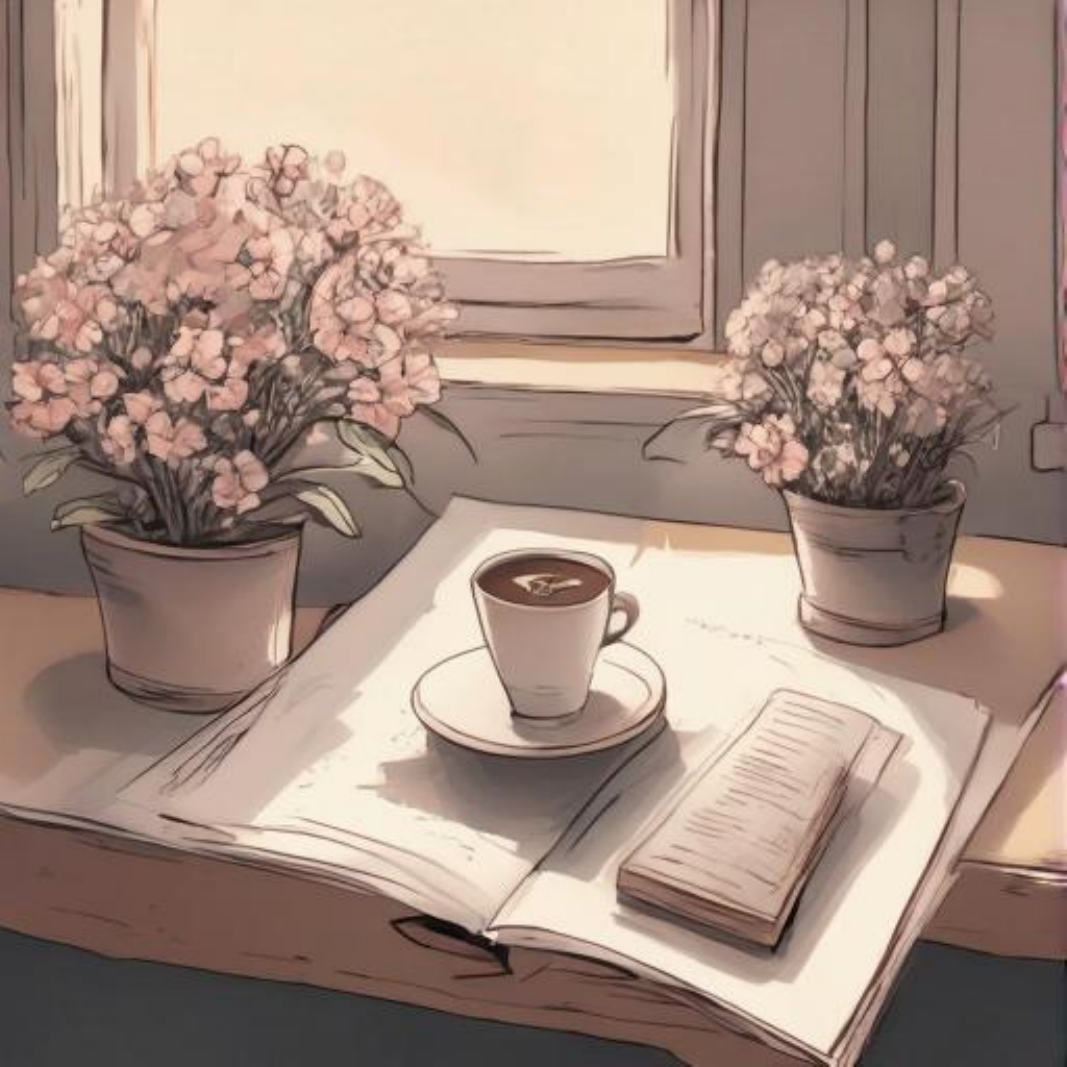} &
\tlimg{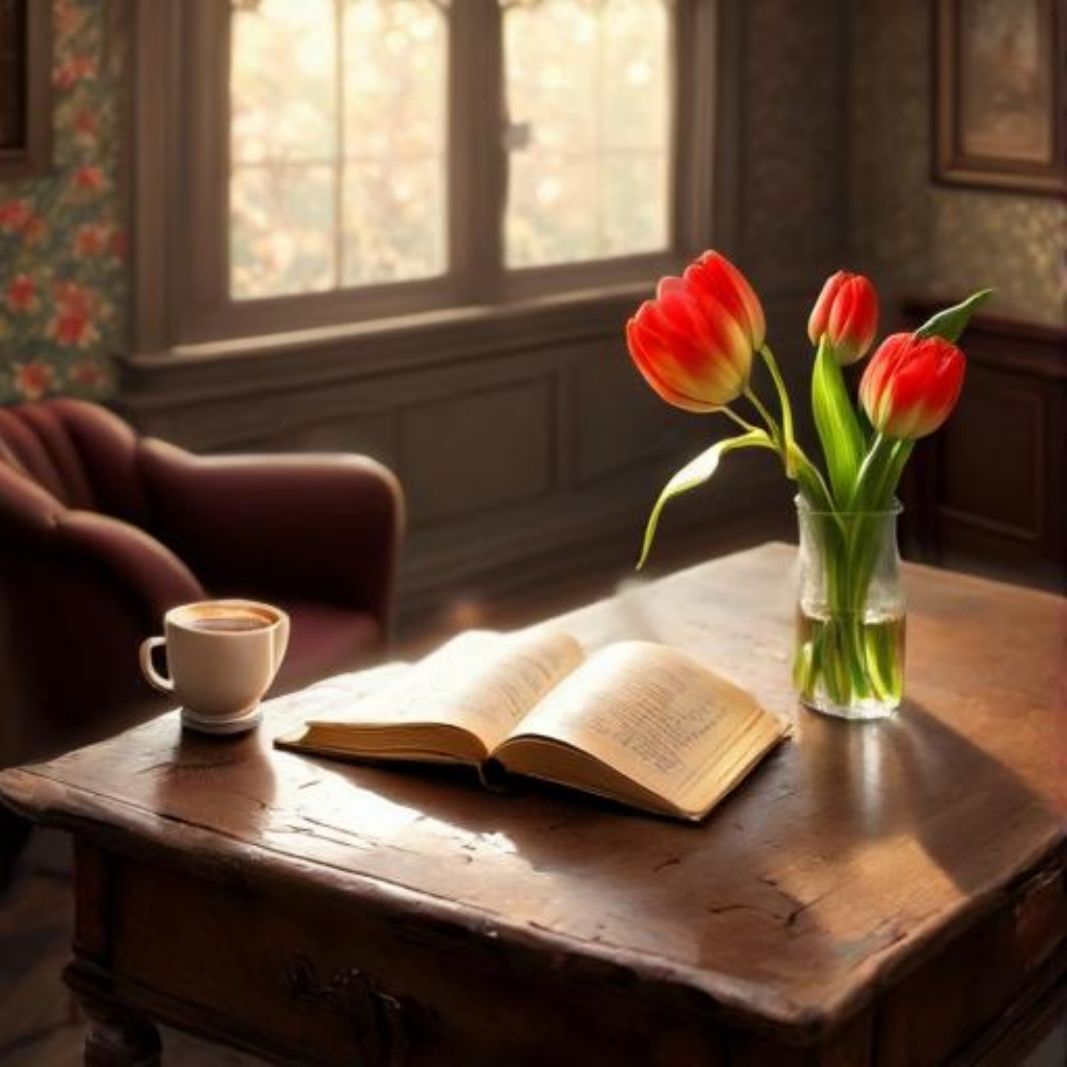} &
\tlimg{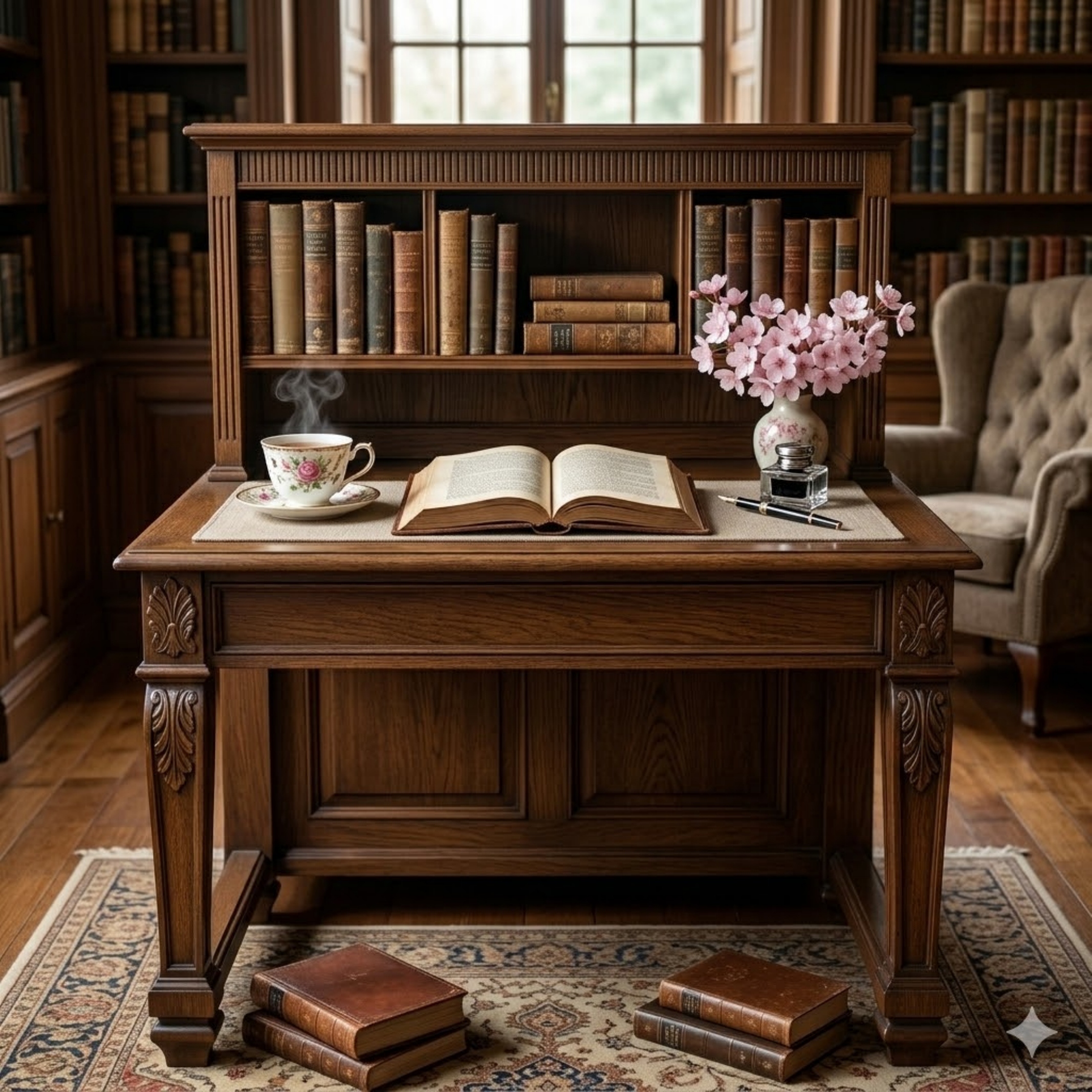} \\
\bottomrule
\end{tabular}
\caption{Textual layout interpretation: All prompts use the prefix ``Convert this drawing to a realistic image: '' followed by textual layouts in ASCII format (optionally with emoji) shown in the left column. \mural~interprets textual layouts and generates corresponding realistic images.}
\label{fig:boxdrawing}
\end{figure*}

\begin{figure}[t]
\centering
\setlength{\tabcolsep}{2pt}
\scriptsize
\begin{tabular}{@{}m{2.4cm}m{0.13\linewidth}m{3.6cm}m{0.13\linewidth}@{}}
\toprule
Prompt & \centering \mural & \centering CoT & \centering\arraybackslash \mural+CoT \\
\midrule
\raggedright \tiny Generate an image of a large animal, a symbol of national pride in Thailand &
\centering\includegraphics[width=\linewidth]{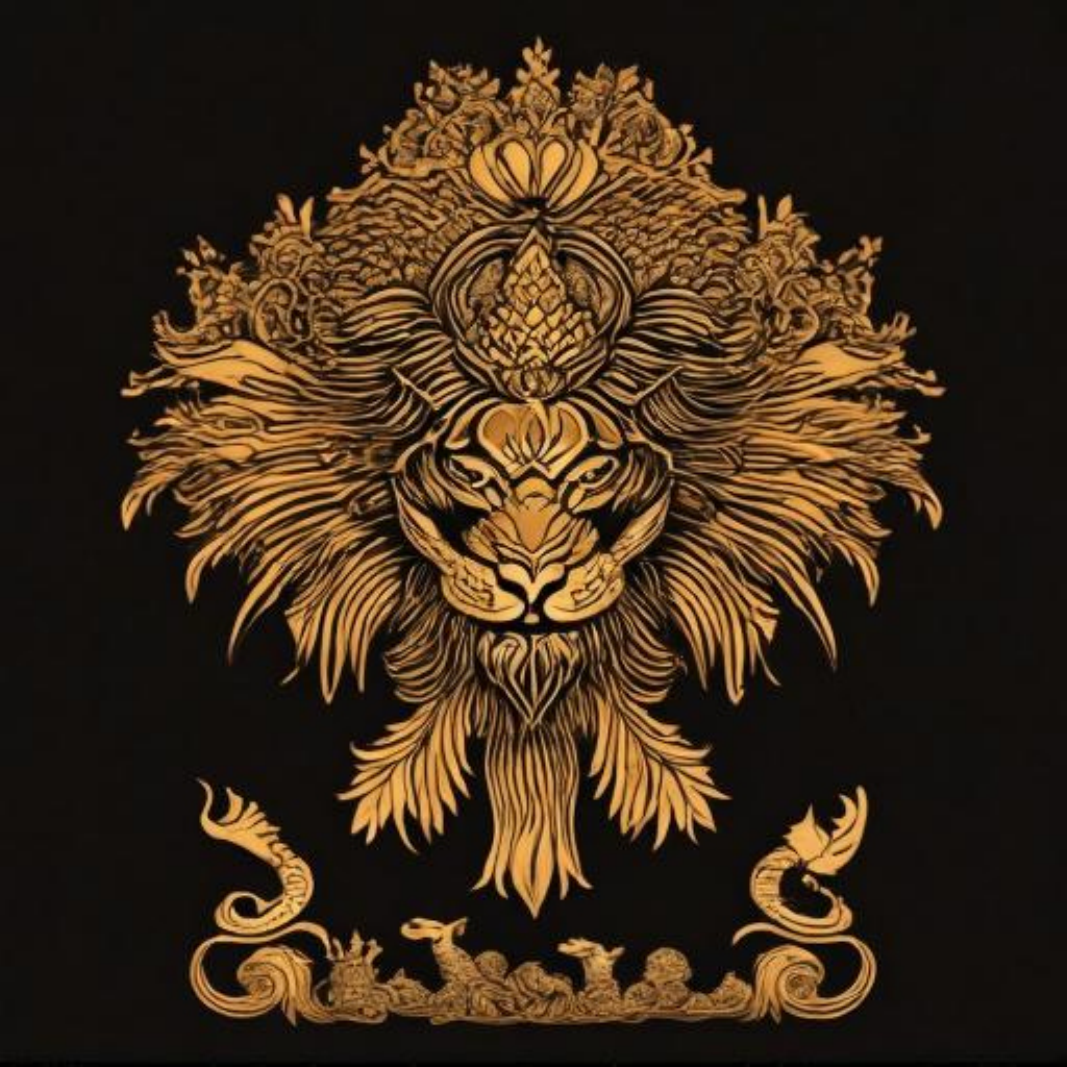} &
\raggedright\tiny \textit{The large animal that symbolizes national pride in Thailand is the elephant. Elephants are a significant cultural and historical symbol in Thailand, often associated with royalty and strength\ldots}  &
\centering\arraybackslash\includegraphics[width=\linewidth]{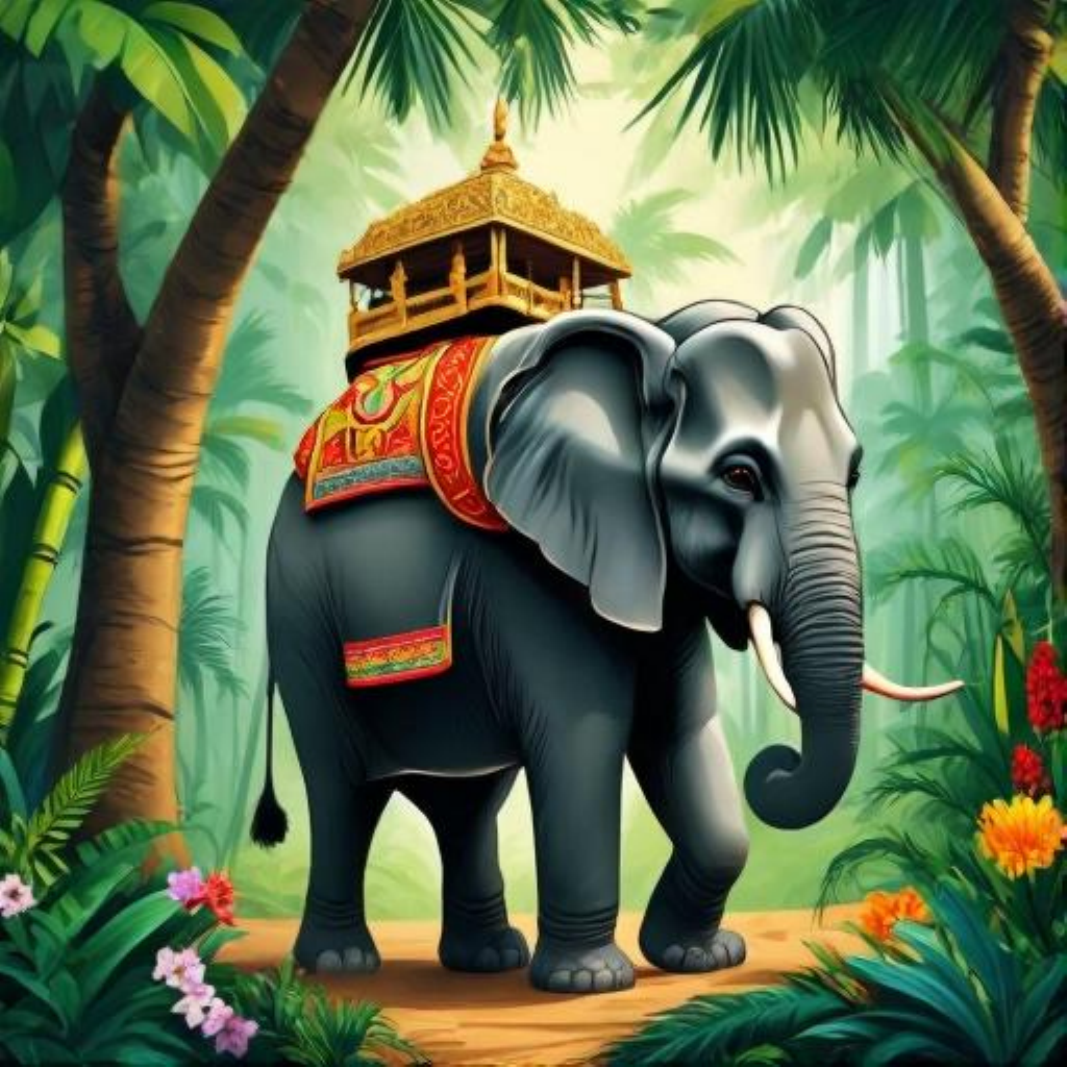} \\[2pt]
\raggedright \tiny A fruit renowned for its sweetness, often regarded as one of the best in the world, particularly in the Philippines &
\centering\includegraphics[width=\linewidth]{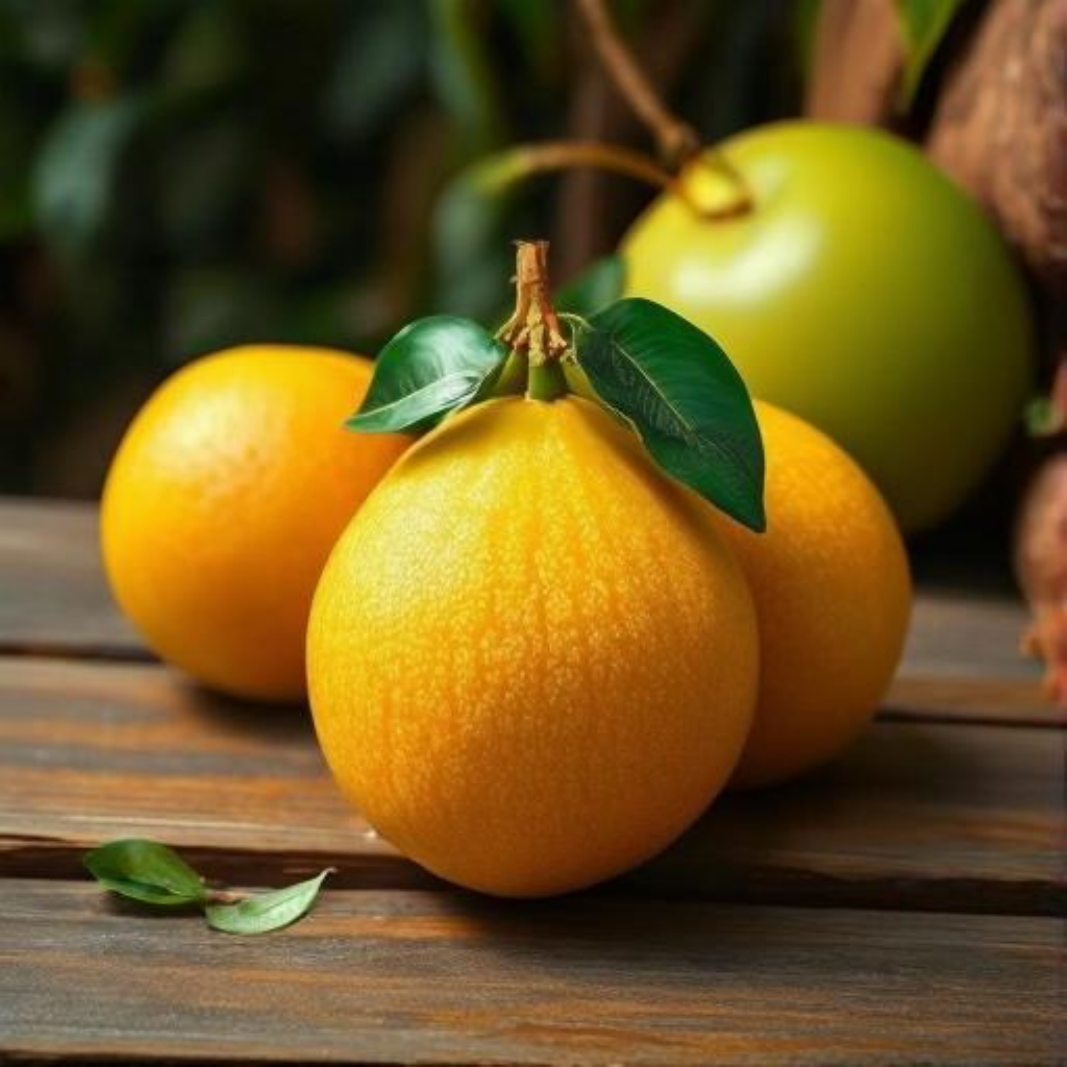} &
\raggedright\tiny \textit{The fruit being described is likely the mango, known for its smooth, sweet flesh and high regard in the Philippines. I should depict a ripe, golden mango\ldots} &
\centering\arraybackslash\includegraphics[width=\linewidth]{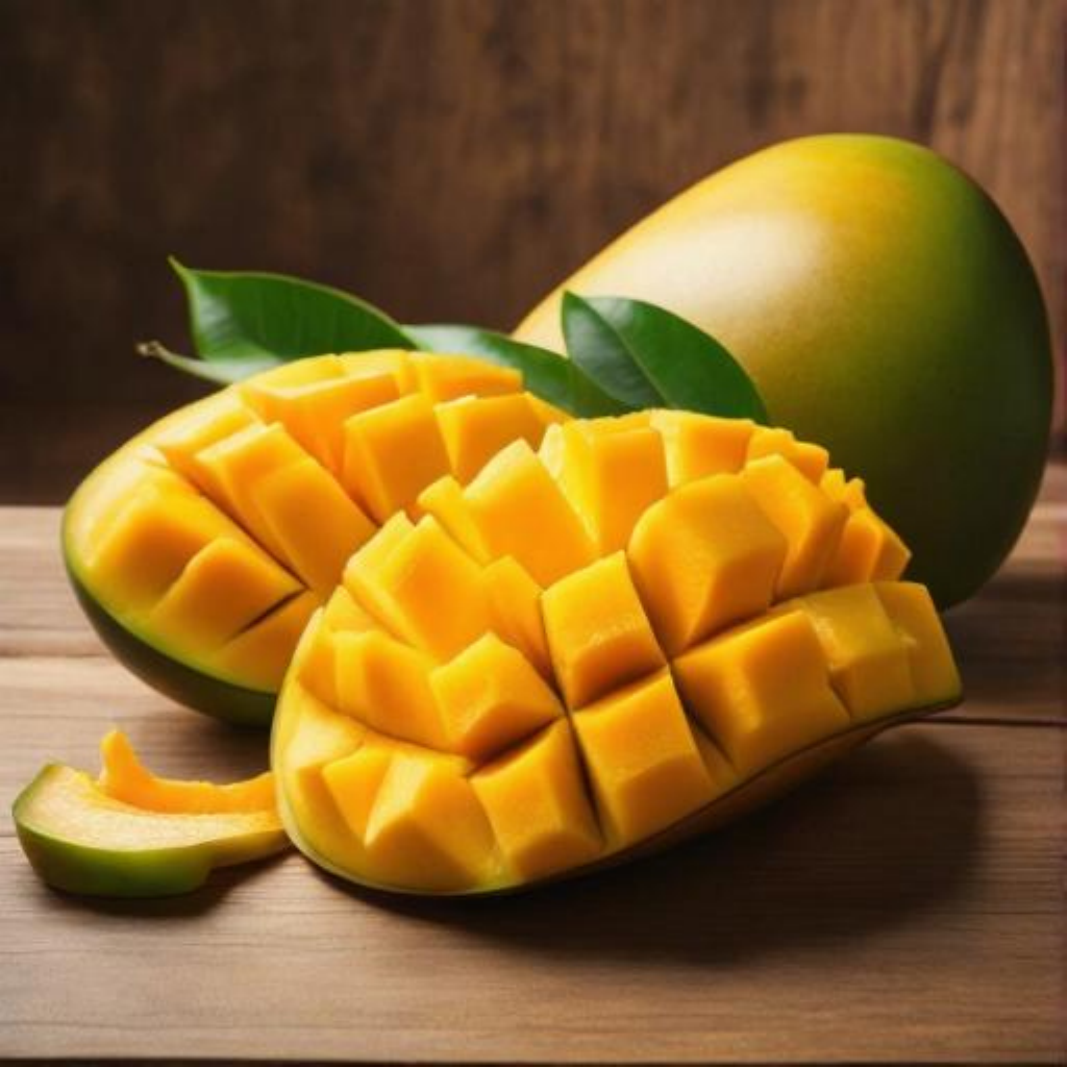} \\
\bottomrule
\end{tabular}
\caption{Reasoning-based on world knowledge: When direct generation without CoT produces incorrect results, prompts requiring world knowledge are resolved correctly when CoT thinking is enabled at inference.}
\label{fig:reasoning}
\end{figure}

\noindent\textbf{Reasoning-based on world knowledge.}
Prompts requiring world knowledge, such as \emph{``Generate an image of a large animal, a symbol of national pride in Thailand''}, are difficult for image generation models trained solely on text-image pairs, as they must deduce the intended subject (in this case, an elephant) without access to broad factual knowledge.
Frozen LLMs, trained on rich text corpora, naturally possess this knowledge.
We find that shared attention alone (without CoT) already improves the image expert's ability to handle such prompts, as evidenced by the WISE benchmark~\cite{wise2024} scores comparing the dense baseline (row 1) and MoT (row 2) in Table~\ref{tab:dense_vs_mot}. See more on this in Section~\S\ref{SS:dense_vs_mot}.
In cases where shared attention is insufficient, enabling CoT allows the LLM to explicitly reason through the prompt and correctly identify the subject as an elephant (previously also demonstrated in~\cite{bagel2025}). Thus, the generation expert produces a corresponding image (Figure~\ref{fig:reasoning}). This accounts for the further gains in row 3 versus row 2 in Table~\ref{tab:dense_vs_mot}.
Again, this capability requires no reasoning data during training; it emerges directly from the LLM's pretrained knowledge.

\subsection{Decoupling effects of shared attention: Dense vs MoT}
\label{SS:dense_vs_mot}

In a controlled experiment, we analyze the gains of the MoT architecture (14B total parameters, with a 7B frozen LLM) over a dense 7B baseline on image generation benchmarks.
Both models are initialized from Qwen2.5-7B-Instruct and trained with identical data, recipe, and compute budget, thus this setup isolates the effect of the underlying MoT design used in \mural.

Since we train solely on image-text with diffusion loss, as expected, the dense model exhibits catastrophic forgetting on text understanding benchmarks, dropping to chance-level performance on MMLU-Pro within 2K steps.
In contrast, MoT preserves all text understanding metrics by construction, since the LLM backbone is frozen.

\begin{table}[!t]
\centering
\caption{Dense versus MoT comparison: MoT with shared attention outperforms the dense baseline on image generation benchmarks while preserving text understanding. MoT strictly improves over the dense baseline. Enabling CoT thinking at inference further improves on WISE, which requires world knowledge.}
\label{tab:dense_vs_mot}
\footnotesize
\begin{tabular}{l|c|c|c|c}
\toprule
\textbf{Model} & \ \textbf{GenEval}$\uparrow$ \ & \ \textbf{WISE}$\uparrow$ \ & \ \textbf{DPGBench}$\uparrow$ \ & \ \textbf{Text Understanding} \\
\midrule
Dense & 0.84 & 0.44 & 86.32 & \ding{55} degraded \\
MoT & \textbf{0.85} & 0.51 & 86.75 & \ding{51} preserved \\
MoT + CoT \ & -- & \textbf{0.66} & -- & \ding{51} preserved \\
\bottomrule
\end{tabular}
\end{table}

We report our findings in Table~\ref{tab:dense_vs_mot}.
First, the dense model achieves competitive GenEval and DPGBench scores, ranking above most open-source models. However, its weak WISE score (0.44) indicates that the training data used throughout this work lacks in world knowledge diversity.
Second, MoT without CoT exceeds the dense model on both GenEval and DPG-Bench, confirming that freezing the LLM incurs no loss in image generation quality. As a consequence, any future data improvements specifically targeted to improve standalone image generation models should translate to our MoT design.
Moreover, the shared attention in MoT boosts the WISE score to 0.51 by leveraging the frozen LLM's factual knowledge.
Third, with CoT thinking enabled at inference, MoT achieves a WISE score of 0.66, surpassing state-of-the-art models including QwenImage (0.62).
Fourth, the dense model does not generate meaningful results on all emergent capability tasks described in \S\ref{SS:emergent} confirming that these capabilities arise from the shared attention mechanism rather than the training data. With the exception of world-knowledge reasoning as measured by WISE, these emergent capabilities are not captured by existing quantitative benchmarks.

\subsection{Comparison to the state-of-the-art models}
\label{SS:sota}

\noindent\textbf{Comparison with unified models.}
Table~\ref{tab:unified_comparison} compares two variants of \mural -- \mural~3B (6B total parameters) and \mural~7B (14B total parameters) -- against unified understanding and generation models.
We also mark data used by different models.
Nomenclature to read data types: T2T: text-to-text, I2T: image-to-text, T2I: text-to-image, INT: interleaved, CoT: chain-of-thought/reasoning/reflection data.

\begin{table}[!t]
\centering
\caption{Comparison with unified understanding and generation models. $^*$denotes parameters of the image generation expert, $^{**}$parameters of the text encoder, \textsuperscript{\dag}inference with CoT, and \textsuperscript{\ddag}evaluation with prompt rewriting. \mural~achieves best results among all models that preserve text/visual understanding.}
\label{tab:unified_comparison}
\resizebox{\linewidth}{!}{
\begin{tabular}{l|c|ccc|cc|c}
\toprule
 \multirow{2}{*}{\textbf{Model}} & \multirow{2}{*}{\textbf{Size}} & \multicolumn{3}{c|}{\textbf{Image Generation}} & \multicolumn{2}{c|}{\textbf{Understanding}} & \multirow{2}{*}{\textbf{Data type}} \\
\cmidrule(lr){3-5}\cmidrule(lr){6-7}
 & & \ \textbf{GenEval}$\uparrow$ \ & \ \textbf{WISE}$\uparrow$ \ & \ \textbf{DPG}$\uparrow$ \ & \ \textbf{Text} \ & \ \textbf{Visual} \ & \\
\midrule
\multicolumn{8}{l}{\textit{Joint multi-modal training}} \\
Chameleon~\cite{chameleon2024} & 7B & 0.39 & -- & -- & \ding{55} & \ding{55} & T2T T2I INT \\
Transfusion~\cite{transfusion2024} & 7B & 0.63 & -- & -- & \ding{55} & \ding{55} & T2T T2I \\
Show-o~\cite{showo2024} & 1.3B & 0.53 & 0.35 & 67.27 & \ding{55} & \ding{55} & T2T I2T T2I \\
Emu3~\cite{emu3_2024} & 8B & 0.54 & 0.39 & 80.6 & \ding{55} & \ding{55} & T2T I2T T2I \\
Janus-Pro~\cite{januspro2025} & 7B & 0.80 & 0.35 & 82.63 & \ding{55} & $\ominus$ & I2T T2I \\
Bagel~\cite{bagel2025} & 7B+7B$^*$ & \textbf{0.88}\textsuperscript{\ddag} & 0.52/\textbf{0.70}\textsuperscript{\dag} & \underline{85.07} & \ding{55} & $\ominus$ & I2T T2I INT CoT \\
\midrule
\multicolumn{8}{l}{\textit{Frozen LLM}} \\
MetaQuery-XL~\cite{metaquery2025} & 7B+2B$^*$ & 0.80 & 0.55 & 82.05 & \ding{51} & \ding{51} & T2I \\
UniWorld-v1~\cite{uniworld2025} & 7B+12B$^*$+5B$^{**}$ & 0.80 & 0.55 & 81.38 & \ding{51} & \ding{51} & I2T T2I \\
OmniGen2~\cite{omnigen2_2025} & 3B+4B$^*$ & 0.80 & -- & 83.57 & \ding{51} & \ding{51} & T2I INT CoT \\
\midrule
\mural~3B & 3B+3B$^*$ & 0.81 & 0.46/0.56\textsuperscript{\dag} & 85.01 & \ding{51} & \ding{51} & T2I \\
\mural~7B & 7B+7B$^*$ & \underline{0.85} & 0.51/\underline{0.66}\textsuperscript{\dag} & \textbf{86.75} & \ding{51} & \ding{51} & T2I \\
\bottomrule
\end{tabular}
}
\end{table}

Among models that preserve both text and visual understanding (MetaQuery-XL, UniWorld-v1, OmniGen2, and \mural), \mural~achieves the best image generation metrics using only standard text-to-image training data. \mural~3B (6B total) surpasses MetaQuery-XL (9B), UniWorld-v1 (26B), and OmniGen2 (7B) on all benchmarks despite having fewer parameters. The metrics shown for \mural~3B~are at the end of 512p stage and should further improve during 1024p training. \mural~7B achieves the highest scores overall: GenEval (0.85), DPG-Bench (86.75), and WISE (0.66).

Bagel achieves a WISE score of 0.70 (Table~\ref{tab:unified_comparison}) with CoT by incorporating reasoning training data, but at the cost of degraded (denoted by \ding{55}) text and mixed (denoted by $\ominus$) visual understanding performance -- a consequence of complex data mixture optimization. \mural~7B (same size as Bagel), in contrast, leverages the LLM's reasoning capabilities at inference time without any interleaved or reasoning data during training, achieving a competitive WISE score of 0.66 while fully preserving (denoted by \ding{51}) all understanding capabilities.
The degraded understanding metrics observed in several prior unified models underscore precisely the complexity that the frozen-LLM approach in \mural~eliminates. Moreover, as shown in Section~\ref{SS:emergent}, this approach gives rise to emergent capabilities absent from the training data.

\noindent\textbf{Comparison with standalone T2I models.}
Table~\ref{tab:t2i_comparison} compares \mural~against dedicated text-to-image models.
\mural~achieves competitive results, with a DPG-Bench score of 86.75 that surpasses most standalone models including FLUX.1-dev (83.84), GPT Image~1 (85.15), and HiDream-I1 (85.89), and approaches QwenImage (88.32) at roughly half the parameter count and arguably smaller training budget.
Most notably, \mural's WISE score of 0.66 (with CoT) exceeds QwenImage (0.62), making a strong case for leveraging frozen LLM understanding via shared attention to improve knowledge-grounded image generation.

\begin{table}[t]
\centering
\caption{Comparison with standalone text-to-image models. $^*$denotes parameters of the image generation expert, $^{**}$parameters of the text encoder, and \textsuperscript{\dag}without RL. \mural~surpasses QwenImage in WISE while being competitive on other benchmarks.}
\footnotesize
\label{tab:t2i_comparison}
\begin{tabular}{lcccc}
\toprule
\textbf{Model} & \textbf{Params} & \ \textbf{GenEval}$\uparrow$ \ & \ \textbf{WISE}$\uparrow$ \ & \ \textbf{DPGBench}$\uparrow$ \ \\
\midrule
SDXL~\cite{sdxl2023} & 2.6B+0.7B$^{**}$ & 0.55 & 0.43 & 74.65 \\
DALL-E 3~\cite{dalle3_2023} & -- & 0.67 & -- & 83.5 \\
FLUX.1-dev~\cite{flux2024} & 12B+5B$^{**}$ & 0.82 & 0.50 & 83.84 \\
SD-3.5-Large~\cite{sd3_2024} & 8B+6B$^{**}$ & 0.71 & 0.46 & -- \\
Infinity~\cite{infinity2024} & 8B+3B$^{**}$ & 0.73 & -- & 83.46 \\
HiDream-I1~\cite{hidream2025} & 17B+22B$^{**}$ & 0.83 & -- & 85.89 \\
GPT Image 1~\cite{gpt4o2024} & -- & 0.84 & -- & 85.15 \\
Seedream 3.0~\cite{seedream2025} & -- & 0.84 & -- & 88.27 \\
QwenImage~\cite{qwenimage2025} & 20B+7B & \textbf{0.87}\textsuperscript{\dag} & 0.62 & \textbf{88.32} \\
\midrule
\mural~7B & 7B+7B$^*$ & 0.85 & \textbf{0.66} & 86.75\\
\bottomrule
\end{tabular}
\end{table}

\subsection{Model scaling}
\label{SS:scaling}

\begin{figure}[t]
\centering
\includegraphics[width=0.65\linewidth]{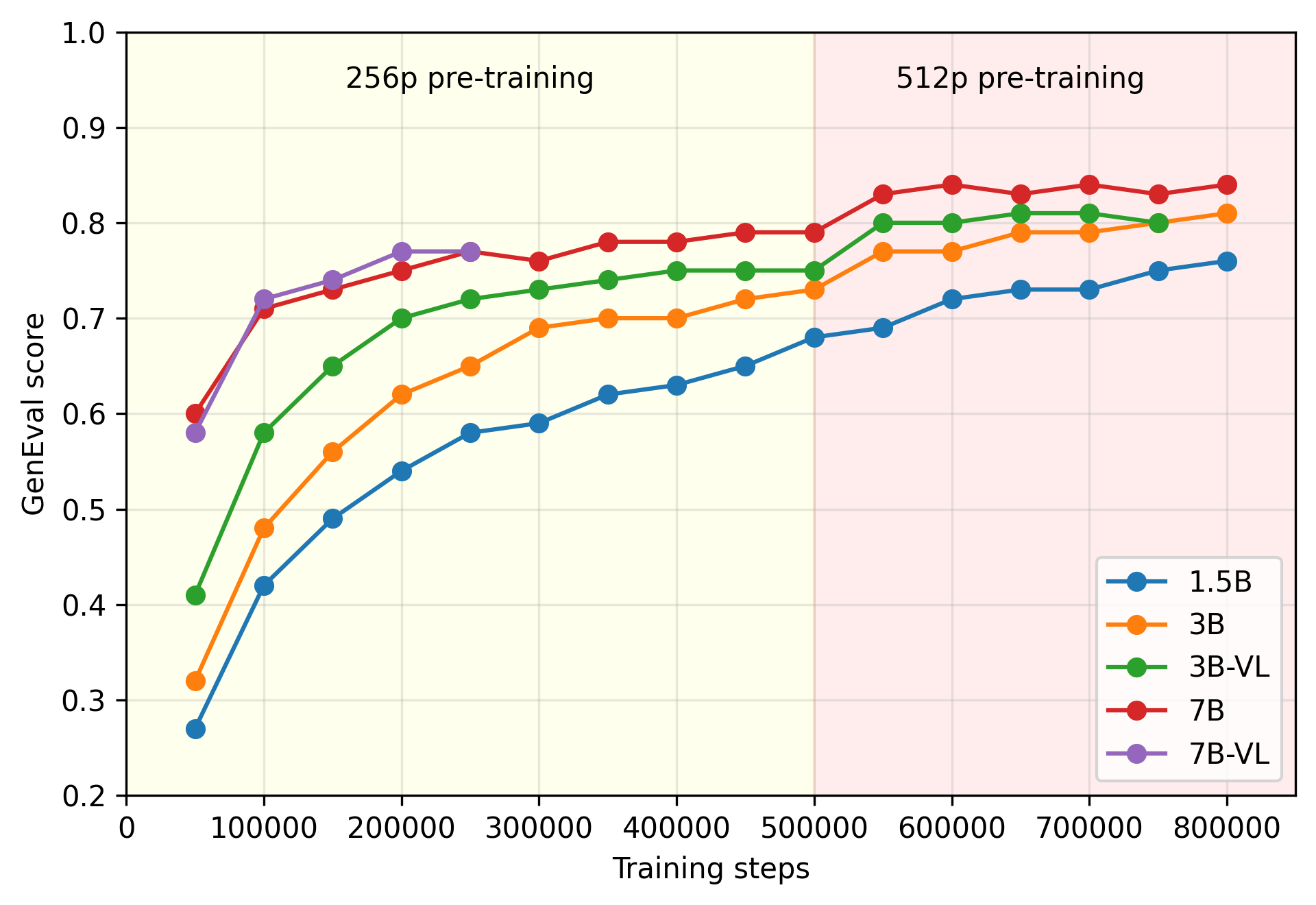}
\caption{GenEval scores across training steps for different model scales and initializations. Qwen2.5-VL initialization provides faster convergence at 3B scale, but this advantage diminishes at 7B.}
\label{fig:scaling}
\end{figure}

We train \mural~at three scales -- 1.5B, 3B, and 7B -- using both text-only (Qwen2.5) and vision-language (Qwen2.5-VL) initializations where available (the 1.5B model does not have a VL variant).
We report observations during pretraining at 256p and 512p stages.
Figure~\ref{fig:scaling} shows GenEval scores across training steps.
We observe consistent improvements with scale: 7B model achieves the highest scores, followed by 3B and 1.5B.
\mural~1.5B with CoT at inference does better \mural~3B and \mural~3B with CoT does better than \mural~7B on WISE.
Notably, the 3B model achieves results competitive with most unified models of larger size (as shown in Table~\ref{tab:unified_comparison}), making it a practical choice for resource-constrained settings.

We note an interesting observation with respect to vision-language initialization.
At the 3B scale, MoT initialized with Qwen2.5-VL-3B-Instruct converges faster than with Qwen2.5-3B-Instruct, suggesting that cross-modality alignment at initialization accelerates learning.
However, at the 7B scale, this advantage diminishes, indicating that larger models can learn the necessary cross-modal representations from scratch given sufficient data.

\begin{table}[t]
\centering
\caption{Image generation metrics at the end of 512p resolution training across model scales. WISE is reported as without / with CoT.}
\label{tab:scaling}
\begin{tabular}{lcccc}
\toprule
\textbf{Model} & \textbf{GenEval}$\uparrow$ & \textbf{WISE}$\uparrow$ & \textbf{DPGBench}$\uparrow$ \\
\midrule
\mural~1.5B & 0.76 & 0.38 / 0.47 & 82.27 \\
\mural~3B & 0.81 & 0.46 / 0.56 & 85.01 \\
\mural~7B & 0.84 & 0.50 / 0.65 & 86.16 \\
\bottomrule
\end{tabular}
\end{table}

\section{Discussion}

In this paper, we study how knowledge from frozen large language models can influence text-to-image generation when training is restricted to standard text–image pairs. 
Our results suggest that shared attention between a frozen LLM and an image generation model allows the generation process to access the knowledge and reasoning abilities encoded in the LLM, even though the LLM itself is never updated during training.
Across a range of experiments, we observe that this interaction gives rise to several emergent behaviors not present in the training data, including cross-lingual generation, color-guided composition, emoji-to-scene translation, and generation guided by world knowledge. 
These findings indicate that knowledge of pretrained language models can transfer to image generation systems without interleaved multimodal data mixtures or explicit reasoning supervision. 
Empirically, models trained under this setup achieve strong performance on compositional generation benchmarks such as DPG-Bench, GenEval, and WISE, while using only text-to-image training data. 
Scaling experiments further show consistent improvements across model sizes from 1.5B to 7B parameters.

\textbf{Limitations and future work.} 
The emergent capabilities demonstrated in this work are observed using Qwen2.5, a text-only LLM. 
Extending this study to multimodal LLM backbones such as Qwen2.5-VL (Figure~\ref{fig:scaling}) could reveal how visual reasoning capabilities interact with image generation. 
Such extensions may enable new applications including instruction-based image editing, subject-driven generation, and multi-image reasoning. 
More broadly, understanding how reasoning abilities from large pretrained models transfer across modalities remains an important direction for future work.

%
%
\bibliographystyle{splncs04}
\bibliography{main}

@String(CVPR  = {CVPR})

@String(ICCV  = {ICCV})

@String(NeurIPS = {NeurIPS})

@String(ICML  = {ICML})

@String(ICLR  = {ICLR})

@article{chameleon2024,
  title={Chameleon: Mixed-Modal Early-Fusion Foundation Models},
  author={{Chameleon Team}},
  journal={arXiv preprint arXiv:2405.09818},
  year={2024}
}

@article{transfusion2024,
  title={Transfusion: Predict the Next Token and Diffuse Images with One Multi-Modal Model},
  author={Zhou, Chunting and Yu, Lili and Babu, Arun and Tirumala, Kushal and Yasunaga, Michihiro and Shamis, Leonid and Kahn, Jacob and Ma, Xuezhe and Zettlemoyer, Luke and Levy, Omer},
  journal={arXiv preprint arXiv:2408.11039},
  year={2024}
}

@article{showo2024,
  title={Show-o: One Single Transformer to Unify Multimodal Understanding and Generation},
  author={Xie, Jinheng and Mao, Weijia and Bai, Zechen and Zhang, David Junhao and Wang, Weihao and Lin, Kevin Qinghong and Gu, Yuchao and Chen, Zhijie and Yang, Zhenheng and Shou, Mike Zheng},
  journal={arXiv preprint arXiv:2408.12528},
  year={2024}
}

@article{emu3_2024,
  title={Emu3: Next-Token Prediction is All You Need},
  author={Wang, Xinlong and Zhang, Xiaosong and Luo, Zhengxiong and Sun, Quan and Cui, Yufeng and Wang, Jinsheng and Zhang, Fan and Wang, Yueze and Li, Zhen and Yu, Qiying and others},
  journal={arXiv preprint arXiv:2409.18869},
  year={2024}
}

@article{januspro2025,
  title={Janus-Pro: Unified Multimodal Understanding and Generation with Data and Model Scaling},
  author={Chen, Xiaokang and Wu, Zhiyu and Liu, Xingchao and Pan, Zizheng and Liu, Wen and Xie, Zhenda and Yu, Xingkai and Ruan, Chong},
  journal={arXiv preprint arXiv:2501.17811},
  year={2025}
}

@article{bagel2025,
  title={Emerging Properties in Unified Multimodal Pretraining},
  author={Deng, Chaorui and Zhu, Deyao and Li, Kunchang and Gou, Chenhui and Li, Feng and Wang, Zeyu and Zhong, Shu and Yu, Weihao and Nie, Xiaonan and Song, Ziang and others},
  journal={arXiv preprint arXiv:2505.14683},
  year={2025}
}

@article{metaquery2025,
  title={Transfer between Modalities with MetaQueries},
  author={Pan, Xichen and Shukla, Satya Narayan and Singh, Aashu and Zhao, Zhuokai and Mishra, Shlok Kumar and Wang, Jialiang and Xu, Zhiyang and Chen, Jiuhai and Li, Kunpeng and Juefei-Xu, Felix and others},
  journal={arXiv preprint arXiv:2504.06256},
  year={2025}
}

@article{uniworld2025,
  title={Uniworld-v1: High-Resolution Semantic Encoders for Unified Visual Understanding and Generation},
  author={Lin, Bin and Li, Zongjian and Cheng, Xinhua and Niu, Yuwei and Ye, Yang and He, Xianyi and Yuan, Shenghai and Yu, Wangbo and Wang, Shaodong and Ge, Yunyang and others},
  journal={arXiv preprint arXiv:2506.03147},
  year={2025}
}

@article{omnigen2_2025,
  title={Omnigen2: Exploration to Advanced Multimodal Generation},
  author={Wu, Chenyuan and Zheng, Pengfei and Yan, Ruiran and Xiao, Shitao and Luo, Xin and Wang, Yueze and Li, Wanli and Jiang, Xiyan and Liu, Yexin and Zhou, Junjie and others},
  journal={arXiv preprint arXiv:2506.18871},
  year={2025}
}

@article{cosmos32026,
  title={{Cosmos 3: Omnimodal World Models for Physical AI}},
  author={Nvidia},
  journal={arXiv preprint arXiv:2606.02800},
  year={2026}
}

@inproceedings{lmfusion2025,
  title={{LMFusion}: {Adapting} Pretrained Language Models for Multimodal Generation},
  author={Shi, Weijia and Han, Xiaochuang and Zhou, Chunting and Liang, Weixin and Lin, Xi and Zettlemoyer, Luke and Yu, Lili},
  booktitle=NeurIPS,
  year={2025}
}

@article{mot2024,
  title={Mixture-of-Transformers: A Sparse and Scalable Architecture for Multi-Modal Foundation Models},
  author={Liang, Weixin and Yu, Lili and Luo, Liang and Iyer, Srinivasan and Dong, Ning and Zhou, Chunting and Ghosh, Gargi and Lewis, Mike and Yih, Wen-tau and Zettlemoyer, Luke and others},
  journal={arXiv preprint arXiv:2411.04996},
  year={2024}
}

@article{moe_shazeer2017,
  title={Outrageously Large Neural Networks: The Sparsely-Gated Mixture-of-Experts Layer},
  author={Shazeer, Noam and Mirhoseini, Azalia and Maziarz, Krzysztof and Davis, Andy and Le, Quoc and Hinton, Geoffrey and Dean, Jeff},
  journal={arXiv preprint arXiv:1701.06538},
  year={2017}
}

@article{qwen25_2024,
  title={Qwen2.5 Technical Report},
  author={Yang, An and Yang, Baosong and Zhang, Beichen and Hui, Binyuan and Zheng, Bo and Yu, Bowen and Li, Chengyuan and Liu, Dayiheng and Huang, Fei and Wei, Haoran and others},
  journal={arXiv preprint arXiv:2412.15115},
  year={2024}
}

@article{qwen2vl2024,
  title={Qwen2-{VL}: Enhancing Vision-Language Model's Perception of the World at Any Resolution},
  author={Wang, Peng and Bai, Shuai and Tan, Sinan and Wang, Shijie and Fan, Zhihao and Bai, Jinze and Chen, Keqin and Liu, Xuejing and Wang, Jialin and Ge, Wenbin and others},
  journal={arXiv preprint arXiv:2409.12191},
  year={2024}
}

@misc{gpt4o2024,
  title={{GPT-4o}},
  author={{OpenAI}},
  year={2024},
  howpublished={\url{https://openai.com/index/hello-gpt-4o/}}
}

@article{ddpm2020,
  title={Denoising Diffusion Probabilistic Models},
  author={Ho, Jonathan and Jain, Ajay and Abbeel, Pieter},
  journal=NeurIPS,
  year={2020}
}

@inproceedings{ldm2022,
  title={High-Resolution Image Synthesis with Latent Diffusion Models},
  author={Rombach, Robin and Blattmann, Andreas and Lorenz, Dominik and Esser, Patrick and Ommer, Bj{\"o}rn},
  booktitle=CVPR,
  year={2022}
}

@inproceedings{sdxl2023,
  title={{SDXL}: Improving Latent Diffusion Models for High-Resolution Image Synthesis},
  author={Podell, Dustin and English, Zion and Lacey, Kyle and Blattmann, Andreas and Dockhorn, Tim and M{\"u}ller, Jonas and Penna, Joe and Rombach, Robin},
  booktitle=ICLR,
  year={2024}
}

@article{dalle3_2023,
   title={Improving image generation with better captions},
  author={Betker, James and Goh, Gabriel and Jing, Li and Brooks, Tim and Wang, Jianfeng and Li, Linjie and Ouyang, Long and Zhuang, Juntang and Lee, Joyce and Guo, Yufei and others},
  journal={Computer Science},
  year={2023}
}

@inproceedings{sd3_2024,
  title={Scaling Rectified Flow Transformers for High-Resolution Image Synthesis},
  author={Esser, Patrick and Kulal, Sumith and Blattmann, Andreas and Entezari, Rahim and M{\"u}ller, Jonas and Saini, Harry and Levi, Yam and Lorenz, Dominik and Sauer, Axel and Boesel, Frederic and others},
  booktitle=ICML,
  year={2024}
}

@misc{flux2024,
  title={{FLUX.1}},
  author={{Black Forest Labs}},
  year={2024},
  howpublished={\url{https://blackforestlabs.ai/}}
}

@inproceedings{dit2023,
  title={Scalable Diffusion Models with Transformers},
  author={Peebles, William and Xie, Saining},
  booktitle=ICCV,
  year={2023}
}

@inproceedings{uvit2023,
  title={All are Worth Words: A {ViT} Backbone for Diffusion Models}, 
  author={Fan Bao and Shen Nie and Kaiwen Xue and Yue Cao and Chongxuan Li and Hang Su and Jun Zhu},
  booktitle=CVPR,
  year={2023},
}

@article{seedream2025,
  title={Seedream 3.0 Technical Report},
  author={Gao, Yu and Gong, Lixue and Guo, Qiushan and Hou, Xiaoxia and Lai, Zhichao and Li, Fanshi and Li, Liang and Lian, Xiaochen and Liao, Chao and Liu, Liyang and others},
  journal={arXiv preprint arXiv:2504.11346},
  year={2025}
}

@article{hidream2025,
  title={{HiDream-I1}: A High-Efficient Image Generative Foundation Model with Sparse Diffusion Transformer},
  author={Cai, Qi and Chen, Jingwen and Chen, Yang and Li, Yehao and Long, Fuchen and Pan, Yingwei and Qiu, Zhaofan and Zhang, Yiheng and Gao, Fengbin and Xu, Peihan and others},
  journal={arXiv preprint arXiv:2505.22705},
  year={2025}
}

@article{qwenimage2025,
    title={Qwen-Image Technical Report}, 
    author={Chenfei Wu and Jiahao Li and Jingren Zhou and Junyang Lin and Kaiyuan Gao and Kun Yan and Sheng-ming Yin and Shuai Bai and Xiao Xu and Yilei Chen and Yuxiang Chen and Zecheng Tang and Zekai Zhang and Zhengyi Wang and An Yang and Bowen Yu and Chen Cheng and Dayiheng Liu and Deqing Li and Hang Zhang and Hao Meng and Hu Wei and Jingyuan Ni and Kai Chen and Kuan Cao and Liang Peng and Lin Qu and Minggang Wu and Peng Wang and Shuting Yu and Tingkun Wen and Wensen Feng and Xiaoxiao Xu and Yi Wang and Yichang Zhang and Yongqiang Zhu and Yujia Wu and Yuxuan Cai and Zenan Liu},
    journal={arXiv preprint arXiv:2508.02324},
    year={2025},
}

@article{llamagen2024,
  title={Autoregressive Model Beats Diffusion: {Llama} for Scalable Image Generation},
  author={Sun, Peize and Jiang, Yi and Chen, Shoufa and Zhang, Shilong and Peng, Bingyue and Luo, Ping and Yuan, Zehuan},
  journal={arXiv preprint arXiv:2406.06525},
  year={2024}
}

@inproceedings{mar2024,
  title={Autoregressive Image Generation without Vector Quantization},
  author={Li, Tianhong and Tian, Yonglong and Li, He and Deng, Mingyang and He, Kaiming},
  booktitle=NeurIPS,
  year={2024}
}

@inproceedings{infinity2024,
  title={Infinity: Scaling Bitwise AutoRegressive Modeling for High-Resolution Image Synthesis},
  author={Han, Jian and Liu, Jinlai and Jiang, Yi and Yan, Bin and Zhang, Yuqi and Yuan, Zehuan and Peng, Bingyue and Liu, Xiaobing},
  booktitle=CVPR,
  year={2025}
}

@inproceedings{flowmatching2023,
  title={Flow Matching for Generative Modeling}, 
  author={Yaron Lipman and Ricky T. Q. Chen and Heli Ben-Hamu and Maximilian Nickel and Matt Le},
  booktitle=ICLR,
  year={2023},
}

@article{geneval2023,
  title={Geneval: An object-focused framework for evaluating text-to-image alignment},
  author={Ghosh, Dhruba and Hajishirzi, Hannaneh and Schmidt, Ludwig},
  journal=NeurIPS,
  year={2023}
}

@article{dpgbench2024,
  title={{ELLA}: Equip Diffusion Models with {LLM} for Enhanced Semantic Alignment},
  author={Hu, Xiwei and Wang, Rui and Fang, Yixiao and Fu, Bin and Cheng, Pei and Yu, Gang},
  journal={arXiv preprint arXiv:2403.05135},
  year={2024}
}

@article{wise2024,
  title={{WISE}: A World Knowledge-Informed Semantic Evaluation for Text-to-Image Generation},
  author={Niu, Yuwei and Ning, Munan and Zheng, Mengren and Jin, Weiyang and Lin, Bin and Jin, Peng and Liao, Jiaqi and Feng, Chaoran and Ning, Kunpeng and Zhu, Bin and others},
  journal={arXiv preprint arXiv:2503.07265},
  year={2025}
}

@article{rmsnorm2019,
  title={Root Mean Square Layer Normalization},
  author={Zhang, Biao and Sennrich, Rico},
  journal=NeurIPS,
  year={2019},
}

@inproceedings{adamw2019,
    title={Decoupled Weight Decay Regularization}, 
    author={Ilya Loshchilov and Frank Hutter},
    booktitle=ICLR,
    year={2019},
}
\end{document}